%% file: main.tex
\documentclass[english,11pt, letterpaper]{article}
\usepackage[T1]{fontenc}
\usepackage[utf8]{inputenc}
\usepackage{pifont} 
\usepackage{geometry}

\usepackage{amsmath}
\usepackage{picinpar}
\usepackage{multicol}
\usepackage{amsthm}
\usepackage{amssymb}
\usepackage{algorithm}
\usepackage{algorithmic}
\newlength\myindent
\usepackage{setspace}
\usepackage{amsmath,amsfonts,bm,amsthm}
\usepackage{xfrac}
\usepackage{empheq}
\usepackage{natbib} 
\usepackage{float}
\usepackage{wrapfig}
\usepackage{verbatim}
\usepackage{empheq}

\makeatletter
\numberwithin{equation}{section}
\numberwithin{figure}{section}
\theoremstyle{plain}
\newtheorem{thm}{\protect\theoremname}
\theoremstyle{definition}

\theoremstyle{plain}

\theoremstyle{plain}

\pdfoutput=1

\pdfoutput=1

\usepackage{amsmath, amsthm, amssymb, amsfonts,bbold}
\usepackage[basic]{complexity}
\usepackage[pdftex,bookmarks,colorlinks]{hyperref}
\usepackage{enumitem}

\DeclareMathOperator*{\argmaxTex}{arg\,max}
\DeclareMathOperator*{\argminTex}{arg\,min}
\DeclareMathOperator*{\signTex}{sign}
\DeclareMathOperator*{\rankTex}{rank}
\DeclareMathOperator*{\diagTex}{diag}
\DeclareMathOperator*{\imTex}{im}

\hypersetup{colorlinks=true,citecolor=red}

\renewcommand{\varepsilon}{\epsilon}

\usepackage{thmtools}
\usepackage{thm-restate}
\usepackage[usenames]{color}

\usepackage[lined,boxed,ruled,norelsize,algo2e]{algorithm2e}
\usepackage{tikz}
\usetikzlibrary{plotmarks}
\usepackage[small,compact]{titlesec}
\usepackage{setspace}
\usepackage{subfigure}

\usepackage{tcolorbox}

\makeatother
\title{The Optimality of (Accelerated) SGD  for\\ High-Dimensional
Quadratic Optimization}
\usepackage{babel}
\providecommand{\corollaryname}{Corollary}
\providecommand{\definitionname}{Definition}
\providecommand{\lemmaname}{Lemma}
\providecommand{\theoremname}{Theorem}

\author{%
  Haihan Zhang$^*$, Yuanshi Liu$^*$, Qianwen Chen, Cong Fang\textsuperscript{\ding{41}}, \vspace{5pt}\\
     School of Intelligence Science and Technology, Peking University \\
}

\date{}
\begin{document}

\include{mathdefs}

\include{notation}
\global\long\def\vol{\mathrm{vol}}%
\global\long\def\rPos{\R_{> 0}}%
\global\long\def\dWeight{\R_{>0}^{m}}%
\global\long\def\dWeights{\rPos^{m}}%

\global\long\def\cO{\mathcal{O}}%
\global\long\def\tO{\tilde{\cO}}%

\global\long\def\specGeq{\succeq}%

\global\long\def\specLeq{\preceq}%

\global\long\def\specGt{\succ}%

\global\long\def\specLt{\prec}%
\global\long\def\gradient{\nabla}%
\global\long\def\weight{w}%

\global\long\def\vWeight{\vvar{\weight}}%

\global\long\def\mWeight{\mvar W}%
\global\long\def\mNormProjLap{\bar{\mLambda}}%
\global\long\def\volPot{\mathcal{V}}%
\global\long\def\dInterior{\Omega^{\circ}}%
\global\long\def\dFull{\{\dInterior\times\R_{>0}^{m}\}}%
\global\long\def\vWeight{\vvar w}%

\global\long\def\mWeight{\mvar W}%
\global\long\def\polytope{\Omega}%
\global\long\def\interior{\Omega^{\circ}}%
\maketitle

\begin{abstract}
Stochastic gradient descent (SGD) is a widely used algorithm in machine learning, particularly for neural network training. Recent  studies on SGD  for  canonical quadratic optimization or linear regression show it attains well generalization under suitable high-dimensional settings.  %benign overfitting or attains well generalization with no explicit regularization
However, a fundamental question\textbf{---}\emph{for what kinds of high-dimensional learning problems SGD and its accelerated variants can achieve optimality}\textbf{---} has yet to be well studied. This paper investigates SGD with two essential components in practice: exponentially decaying step size schedule and momentum. We establish the convergence upper bound for momentum accelerated SGD (ASGD) and propose concrete classes of learning problems under which SGD or ASGD achieves  min-max \emph{optimal} convergence rates. The characterization of the target function is based on standard power-law decays in (functional) linear regression. 
Our results unveil new insights for understanding the learning bias of SGD: 
(i)  SGD  is efficient in learning ``dense'' features where the corresponding weights are subject to an $\ell_{\infty}$ constraint; (ii) SGD is efficient for easy problem without suffering from the saturation effect; (iii)  momentum  can  accelerate the convergence rate by order when the learning problem is relatively hard.  
To our knowledge, this is the \emph{first} work to clearly identify the optimal boundary of SGD versus ASGD for the problem under mild settings.

\end{abstract}

\footnotetext[1]{Equal Contribution.}

\newpage

\section{Introduction}\label{sec: intro}

In the realm of solving large-scale machine learning problems, stochastic gradient descent (SGD)~\citep{Robbins1951ASA} plays an indispensable role and has achieved significant success in practice \citep{lecun1998gradient,bengio2009learning,bottou2007tradeoffs,lecun2015deep}.

Towards the analysis of SGD, a majority of theoretical works delve into stochastic quadratic optimization problems and focus on the optimality of SGD within these settings~\citep{dieuleveut2016nonparametric,Jain2016ParallelizingSG,ge2019step}. Although quadratic optimization problems are %among 
plausibly simple, %learning tasks, 
they involve fundamental aspects of machine learning such as linear regression~\citep{Hsu2011RandomDA}, over-parametrized models~\citep{gunasekar2017implicit,li2018algorithmic}, and neural networks in the kernel regime~\citep{jacot2018neural,arora2019fine}. Meanwhile, %we are able to
one is able to conduct  %rigorous theoretical
theoretical
fine-grained
 analysis for quadratic optimization,  providing new insights for  general  algorithm analysis and design. 
Given these foundational aspects, we restrict our analysis to this canonical
%fundamental
problem, namely,
\begin{equation}  \label{eq: objective}
    \min_{\mathbf{w}\in\mathbb{R}^d }f\left (\mathbf{w}  \right ) =\mathbb{E}_{\bm{\xi} }\left [ F\left ( \mathbf{w} ,\bm{\xi} \right )  \right ], \   F\left ( \mathbf{w} ,\bm{\xi}  \right )=\frac{1}{2}\mathbf{w}^{\top}\mathbf{H}\left (  \bm{\xi}\right ) \mathbf{w} -\mathbf{b}\left ( \bm{\xi} \right ) ^{\top }\mathbf{w},
\end{equation}
where $\bm\xi\sim\mathbb{P}_{\bm\xi}$ encodes the randomness, and $\mathbf{H}(\bm{\xi})$ is symmetric and positive semidefinite. We denote the optimum of $f$ in \eqref{eq: objective} as $\mathbf{w}_*$. For this stochastic optimization problem, any algorithm after inputting  $\mathbf{w}$  receives an unbiased stochastic gradient   $\hat{\nabla}f(\mathbf{w}) = \mathbf{H}(\bm{\xi})\mathbf{w} - \mathbf{b}(\bm{\xi})$,  where $\bm{\xi}$ is an independent sample drawn from the 
distribution $\mathbb{P}_{\bm\xi}$.  A critical assumption regarding \eqref{eq: objective} is the anisotropic gradient noise, given by $\mathrm{Var}\left(\tilde{\nabla} f(\mathbf{w})\right) \preceq \sigma^2 \mathbb{E}\left[\mathbf{H}(\bm\xi)\right]$, which is adopted  in previous quadratic optimization problems such as least squares regression and stochastic approximation \citep{bach2013non,dieuleveut2017harder,pan2021eigencurve,pan2023accelerated}. The specific structure of anisotropic gradient noise makes it possible to obtain more accurate conclusions compared to scenarios with unstructured noise~\citep{ghadimi2012optimal}.

In the fundamental problem described above,  the estimation behavior is primarily governed by two separate factors, according to the widely used bias-variance decomposition~\citep{Hsu2011RandomDA,bach2013non,jain2017markov}. Here, the estimation variance term $\mathrm{Var}(\hat{\mathbf{w}})$, which stems from gradient noise, reflects the statistical challenges inherent in the estimation process. The bias term, defined as the deviation between $\hat{\mathbf{w}}$ and $\mathbf{w}_*$,  represents the behavior of the noiseless optimization process.

The low-dimensional setting, where the gradient query number significantly exceeds the parameter dimension $d$, has been thoroughly examined in existing works~\citep{defossez2014constant,dieuleveut2016nonparametric,Jain2016ParallelizingSG,jain2017markov,dieuleveut2017harder,jain2018accelerating,ge2019step,pan2023accelerated}. In this scenario, the large number of gradient queries ensures the convergence of the bias term. Consequently, the primary difficulty resides in the variance caused by the stochastic gradient noise, leading to the classical $d\sigma^2/T$ optimal excess risk~\citep{van2000asymptotic,polyak1992acceleration}, where $T$ refers to the number of stochastic gradient oracle accesses. This optimal rate is achieved by both SGD and its acceleration using tail-averaging \citep{defossez2014constant,dieuleveut2016nonparametric,Jain2016ParallelizingSG,jain2017markov,dieuleveut2017harder,jain2018accelerating} or exponential step decay schedule \citep{ge2019step,pan2023accelerated}.

When it comes to the modern high-dimensional setting, fewer gradient queries provide insufficient information to address high-dimensional, ill-conditioned problems. In such cases, the deviation $\mathbf{w}_* $ significantly affects the optimality condition. Treating the noise level $\sigma^2$ as a constant and adopting the common prior assumptions of polynomially decaying Hessian eigenvalues and $\mathbf{w_*}$ coordinates results in a dimension-free optimal rate that is asymptotically slower than $1/T$~\citep{caponnetto2007optimal,dieuleveut2016nonparametric}.
In recent progress, there have been substantial studies on SGD for linear
regression that show it attains well generalization under high-dimensional
settings~\citep{zou2021benign,wu2022last,li2023risk}. However, relatively limited studies consider when SGD can achieve optimal convergence rates.
 Existing theoretical works~\citep{dieuleveut2016nonparametric,dieuleveut2017harder} consider vanilla SGD and show its optimality within a narrow regime of decay rate, with the goal of it to the ridge regression method.  
 A  more comprehensive understanding of the performance and the optimality of SGD with its variants under different decay rates remains open.

Our paper delves into the study of SGD along the thread of high-dimensional problems, adopting a more realistic framework that incorporates the exponentially decaying step size schedule~\citep{goffin1977convergence,ge2019step} and the momentum acceleration~\citep{polyak1964some}. Theoretical analysis substantiates the efficiencies of SGD with these practical tricks in the context of \eqref{eq: objective}. We prove that these enhancements can effectively control the last-iterate variance, accelerate the convergence of the bias term, and enable SGD to achieve optimal rates across a broader range of settings, as discussed below.

Our paper considers the functions whose Hessian eigenvalues have a polynomial decay rate and the optimum $\mathbf{w}_*$ satisfies the source condition. These are commonly used assumptions for evaluating optimality, particularly for non-parametric regression~\citep{cai2006prediction,caponnetto2007optimal,dieuleveut2016nonparametric}.
Specifically, with constants $a,b>1$, we assume that the Hessian eigenvalues decay as $\lambda_i\asymp i^{-a}$  and that  $\mathbf{w}_*^{(i)}$ the optimum $\mathbf{w}_*$ in the $i$-th Hessian eigenvector coordinate, satisfies $\lambda_i\left(\mathbf{w}^{(i)}\right)^2\leq i^{-b}$.
 Our results show that SGD variants attain optimality when $a\leq 2b$, whereas the analysis on vanilla SGD achieves optimality only when $a\leq b$. 
When the noise $\sigma^2$ is treated as a constant, SGD variants can achieve optimality even when $b\leq a \leq 2b$, where vanilla SGD is shown to be suboptimal. %Given the large dimension $d$, 
This setting actually corresponds to a highly misspecified setting in  kernel regression. This optimality persists until entering the region $a>2b$, where further acceleration becomes infeasible. 
%Addition to the constant $\sigma^2$ analysis, the results also include the small noise level setting.
Furthermore, we provide optimality conditions for any instance of the Hessian. Namely, we show that the vanilla SGD with exponentially decaying step size schedule achieves instance optimality for any Hessian subject to an $\ell_{\infty}$ constraint on $\mathbf{w}_*$ with respect to the standard orthonormal basis dictated by the Hessian eigenspace.

Towards the claims above, the analysis addresses technical challenges associated with the non-commutative diminishing matrices introduced by momentum and step decay, establishing a last-iterate error upper bound within this unified framework. Meanwhile, following the min-max error framework~\citep{yang1999information}, we construct a binary hypothesis testing under the constrained risk inequality over high dimension~\citep{brown1996constrained}, and  attain an information-theoretic lower bound on \eqref{eq: objective} for the function class that has a prior constraint on the optimum $\mathbf{w}_*$ and any instance Hessian.
 Both lower and upper bound are based on finer eigenstructure analysis and adapt to classical results.

Understanding when SGD with practical tricks achieves optimality is a fundamental area of research in machine learning. The ``no free lunch" principle~\citep{wolpert1997no} suggests that no single algorithm is optimal for all types of problems. This indicates that identifying the learning bias inherent in SGD is crucial for determining the specific problems it best addresses, thus guiding informed decisions about its practical application.
To further investigate this,  we present key insights into when SGD, equipped with the exponential step decay schedule and momentum, achieves optimality across various settings. Our key findings include:
%Such analysis helps illustrate the learning bias of SGD. The high-level implications of our work include: 
(i) SGD  is efficient in learning ``dense'' features whose corresponding weights have an $l_{\infty}$ constraint; 
(ii) SGD is efficient when the learning problem is easy and does not suffer from the saturation effect faced by kernel ridge regression (KRR);
(iii) The momentum technique can truly accelerate the convergence rate by order when a learning problem is hard  (or highly misspecified in kernel regression).

For (i), the key insight for generalization lies in the algorithm's implicit preference, which plays the role
of regularization/bias.  In offline quadratic optimization, previous studies \citep{soudry2018implicit,gunasekar2017implicit} have shown that the full gradient descent (GD) converges to the minimum-norm or sparse solution under different parameterizations.  In comparison,   our result reveals that  anisotropy of gradient noise instead leads the algorithm learning towards a ``dense'' solution. Building on these findings, we provide concrete examples to demonstrate the provable statistical efficiency of SGD over offline GD.

For (ii), the key insight for generalization comes from the implicit regularization of SGD with an appropriate initial step size, which can adaptively select an effective dimension based on the smoothness of the optimum $\mathbf{w}_*$, thereby achieving optimality.  In contrast, offline algorithms such as KRR use explicit regularization, which incorporates all dimensions into the error and increases the bias term when the target function is exceptionally smooth, preventing it from achieving optimality---a phenomenon commonly referred to as the saturation effect~\citep{neubauer1997converse,li2023on}. This suggests that when the optimum $\mathbf{w}_*$ is sufficiently smooth, SGD outperforms KRR.

For  (iii), since the early work from \citet{polyak1964some,nemirovskij1983problem,10.1145/3055399.3055464,carmon2018accelerated}, the momentum technique is provably efficient in optimization, as it can accelerate convergence rates by order for quadratic, convex, and non-convex optimization under broad conditions. However, despite the wide use of the trick, limited work has demonstrated provable facilitation in improving statistical efficiency. For example, for general convex optimization where bounded optimum norm $\left \| \mathbf{w}  \right \| _2$ and noise are assumed, \citet{shamir2013stochastic} and \citet{jain2019making} show that SGD with polynomially decaying step size already
achieves the optimal $1/\sqrt{T}$ rate of convergence. \citet{pan2023accelerated} considers ASGD for quadratic optimization in the low-dimensional regime, and obtains $\sqrt{\kappa}+ d\sigma^2/T$ rate. Though acceleration in optimization leads to faster bias convergence, it does not improve the order of variance convergence, which is the primary error when the noise is regarded as constant. Thus, this advantage holds only when the noise is small. \citet{dieuleveut2017harder} shows that averaged accelerated regularized gradient descent can achieve optimality in a broader region; however, adding additional regularization makes it difficult to determine whether the improvement stems from the regularization or the momentum. To the best of our knowledge, this paper serves as the \emph{first} work that justifies the statistical advantages under mild settings, novelly based on the slow decay of the source condition.

\subsection{Our Contributions}
To summarize our contributions:
\begin{itemize}
    \item We provide the upper bound of last-iterate error for SGD with exponentially decaying step size schedule on the high-dimensional objective with respect to the magnitude of the added momentum, under the anisotropic gradient noise assumption.
    \item We establish an instance-dependent information-theoretic lower bound for algorithms observing $T$ independent data points $\left \{ \left ( \mathbf{H} \left ( \bm{\xi }_t \right ) ,\mathbf{b} \left ( \bm{\xi }_t  \right )  \right ) \right \} _{t=0}^{T-1}$.
    \item We conduct a comprehensive analysis of when SGD, with both exponentially decaying step size schedule and momentum acceleration, achieves optimality for quadratic optimization problems.
\end{itemize}

\section{Related Works}

SGD is the workhorse algorithm in machine learning.  A common observation from practical applications is the anisotropy of stochastic gradient noise~\citep{neyshabur2017exploring, zhu2018anisotropic},  under which vanilla SGD with polynomially decaying step size is highly sub-optimal even for the basic quadratic optimization or linear regression problem \citep{ge2019step}. In the following, we review existing quadratic optimization works that we believe are most related to ours.

\noindent \textbf{(Accelerated) SGD for low-dimensional quadratic optimization problems}  \qquad 
In low-dimensional settings, for stochastic quadratic optimization problems, \cite{ge2019step} shows that SGD with exponentially decaying step size can nearly reach the min-max lower bound $\widetilde{\mathcal{O}}\left(d\sigma^2/T\right)$ while vanilla SGD with polynomially decaying
step size is highly sub-optimal. \citet{pan2021eigencurve} further provides a family of stepsize decaying schedules that can achieve optimality for SGD. \citet{pan2023accelerated} shows that adding momentum can accelerate SGD with exponential decaying stepsize in the bias term while the variance can achieve optimality. In addition to SGD with decaying stepsize, an alternative approach towards theoretical guarantees of SGD is the combination of the constant stepsize and the tail-averaging method. Compared with decaying stepsize, constant-step-size and tail-averaging methods are less practical but easier to analyze theoretically. Using this method,  many works such as \citet{bach2013non} and \citet{Jain2016ParallelizingSG} show that SGD can achieve optimality. \citet{dieuleveut2017harder,jain2018accelerating,varre2022accelerated} further study the Accelerated SGD and show that adding momentum can accelerate SGD in the bias term while the variance can achieve optimality. 

\noindent \textbf{(Accelerated) SGD for high-dimensional quadratic optimization problems}  \qquad 
\citet{wu2022last} extend the analysis of \citet{ge2019step} to high-dimensional setting, providing a dimension-free and instance-dependent convergence rate for SGD with exponentially decaying step size. As for SGD with constant-stepsize and tail-averaging, \citet{zou2021benign} offers a dimension-free and instance-dependent convergence rate for SGD within each eigen-subspace of the data covariance matrix. \citet{li2023risk} further studies Accelerated SGD, and offers a dimension-free and instance-dependent convergence rate for Accelerated SGD.  To our knowledge, there is still no work analyzing the convergence of accelerate SGD with decaying stepsize in high-dimensional setting.  
Moreover, an understanding of the optimality of SGD and its variants for high-dimensional problems remains  limited.

\noindent \textbf{Nonparametric linear regression}  \qquad 
Our optimality arguments focus on 
the class of target functions that is closely related to the common characterization in nonparametric linear regression~ \citep{cai2006prediction,caponnetto2007optimal,dieuleveut2016nonparametric,zhang2023optimality}, which assumes the polynomially decaying Hessian eigenvalues and the source condition. However,  in nonparametric linear regression, most existing works study offline learning, such as kernel ridge regression~\citep{raskutti2014early,zhang2023optimality}, the minimum-norm solution~\citep{Bartlett_2020}, and full GD with early stopping~\citep{raskutti2014early}, except \citet{dieuleveut2016nonparametric,dieuleveut2017harder}, which study (Accelerated) SGD with constant stepsize and tail-averaging for kernel regression.  Differently, our work analyzes SGD with step
decay schedule and momentum, which aligns better with practical scenarios. 
Our proof is also more involved in dealing with the bias term and provides the convergence rate for each coordinate.  As a result, their optimal region is restricted to $a \le b \le 2a + 1$, which is much narrower compared to ours.

 \noindent \textbf{Implicit Regularization over Linear Models}  \qquad 
When the number of model parameters exceeds the number of training samples, the generalization without explicit regularity often becomes ill-posed.  The key insight
is the implicit regularization imposed by the learning algorithm. There are typically two types of methods to characterize the implicit regularization.  The first one concentrates on offline problems and offers the exact formula of the solution trained by the learning algorithm \citep{gunasekar2017implicit,soudry2018implicit,ji2019implicit,chizat2020implicit}.  For example,  the pioneering work from \citet{soudry2018implicit,chizat2020implicit} demonstrates that the iterates of
GD converges to the max-margin solution for logistic regression when the data is separable.  For linear regression, GD would converge to the minimum-norm solution \citep{gunasekar2017implicit}. The other way instead seeks the  classes of learning problems where 
 the algorithm achieves benign performances,
bridging direct connections between the algorithms and the problems~\citep{gunasekar2017implicit,li2018algorithmic,haochen2021shape,pmlr-v202-jin23a}. For example, when for $d\times r$ low-rank matrix sensing problem, \citet{li2018algorithmic} shows that GD can find a tolerant solution in $\widetilde{\mathcal{O}}  \left ( dr^2 \right ) $ training samples.  Similarly, by injecting suitable label noise, GD can successfully find the solution for sparse linear regression~\citep{haochen2021shape}.
Our work follows the second line of the research and proposes the function classes under the power-law decay of the Hessian eigenvalues and the source condition. Furthermore, we show that, unlike previous works, SGD tends to learn  ``dense'' features whose corresponding weights have an $l_{\infty}$ constraint.

\section{Problem Setup and Preliminaries}
\textbf{Notations.} We use the convention 
$\mathcal{O} \left ( \cdot  \right ) $ and $\Omega \left ( \cdot  \right ) $ to denote lower and upper bounds with a universal constant.
 $\widetilde{\mathcal{O} } \left ( \cdot  \right ) $ and $\widetilde{\Omega  } \left ( \cdot  \right ) $  ignore the   polylogarithmic dependence. Use $f\lesssim g$ to denote $f=\mathcal{O} \left ( g  \right ) $, $f\gtrsim g$ to denote $f=\Omega \left ( g \right ) $. We use notation $f\asymp g$ if $g\lesssim f\lesssim g$. The Frobenius norm is denoted by $\left \| \cdot \right \|_F$ while $\left \| \cdot \right \|$ stands for operator 2-norm for matrices. $|\mathcal{A}|$ denotes the cardinality of the set $\mathcal{A}$.
Bold lowercase letters, for example, $\mathbf{x} \in \mathbb{R}^d$, denote vectors, while bold uppercase letters, for example, $\mathbf{A} \in \mathbb{R}^{m \times n}$, denote matrices. For matrices $\mathbf{A}, \mathbf{B}\in \mathbb{R}^{m \times n}$, their inner product is defined as $\left \langle \mathbf{A},\mathbf{B} \right \rangle \equiv \mathrm{tr}\left ( \mathbf{A}^{\top}\mathbf{B} \right )  $. For matrices $\mathbf{A}, \mathbf{B}\in \mathbb{R}^{m \times m}$, we use notation $\mathbf{A} \preceq \mathbf{B} $, if $\mathbf{B}-\mathbf{A}$ is a positive semidefinite matrix. We use $\mathbf{O}\in \mathbb{R}^{m\times n}$ to denote a matrix whose all elements are zero. Moreover, we let $a\vee b=\mathrm{max}\left \{ a,b \right \}  $ and $a\wedge  b=\mathrm{min}\left \{ a,b \right \}  $. For random vector $\mathbf{x}$, $\mathrm{Var}(\mathbf{x}) = \left(\mathbf{x} - \mathbb{E}\mathbf{x}\right)\left(\mathbf{x} - \mathbb{E}\mathbf{x}\right)^{\top}$ denotes its variance matrix.

We refer to $(\mathbf{H}, \mathbf{b})$ as the population version of $(\mathbf{H}(\bm\xi), \mathbf{b}(\bm\xi))$ in \eqref{eq: objective} thereby the Hessian of objective \eqref{eq: objective} is $\mathbf{H}$ and $\mathbf{w}_* = \mathbf{H}^{-1}\mathbf{b}$. The eigenstructure of $\mathbf{H}$ is expressed as
\begin{equation}\nonumber
    \mathbf{H}\equiv \nabla ^2f\left ( \mathbf{w}  \right ) =\sum_{i=1}^d \lambda_i \mathbf{v}_i \mathbf{v}_i^{\top}=\mathbf{V} \bm{\Sigma} \mathbf{V}^{\top}.
\end{equation}
with descending eigenvalues $\lambda_1\geq\lambda_2\geq\cdots\geq\lambda_d$. With this quadratic structure, a finer dimension-wise analysis can be conducted in the eigenspace dictated by $\mathbf{H}$. For vector $\mathbf{w} $ in $\mathbb{R}^d$, we denote $\mathbf{w} ^{\left ( i \right )} = \mathbf{v}_i^{\top} \mathbf{w}$ as its $i$-th coordinate with respect to the above eigenbasis.

To rigorously characterize the stochastic quadratic problem in \eqref{eq: objective} in the high-dimensional setting, we adopt the stochastic approximation framework as in \citet{kushner2012stochastic}. At the $t$-th iteration, algorithms can access the stochastic gradient based on $\left ( \mathbf{H} \left ( \bm{\xi }_t \right ) ,\mathbf{b} \left ( \bm{\xi }_t  \right )  \right ) $, namely $\nabla_{\mathbf{w}} F(\mathbf{w}_t,\bm\xi_t)$,
where $\left\{  \bm{\xi }_t \right\} _{t=0}^{\infty }$ follow $\mathbb{P}_{\bm{\xi }}$ independently and identically. With the gradient noise denoted as $\bm{\zeta}_t = \hat{\nabla} f(\mathbf{w}_t) - \nabla f(\mathbf{w}_t)$, the anisotropic gradient noise assumption is rigorously defined as follows.
\begin{assumption}\label{a2}
 The randomness $\left \{  \bm{\xi}_t \right \} _{t=0}^{\infty }$ are independent. The expectation and variance of the gradient noise satisfy
\begin{equation}\label{noise}
  \mathbb{E} \left[\bm{\zeta}_t\mid  \mathbf{w}_t  \right] =0, \quad~~
  \mathbb{E}\left[\bm{\zeta}_t\bm{\zeta}_t^{\top}\mid \mathbf{w}_t \right] \preceq  \sigma^2 \mathbf{H}.
\end{equation}
\end{assumption}
As discussed in Section~\ref{sec: intro}, Assumption~\ref{a2} serves as a standard assumption in analyses on stochastic quadratic optimization problems. It is widely adopted in the previous theoretical literature~\citep{bach2013non,ghadimi2012optimal,ghadimi2013optimal,bach2013non,defossez2014constant,dieuleveut2017harder,pan2021eigencurve,pan2023accelerated}, and is considered consistent with practice~\citep{sagun2017empirical,zhang2018energy,zhu2018anisotropic,wu2022alignment}.

This paper studies SGD under the high-dimensional regime, where $d$ is reasonably large. Thus, benign generalization implies a convergence rate that is independent of $d$.
We propose to prescribe function classes based on the Hessian eigenvalue and optimum coordinates decay rate. 
The assumptions are given by:
\begin{itemize}
    \item Eigenvalue decay rate: there exists constants $c_1>0$, $c_2>0$ and $a>1$ such that $\lambda_i$ satisfies
    \begin{equation}\nonumber
        c_1i^{-a}\le \lambda_i\le c_2i^{-a},\ \forall i=1,\dots,d.
    \end{equation}
    \item  Source  condition: there exists constants $c_3>0$ and $b>1 $ such that the optimum of the quadratic objective $ f(\mathbf{w})$ satisfies
    \begin{equation}\nonumber
        \lambda _i\left ( \mathbf{w}_*^{\left ( i \right ) } \right )^2\le c_3 i^{-b },\ \forall i=1,\dots,d.
    \end{equation}
\end{itemize}
\begin{remark}
The polynomially   decaying Hessian eigenvalues and source  condition are commonly used in nonparametric regression~\citep{caponnetto2007optimal,dieuleveut2016nonparametric,raskutti2014early,zhang2023optimality}.  
For nonparametric regression within the reproducing kernel hilbert space~(RKHS) framework~\citep{caponnetto2007optimal}, the polynomially decaying Hessian eigenvalues correspond to those of the covariance operator and the source condition relates to the smoothness of the regression function $f_*$. 
Typically, the regression function $f_*$  belongs to an interpolation space $\left [ \mathcal{H}  \right ] ^s$ of the RKHS $\mathcal{H}$ for some $s>0$. 
Specifically, let the covariance operator $\Sigma :\mathcal{H} \to \mathcal{H}$  be denoted by $\Sigma\phi_i=\mu_i  \phi_i$. Then, for $s\ge 0$ the interpolation space $\left [ \mathcal{H}  \right ] ^s$ is defined as 
\begin{equation}\nonumber
    \left [ \mathcal{H}  \right ] ^s=\left \{ \sum _{i=1}^{+  \infty }a_i\lambda _i^{\frac{s}{2} }e_i  \Big| \sum _{i=1}^{+  \infty }\left ( a_i \right )^2<+\infty    \right \}. 
\end{equation}
 The well-specified setting refers to $f_*\in \left [ \mathcal{H}  \right ] ^s$ with $s\ge1$ and the misspecified setting refers to $f_*\in \left [ \mathcal{H}  \right ] ^s$ with $0<s<1$.
A function satisfies the polynomially decaying Hessian eigenvalues and the source condition belonging to $\left [ \mathcal{H}  \right ]^{s}$ with $s<\frac{b-1}{a} $.
The setting $b> a+1$ can be roughly regarded as the well-specified setting and the setting $b\le a+1$ turns to the misspecified one.
\end{remark}

\subsection{Algorithm}
We focus on a unified framework of SGD, specifically integrating the momentum acceleration and exponentially decaying step size schedule.
We employ the Heavy Ball method \citep{polyak1964some} to solve problem \eqref{eq: objective}, which is defined by an iteration with two parameters $\left (\mathbf{w}_{t}, \mathbf{u}_{t} \right ) $ satisfying for $t\ge 0$,
\begin{equation}\nonumber
    \begin{aligned}
\mathbf{u}_{t+1}=\beta\mathbf{u}_{t}+\eta_t\widehat{\nabla } f\left ( \mathbf{w}_t  \right ) ,\quad
\mathbf{w}_{t+1}=\mathbf{w}_{t}-\mathbf{u}_{t+1},
    \end{aligned}
\end{equation}
starting from the initial point $\mathbf{w}_0$ and $\mathbf{u}_0=\mathbf{0}$.

We adopt the exponentially decaying step size schedule as described in~\citet{ge2019step, wu2022last,pan2023accelerated}.
The step size is piecewise constant and is decreased by a factor after each stage, given by
\begin{equation}\label{step schedule}
    \begin{aligned}
        \eta_t = \frac{\eta_0}{4^{\ell-1}} \enspace \ \text{if } K(\ell-1) \leq t \leq K  \ell - 1,\enspace \quad \text{for  } 1\leq \ell \leq n,
    \end{aligned}
\end{equation}
where $T$ is the total iteration number, $ n=\mathrm{log}_2T$ is the total number of stages, and $K = \frac{T}{n}$. The algorithm is summarized as Algorithm~\ref{alg}. An alternative analysis for SGD and its variants is based on constant learning rate algorithms and tail-averaged output $\bar{\mathbf{w} } _{s,s+T}=\frac{1}{T} \sum_{t=s}^{s+T-1}\mathbf{w} _t$~\citep{dieuleveut2016nonparametric,Jain2016ParallelizingSG,li2023risk}. The tail average method is more amenable to theoretical analysis, but learning rate decay aligns better with practical scenarios.

One notable nuance distinguishing the schedule in Algorithm~\ref{alg} from previous ones \citep{ge2019step,wu2022last} is the step size decaying to $\frac{1}{T^2}$, compared to $\frac{1}{T}$ in prior settings, at the final stage. This adjustment is necessitated by the momentum augmenting the accumulated noise. The resultant  smaller final-stage step size in Algorithm~\ref{alg} preserves convergence while reducing  gradient noise in the final stage.

\begin{algorithm}[t]
    \caption{Stochastic Heavy Ball (SHB) with exponentially decaying step size }\label{alg}
    \begin{algorithmic}
    \REQUIRE {Initial point $\mathbf{w}_0$, $\mathbf{u}_0=\mathbf{0}$, initial learning rate $\eta_0$, momentum $\beta$ }
    \STATE $t\gets 0$
   \FOR{$\ell=1,2,\cdots,\mathrm{log}_2T$}
\STATE$\eta_{\ell} \gets \frac{\eta_0}{4^{\ell-1}}$
\FOR{$i=1,2,\cdots \frac{T}{\mathrm{log}_2T}$}
\STATE $\mathbf{u}_{t+1}\gets\beta\mathbf{u}_{t}+\eta_t\widehat{\nabla } f\left ( \mathbf{w}_t  \right ) $
\STATE $\mathbf{w}_{t+1}\gets\mathbf{w}_{t}-\mathbf{u}_{t+1}$
\STATE $t\gets t+1$
\ENDFOR
\ENDFOR

   \ENSURE $\mathbf{w}_t$

      \end{algorithmic}
\end{algorithm}

In Algorithm~\ref{alg}, $\beta $ represents the magnitude of the momentum. When $\beta=0$, Algorithm~\ref{alg} reduces to the vanilla SGD algorithm
\begin{equation}\nonumber
    \mathbf{w}_{t+1} = \mathbf{w}_t-\eta_t \widehat{\nabla} f\left ( \mathbf{w}_t \right ).
\end{equation}

 The iterations of Algorithm~\ref{alg} can be represented by a linear recursion with injected noise
\begin{equation}\nonumber
    \tilde{\mathbf{w}} _t =\mathbf{A}_{t-1} \tilde{\mathbf{w}} _{t-1}+\eta_{t-1} \tilde{\bm{\zeta }} _{t-1},
\end{equation}
and initial $\tilde{\mathbf{w}}_0$, where
\begin{equation}\label{eq:ww}
    \begin{aligned}
       \tilde{\mathbf{w}} _t =\begin{bmatrix}\mathbf{w}_{t}-\mathbf{w}_* 
 \\\mathbf{w}_{t-1}-\mathbf{w}_* 
\end{bmatrix},\ \tilde{\bm{\zeta }} _t =\begin{bmatrix}\bm{\zeta }_{t}
 \\ \mathbf{0}
\end{bmatrix},\ \tilde{\mathbf{w}} _0 =\begin{bmatrix}\mathbf{w}_{0}-\mathbf{w}_* 
 \\\mathbf{w}_{-1}-\mathbf{w}_* 
\end{bmatrix}, \mathbf{A}_t=\begin{bmatrix}
\left ( 1+\beta  \right ) \mathbf{I}-\eta_t\mathbf{H}    & -\beta \mathbf{I} \\
\mathbf{ I}  & \mathbf{O
} \end{bmatrix}.
    \end{aligned}
\end{equation}
While the corresponding $\mathbf{A}_t$ matrices for different $\eta_t$ are non-commutative, their cumulative product $\Pi_{t = t_1}^{t_2}\mathbf{A}_t=\mathbf{A}_{t_2} \cdots \mathbf{A}_{t_1}$ can be decomposed into $d$ blocks of $2\times2$ matrices as follows
\begin{align*}
    \Pi_{t = t_1}^{t_2}\mathbf{A}_t = \mathbf{P} \left[
    \begin{array}{ccc}
        \Pi_{t = t_1}^{t_2}\mathbf{A}_t^{\left ( 1 \right )} & & \\
         & \ddots & \\
         & & \Pi_{t = t_1}^{t_2}\mathbf{A}_t^{\left ( d \right )}
    \end{array}
    \right] \mathbf{P}^{\top}, \enspace \text{where} \ \     \mathbf{A}_t^{\left ( j \right )}=\begin{bmatrix}
1+\beta -\eta _t\lambda _j  & -\beta \\
 1 &0
\end{bmatrix} \in \mathbb{R} ^{2\times 2},
\end{align*}
for some orthogonal matrix $\mathbf{P}$. Consequently, the analysis of Algorithm~\ref{alg} relies on the characteristics of $\mathbf{A}_{t_2}^{\left ( j \right )} \cdots \mathbf{A}_{t_1}^{\left ( j \right )}$.  
Note that directly bounding $\mathbf{A}^{\left ( j \right )}_{t_2}\cdots \mathbf{A}^{\left ( j \right )}_{t_1}$ by the spectral radius of $\mathbf{A}^{\left ( j \right )}_{t_2}$ would fail to mitigate the expansion caused by the product of $\mathbf{A}_t$ in certain settings. We will provide a  fine-grained  estimate of the product of the non-commutative matrices, especially for  the space of small eigenvalues.

\section{Main Results}
In this section, we summarize our theoretical results.  Section~\ref{sec: result-upb} presents the upper bound on the last-iterate error of Algorithm~\ref{alg}. Section~\ref{sec: result-lwb} formalizes  our information-theoretic lower bound. In Section~\ref{sec: result-refinedana}, the analysis is restricted to the classical function class, offering a detailed discussion on the algorithm's optimality conditions.

\subsection{Upper Bound}\label{sec: result-upb}

For the upper bound on the last-iterate error of Algorithm~\ref{alg}, we have the following theorem.
\begin{thm}\label{SHB UP}
   Consider applying Algorithm~\ref{alg} to problem \eqref{eq: objective}. Under Assumption~\ref{a2}, with the initial point $\mathbf{w}_0=\mathbf{0}$, step size $\eta_0\le \frac{1}{\lambda_1}$, total iteration number $T\ge 16$, and momentum $\beta \in \left [ 0,\left ( 1-\frac{A}{T}  \right )^2  \right] $ where $A=256\mathrm{log}_2  T\cdot\mathrm{ln}T $, the error of the output can be bounded from above by 
   \begin{equation}\nonumber
    \begin{aligned}
      \mathbb{E} \left [ f\left ( \mathbf{w} _T \right )  \right ] -f\left ( \mathbf{w }_*  \right ) \le &
    \underbrace{ C_1\left ( \sum_{j=1}^{k^*}\frac{\left ( \mathrm{log}_2 T \right ) ^2A\left ( 1-\sqrt[]{\beta }  \right ) }{\eta_0T}\left ( \mathbf{w} _*^{\left ( j \right ) } \right ) ^2+\sum_{j=k^*+1}^{d}\lambda _j\left ( \mathrm{log}_2 T \right ) ^2\left ( \mathbf{w} _*^{\left ( j \right ) } \right ) ^2 \right ) }_{\mathrm{Bias}}\\
    & + \underbrace{C_2\sigma ^2 \left ( \frac{\left ( \lambda _1^2\eta_0^2+A^2 \right )k^*}{T} +\eta_0^2 \sum_{j=k^*+1}^{d}\lambda _j^2T\frac{1}{\left ( 1-\sqrt[]{\beta }  \right )^2} \right ) }_{\mathrm{Variance}},
        \end{aligned}
    \end{equation}
    with expectation taken over $\mathbb{P}_{\bm{\xi }}$, 
    %universal 
    constant $C_1,C_2>0$, and $k^*=\mathrm{max}\left \{ k:\lambda _k\eta_0\ge \frac{ 1-\sqrt{\beta}  }{T}  \right \} $.

\end{thm}
\begin{remark}
    The threshold $k^*$, referred to as the effective dimension, is used to characterize the generalization ability of SGD with different levels of momentum in the high-dimensional setting. When the eigenvalues of the Hessian decay rapidly, $k^*$ is much smaller than the problem dimension $d$. 
 A related concept of effective dimension has appeared in previous works~\citep{Bartlett_2020,li2023risk}, but our $k^*$ is specifically adapted to different levels of momentum. It can be observed that increasing the momentum enlarges $k^*$, which accelerates the convergence of the bias term but also increases variance.
\end{remark}
Theorem~\ref{SHB UP} bounds the last-iterate error of Algorithm~\ref{alg} with respect to any magnitude of the added momentum $\beta$. The theorem indicates that while increasing the momentum $\beta$ accelerates the convergence of the bias term, it also, unfortunately, amplifies the accumulation of noise. This observation parallels the tail-averaging results in \citet{li2023risk}. 

Theorem~\ref{SHB UP} addresses SHB with the close-to-reality exponential decaying step size, which poses technical challenges when evaluating non-commutative diminishing matrices induced by decaying step size.
Here, the main technical contribution of our proof is to provide a fine-grained  evaluation for the space of small eigenvalues. For the small eigenvalues, the spectral radius of momentum matrix $ \rho \left ( \mathbf{A}_{0}^{\left ( j \right )} \right )$ is close to 1, prevents its powers from yielding exponential contraction. 
Directly bounding $\lambda _j\left ( \mathbf{A}_{T-1}^{\left ( j \right )}\cdots \mathbf{A}_{0}^{\left ( j \right )} \mathbf{1}_2 \right ) ^2_1$ using the norm $\left \| \mathbf{A}_{T-1}^{\left ( j \right )}\cdots \mathbf{A}_{0}^{\left ( j \right )} \right \| ^2$ 
leads to a divergent result. Instead, we analyze the decomposition of $\left ( \mathbf{A}_{sK }^{\left ( j \right )} \right )^{K}\mathbf{1}_2$ over $\mathbf{1}_2$ and $\mathbf{e}_2$, and show that $\left | \left ( \mathbf{A}_{T-1}^{\left ( j \right )}\cdots\mathbf{A}_{0}^{\left ( j \right )} \mathbf{1}_2 \right ) _1 \right |$ is at most on the order of $\mathrm{log}_2T$.

The bias-variance trade-off in Theorem~\ref{SHB UP} leads to the considerations of the hyperparameter selection and whether these strategies succeed to fully exploit the information from the oracles.

\subsection{Information-theoretic Lower Bound}\label{sec: result-lwb}
How the attained upper bound compares to the optimal rate is crucial for assessing its effectiveness and understanding the implicit bias of SGD with its variants. For the asymptotic large $T$ regime, the classical analysis indicates a gradient lower bound of $\widetilde{\mathcal{O}}\left(d\sigma^2/T\right)$ for problem~\eqref{eq: objective}~\citep{dieuleveut2016nonparametric,Jain2016ParallelizingSG,ge2019step,pan2023accelerated}. This section presents an information-theoretic lower bound for \eqref{eq: objective} in the overparameterized regime without assuming a specific Hessian eigenvalue decay rate.  The lower bound follows the min-max error framework~\citep{yang1999information} and is established through the dimension-wise binary hypothesis testing under the informational constrained risk inequality~\citep{yu1997assouad, brown1996constrained}.

We study general algorithms that can be potentially randomized and are not confined to  using only stochastic gradient information of the data. 
The considered algorithm class $\mathcal{A}_T$ includes all algorithms $\mathrm{A}_T$ that for iteration $1\le t \le T$, with the observation of   $t$ independent data points $\left \{ \left ( \mathbf{H} \left ( \bm{\xi }_t \right ) ,\mathbf{b} \left ( \bm{\xi }_t  \right )  \right ) \right \} _{t=0}^{t-1}$ %and finally 
, return an estimate $\hat{\mathbf{w} } _T$ of the corresponding optimum $\mathbf{w}_*$. Formally,  $\mathcal{A}_T$ includes all algorithm  $\mathrm{A}_T$ with the update going as 
\begin{align}\label{At1}
    \hat{\mathbf{w} }_t = \mathrm{M}_t\big ( \bm{\bar{\zeta}}_{t},\left \{ \left ( \mathbf{H} \left ( \bm{\xi }_i \right ) ,\mathbf{b} \left ( \bm{\xi }_i  \right )  \right ) \right \} _{i=0}^{t-1}, \mathbf{w}_0, \hat{\mathbf{w} }_1,\dots, \hat{\mathbf{w} }_{t-1} \big ),\quad   t=1,2,\dots, T,  
\end{align}
where $\mathrm{M}_t$ is an arbitrary measurable mapping from $\mathbb{R}  ^{d\times \left (d+T+2  \right ) }$ to $\mathbb{R}^d$ and $\bm{\bar \zeta }_{t}$ is a random variables independent of $\left \{ \left ( \mathbf{H} \left ( \bm{\xi }_i \right ) ,\mathbf{b} \left ( \bm{\xi }_i  \right )  \right ) \right \} _{i=0}^{T-1}$ encoding the randomness in algorithm $\mathrm{A}_T$.

The problem is characterized by the pairs $\left( F(\cdot,\bm\xi), \mathbb{P}_{\bm\xi} \right)$: the individual function $F(\mathbf{w},\bm\xi)$, as defined in \eqref{eq: objective}, contains the full information of the observation $\left(\mathbf{H}(\bm\xi), \mathbf{b}(\bm\xi) \right)$; the distribution of these functions is governed by $\mathbb{P}_{\bm\xi}$, which collectively determines $f(\mathbf{w}) = \mathbb{E}_{\bm\xi}[F(\mathbf{w}, \bm\xi)]$. Obtaining dimension-independent lower bounds requires further priors on the function class. We consider the class $\mathcal{F}_{\bar{\mathbf{w}},\mathbf{H}}$ that has a prior constraint on the optimum $\mathbf{w}_*$ for each coordinate and any instance of Hessian $\mathbf{H}$. Recall that the notation $\mathbf{w}_*$ denotes the optimum of $f(\mathbf{w}) = \mathbb{E}_{\bm\xi}[F(\mathbf{w}, \bm\xi)]$.  Given $\bar{\mathbf{w} }=\sum_{i=1}^{d} \bar{w}^{\left ( i \right ) }\mathbf{v}_i$ with $\bar{w}^{(i)}>0$, $\mathcal{F}_{\bar{\mathbf{w}},\mathbf{H}}$ is defined as 
\begin{equation}\label{wi}
  \mathcal{F}_{\bar{\mathbf{w}},\mathbf{H}}= \left \{ \left ( F\left ( \cdot ,\bm{\xi} \right ),\mathbb{P}_{\bm{\xi }} \right ): \nabla^2 \left[\mathbb{E}_{\bm\xi} F(\cdot,\bm\xi)\right] = \mathbf{H},\  \left | \mathbf{w}_*  ^{\left ( i \right ) }\right |\le \bar{w}^{\left ( i \right ) },\ \forall\  1\le i\le d  \right \}.
\end{equation} 
Meanwhile,  meeting Assumption~\ref{a2} requires the problem to belong to the class 
\begin{align*}
    \mathcal{F}_0 = \left\{ \left ( F\left ( \cdot ,\bm{\xi} \right ),\mathbb{P}_{\bm{\xi }} \right ): \mathrm{Var}_{\bm\xi\sim\mathcal{P}}\left(\nabla F(\cdot, \bm\xi)\right) \preceq \sigma^2\mathbb{E}\mathbf{H}(\bm\xi) \right\}.
\end{align*}

 Given an algorithm $\mathrm{A} _T\in \mathcal{A} _T$ and a pair $\left ( F\left ( \cdot ,\bm{\xi} \right ),\mathbb{P}_{\bm{\xi }} \right )\in \mathcal{F}_{\bar{\mathbf{w} },\mathbf{H}}\cap\mathcal{F}_0$, the error of the algorithm's output is $f\left (\hat{\mathbf{w} } _T  \right )  -\inf_{\mathbf{w}} f\left (\mathbf{w}  \right )=\mathbb{E}_{\bm{\xi} }\left [ F\left ( \hat{\mathbf{w} } _T,\bm{\xi} \right )  \right ]-\inf_{\mathbf{w} }\mathbb{E}_{\bm{\xi} }\left [ F\left ( \mathbf{w},\bm{\xi} \right )  \right ]$. Consequently, the min-max error of algorithm class $\mathcal{A} _T $ on function class $\mathcal{F}_{\bar{\mathbf{w} },\mathbf{H}}\cap\mathcal{F}_0$ is defined as
\begin{equation}\nonumber
    \inf_{\mathrm{A}_{T} \in\mathcal{A} _{T}}\sup_{\left ( F\left ( \cdot ,\bm{\xi} \right ),\mathbb{P}_{\bm{\xi }} \right )\in \mathcal{F}_{\bar{\mathbf{w} },\mathbf{H}}\cap\mathcal{F}_0 }\left ( \mathbb{E}\left [f\left (\hat{\mathbf{w} }_T  \right )   \right ]-\inf_{\mathbf{w}} f\left (\mathbf{w}  \right ) \right )  ,
\end{equation}
where the expectation is taken over the observations  $\left \{ \left ( \mathbf{H} \left ( \bm{\xi }_t \right ) ,\mathbf{b} \left ( \bm{\xi }_t  \right )  \right ) \right \} _{t=0}^{T-1} $ and any additional randomness in $\mathrm{A}_{T}$. The min-max error measures the performance of the best algorithm $\mathrm{A}_{T} \in\mathcal{A} _{T}$ where the performance is required to the uniformly good over the function class $\mathcal{F}_{\bar{\mathbf{w}},\mathbf{H}}\cap\mathcal{F}_0$.

We present our lower bound as below.
\begin{thm}\label{Low}
    For any $T\in\mathbb{N}^+$ and positive semidefinite matrix $\mathbf{H}$, the min-max error of algorithm class $\mathcal{A}_T$ on problem class $\mathcal{F}_{\bar{\mathbf{w} },\mathbf{H}}\cap\mathcal{F}_0$ is bounded from below by
\begin{equation}\nonumber
     \inf_{\mathrm{A}_{T} \in\mathcal{A} _{T}}\sup_{\left ( F\left ( \cdot ,\bm{\xi} \right ),\mathbb{P}_{\bm{\xi }} \right )\in \mathcal{F}_{\bar{\mathbf{w} },\mathbf{H}} \cap \mathcal{F}_0 }\left ( \mathbb{E}\left [f\left (\hat{\mathbf{w} }_T  \right )   \right ]-\inf_{\mathbf{w}} f\left (\mathbf{w}   \right ) \right)  \ge \frac{1}{8} \left ( \frac{\left |  \mathcal{I} \right |  \sigma^2}{T}+\sum_{i\notin \mathcal{I}}^d\lambda _i\left ( \bar{ w} ^{\left ( i \right ) }  \right ) ^2  \right ) ,
\end{equation}
given $T$ independent observations $\left \{ \left ( \mathbf{H} \left ( \bm{\xi }_t \right ) ,\mathbf{b} \left ( \bm{\xi }_t  \right )  \right ) \right \} _{t=0}^{T-1}$ and any additional randomness in $\mathrm{A}_{T}$, where $\mathcal{I}=\left \{ i\in\mathbb{N}^+:\   \lambda _i\left ( \bar{ w} ^{\left ( i \right ) }  \right ) ^2\ge \frac{\sigma ^2}{T}\right \} $.
\end{thm}
Theorem~\ref{Low} establishes the information-theoretic lower bound for the function $f$, constrained within $\mathcal{F}_{\bar{\mathbf{w} },\mathbf{H}}$ and satisfying Assumption~\ref{a2}.  This information-theoretic lower bound holds for any instance of Hessian and, as a result,  does not explicitly depend on the dimension which allows for a detailed discussion discuss when (accelerated) SGD can achieve optimality. A comprehensive analysis is presented in the following subsection.

\subsection{Optimality Analysis of (Accelerated) SGD}\label{sec: result-refinedana}
In this section, we discuss when (accelerated) SGD can achieve optimality for function classes that have polynomially decaying Hessian spectrum and satisfy the source condition in Section~\ref{sec:decay}, or for those whose optimum is subject to $\ell_{\infty}$ constraints in Section~\ref{sec: l infty}. 

\subsubsection{OPTIMALITY OF (ACCELERATED) SGD UNDER POLYNOMIALLY DECAY HESSIAN SPECTRUM AND SOURCE CONDITION} \label{sec:decay}

For the function class satisfying polynomially decaying Hessian eigenvalues and the source condition, a tight information-theoretic lower bound can be obtained by setting $\bar{ w} ^{\left ( i \right ) }$ in Theorem~\ref{Low} as $i^{\frac{a-b}{2}}$, yielding a min-max lower bound of
\begin{equation}\label{eq:low432}
           \inf_{\mathrm{A}_{T} \in\mathcal{A} _{T}}\sup_{\left ( F\left ( \cdot ,\bm{\xi} \right ),\mathbb{P}_{\bm{\xi }} \right )\in\mathcal{F}_{\bar{\mathbf{w} },\mathbf{H}} \cap \mathcal{F}_0  }\left ( \mathbb{E}\left [f\left (\hat{\mathbf{w} }_T  \right )   \right ]-\inf_{\mathbf{w}} f\left (\mathbf{w}   \right ) \right ) \gtrsim T^{-1+\frac{1}{b} }\sigma^{2-\frac{2}{b} },
\end{equation}
as per Theorem~\ref{Low}.

% We first provide the information-theoretic lower bound under eigenvalues decay polynomially and smoothness condition. Set $\overline{ w} ^{\left ( i \right ) }$ in~\eqref{wi} to be $i^{\frac{a-b}{2}}$. By Theorem~\ref{Low}, the lower bound becomes 

% \myred{Next, we will explore under what parameter choices Algorithm~\ref{alg} can achieve optimality based on the relationship between $a$ and $b$. If $T\le \sigma^2$, then the rate in the right hand in \eqref{eq:low432} is divergent. Therefore we consider the setting $T> \sigma^2$.}

% Regarding the upper bound, in accordance with the decaying eigenvalue and $\mathbf{w}_*$ coordinate assumptions, the following theorem presents the optimally tuned parameters and the resulting error.

On the other hand, by optimally tuning parameters according to the decay rate, the following upper bound can be derived from Theorem~\ref{SHB UP}.
\begin{thm}\label{thm: opt rigem} 
When the observation noise $\sigma^2$ is regarded as constant, for the region of $b<a\le 2b$, set the parameter in Algorithm~\ref{alg} as
    \begin{equation}\label{parmeter:SHB}
        \beta=\left ( 1- A\cdot T^{-\frac{a-b}{b}}  \right ) ^2, \, \qquad \eta_0=\left ( 2\lambda_1 \right ) ^{-1}, \qquad \mathbf{w}_0=\mathbf{0},
    \end{equation}
    where $0<\frac{a-b}{b} \leq 1$ and $\text{A} = 256\mathrm{log}_2 T\cdot \mathrm{ln}T$. The error of the output from Algorithm~\ref{alg} is bounded from above by 
    \begin{equation}\nonumber
    \begin{aligned}
         \mathbb{E} \left [ f\left ( \hat{\mathbf{w} }_T^{SHB} \right )  \right ]-\inf_{\mathbf{w}}f\left (\mathbf{w}   \right )
\lesssim  \underbrace{\left ( \mathrm{log}_2T \right ) ^{6  
}\cdot T^{-1+\frac{1}{b}   }}_{\mathrm{Bias} }  
+ \underbrace{\sigma^{2 }\left ( \mathrm{log}_2T \right ) ^8\cdot  T^{-1+\frac{1}{b}   }} _{\mathrm{Variance} }.
    \end{aligned}
\end{equation}
    For the region of $a\le b$, set the parameter in Algorithm~\ref{alg} as
    \begin{equation}\label{parmeter:SGD}
        \beta = 0, \qquad \eta_0=\frac{1}{2\lambda_1}T^{-1+\frac{a}{b} }\le \frac{1}{2\lambda_1},\qquad \mathbf{w}_0=\mathbf{0}.
     \end{equation}
     The error of the output from Algorithm~\ref{alg} is bounded from above by 
    \begin{equation}\nonumber
    \begin{aligned}
         \mathbb{E} \left [ f\left ( \hat{\mathbf{w} }_T^{SGD} \right )  \right ]-\inf_{\mathbf{w}}f\left (\mathbf{w}   \right )
\lesssim  &\underbrace{\left ( \mathrm{log}_2T\right )^{\frac{2\left ( b-1 \right ) }{a} }\cdot T^{-1+\frac{1}{b} }}_{\mathrm{Bias} }  
+ \underbrace{\sigma^2\left ( \mathrm{log}_2T\right )^4 \cdot T^{-1+\frac{1}{b} }} _{\mathrm{Variance} }.
    \end{aligned}
\end{equation}
\end{thm}
When ignoring the polylogarithmic factors and regarding the noise $\sigma^2$ as constant,
comparing Theorem~\ref{thm: opt rigem} and the corresponding optimal rate \eqref{eq:low432} identifies SGD optimality region of $1<a\le b$ and SHB optimality region of $1<b<a\le 2b$. 

We highlight that SGD achieves optimality across the entire region $1<a\le b$, without suffering from the saturation effect discussed in \citet{dieuleveut2016nonparametric} and \citet{dieuleveut2017harder}, which suggests that SGD can only reach optimality in the region $a\le b\le 2a+1$. This improvement is due to our more involved estimation of the bias term, where we provide the convergence rate of the bias term for each coordinate. 
In fact, for coordinates up to the effective dimension $k^*$ defined in \eqref{SHB UP} divided by a  polylogarithmic factor, the bias term converges exponentially, while the remaining coordinates account for the error.
For any decay rate of the true parameter $\mathbf{w}_*$, SGD with  parameters defined in \eqref{parmeter:SGD} adaptively selects a threshold and achieves optimality. The saturation effect has also been widely observed in kernel ridge regression (KRR). \citet{li2023on} demonstrated that KRR does indeed suffer from this effect. This suggests that when the true target parameter is sufficiently smooth, SGD surpasses KRR in performance. 
In the limit case where the regularization in KRR approaches zero, the solution obtained by GD for such ridgeless regression is the so-called minimum-norm solution \citep{gunasekar2017implicit}. It is shown in Theorem~2 from ~\citep{Bartlett_2020} that  this solution leads to a divergent error when the Hessian eigenvalues decay polynomially. This suggests that SGD consistently outperforms the minimum-norm solution in this setting.

In the region $b<a<2b$, SHB with parameters defined in \eqref{parmeter:SHB} can achieve optimality. In this region, the true parameter is less smooth, resulting in a large initial bias. Introducing appropriate momentum can accelerate the convergence of the bias term and balance the increase in variance caused by the momentum, ultimately achieving optimality. This is analogous to the conclusion in \citet{dieuleveut2017harder}, while \citet{dieuleveut2017harder} examines the error of SHB with a regularization term, constant step size, and tail-averaging, the additional regularization makes it difficult to determine whether the improvement stems from the regularization or the momentum. Whereas,  we directly analyze the last-iterate error of SHB with exponentially decaying step size.

\begin{wrapfigure}{l}{3.7in}
\centering
\resizebox{0.8\linewidth}{!}{
      \begin{tikzpicture}
    % Axes
    \draw[->] (0,0) -- (6,0) node[right] {$b$};
    \draw[->] (0,0) -- (0,6) node[above] {$a$};

    % Lines
    \draw[thick] (0,0) -- (4.5,4.5) node[above right] {$a = b$};
    \draw[thick] (0,0) -- (3,6) node[above right] {$a = 2b$};
    \draw[dashed] (2,0) -- (4.5,1.5) node[right, xshift=25pt, yshift = 20pt] {$2a+1 = b$};

    % Labels and shading
    \fill[red!20] (0,0) -- (3,6) -- (0,6) -- cycle;
    \fill[green!20] (0,0) -- (3,6) -- (4.5,4.5) -- cycle;
    \fill[blue!20] (0,0) -- (4.5,4.5) -- (75/14,27/14) -- (1.5,0) -- cycle;
    \fill[blue!10] (1.5,0) -- (75/14,27/14)  -- (6,0) -- cycle;

    % Add thick border for the blue region
    \draw[line width=1mm, orange, opacity=0.5] (0,0) -- (4.5,4.5) -- (75/14,27/14) -- (1.5,0) -- cycle;
    \draw[line width=1mm, orange, opacity=0.5] (5.2,2.4) -- (5.5,3.1);
    \node at (6.4,3.8) {\textcolor{blue}{SGD is optimal}};
\node at (6.5, 3.3) {\textcolor{blue}{in \citet{dieuleveut2016nonparametric}}};

    \draw[dashed] (1.5,0) -- (1.5,1.5) -- (0,1.5);

    % Node placements
    \node at (1.2, 5) {$a > 2b$};
    \node at (1.2, 4.5) {\small SHB is not};
    \node at (1, 4) {\small optimal};

    \node at (3.4, 4.6) { $b<a<2b$};
    \node at (3, 4.1) {\small SHB is};
    \node at (2.7, 3.6) {\small optimal};

    \node at (3.7, 1.7) {\small $a\le b$};
    \node at (3.7, 1.2) {\small SGD is optimal};

    \node at (-0.3,1.5) {$1$};
    \node at (1.5,-0.3) {$1$};
\end{tikzpicture}
}
  \caption{Optimality of SGD with constant $\sigma^2$.   }
  \label{fig: opt contsigma}
\end{wrapfigure}
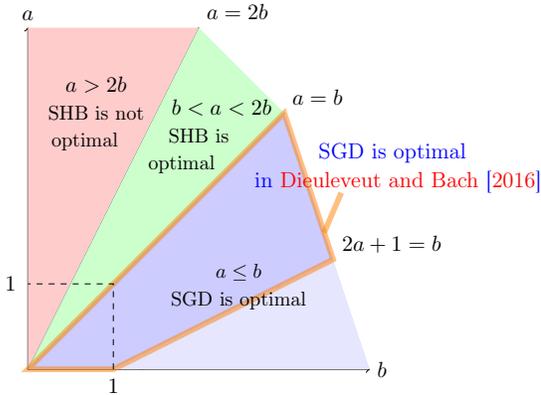

Typically, the observation noise $\sigma^2$ is regarded as a constant~\citep{dieuleveut2016nonparametric}. In such scenario, vanilla SGD with smaller initial step size achieves optimality when $a\leq b$, whereas \citet{dieuleveut2016nonparametric} only shows SGD can reach optimality when $a\leq b\leq 2a+1$ (as shown in the orange-boxed region of Figure~\ref{fig: opt contsigma}). As for the highly misspecified setting when $b<a\le2b$, the result indicates that SHB can still attain the optimal rate under the momentum value of $1-\beta = \Theta \left( T^{-\frac{a-b}{b}} \right)$, as illustrated in Figure~\ref{fig: opt contsigma}.  \ \ \ \ \ \ \ \ \ \ \ \ \ \ \ \ \ \ \ \ \ \ \ \ \ \ \ \ \ \ \ \ \ \ \ \ \ \ \ \ \ \ \ \ \ \ \ \ \ \ \ \ \ \ \ \ \ \ \ \ \ \ \ \ \ \ \ \ \ \ \ \ \ \ \ \ \ \ \ \ \ \ \ \ \ \ \ \ \ \ \ \ \ \ \ \ \ \ \ \ \ \ \ \ \ \ \ \ \ \ \ \ \ \ \ \ \ \ \ \ \ \ \quad \quad \quad \quad \quad \quad \quad \quad  \ \ \ \ \ \ \ \ \ \ \ \ \ \ \ \ \ \ \ \ \ \ \ \ \quad \quad \quad \quad \quad \quad \quad \quad

 On the other hand, when $a>2b$,  neither SGD and SHB can achieve optimality. We suspect that in this regime, greater computational resources, such as multi-pass SGD \citep{pillaud2018statistical} or principal component analysis (PCA) to reduce dimension \citep{silin2022canonical}, are required. We leave the deep investigation as a future work. 
When the problem transitions into the small noise regime such as $\sigma^2=0$, The primary challenge shifts from the convergence of the variance term to the convergence of the bias term.  Intuitively, the main challenge in this regime turns to optimization and its analysis is out of the  scope of our work.

% Theorem~\ref{thm: opt rigem} demonstrates that  Algorithm~\ref{alg} can achieve optimality when $T^{\frac{a-2b}{a}}< \sigma^2<T$. When $\sigma^2<T^{\frac{a-2b}{a}} $, the $\mathrm{bias}$ term becomes the leading error, preventing Algorithm~\ref{alg} from reaching optimality. 

% When considering the common scenario where $\sigma^2$ is at a constant level,  we derive the following conclusions from Theorem~\ref{thm: opt rigem}. 

% If $1<a\le b$, the vanilla SGD with initial step size $\eta_0=\frac{1}{2}T^{-1+\frac{a}{b} }\sigma^{-\frac{2a}{b} }<1$ can reach optimality. If $b<a\le2b$, SHB with momentum $\beta=\left ( 1-\frac{A}{T^{s }}  \right ) ^2$, where $0<s=\frac{a-b}{b}-\frac{2a}{b} \mathrm{log}_T\sigma \le 1$, $A=48\mathrm{log}_2 T\cdot \mathrm{ln}T$, and initial step size  $\eta_0=0.5$ can reach optimality. If $b<a\le2b$, Algorithm~\ref{alg} can not reach optimality.

\subsubsection{OPTIMALITY OF VANILLA SGD UNDER INFINITY NORM CONSTRAINT}\label{sec: l infty}

In the previous section, we presented the optimality condition based on the eigenvalue and coordinate decay rates. This section advances to an optimal condition that SGD achieves the optimality for any instance of the Hessian matrix.

 Specifically, consider the function class $\mathcal{F}_{c\mathbf{1},\mathbf{H}}$ with $c>0$ being a constant and $\bm 1$ denoting the all-ones vector. This is equivalent to imposing
 $\ell_{\infty}$ constraints on the optimum $\mathbf{w}_*$ with respect to the standard orthonormal basis dictated by the Hessian eigenspace.

 Setting the initial point $\mathbf{w}_0=\mathbf{u}_0=0$, the step size $\eta_0\le \frac{1}{\lambda_1}$, and the momentum $\beta=0$, Algorithm~\ref{alg} recovers vanilla SGD with exponentially decaying step size. Theorem~\ref{SHB UP} shows that its error can be bounded from above by
\begin{equation}\label{eq: linf up}
    \mathbb{E} \left [ f\left ( \mathbf{w} _T \right )  \right ] -\inf_{\mathbf{w}}f\left ( \mathbf{w }  \right ) \lesssim \left ( 1+\sigma^2 \right )  \left ( \frac{k^*\left ( \mathrm{log}_2T \right ) ^4}{T}+\left ( \mathrm{log}_2T \right ) ^2\sum_{j=k^*+1}^d\lambda_{j}c^2 \right ),
\end{equation}
where $k^*=\mathrm{max}\left \{ k:\lambda _k\ge \frac{1}{T}  \right \} $. Meanwhile, for any Hessian $\mathbf{H}$ under the $l_{\infty}$ constraint, Theorem~\ref{Low} indicates the min-max rate bound 
\begin{equation}\label{eq: linf low}
     \inf_{\mathrm{A}_{T} \in\mathcal{A} _{T}}\sup_{\left ( F\left ( \cdot ,\bm{\xi} \right ),\mathbb{P}_{\bm{\xi }} \right )\in \mathcal{F}_{c\mathbf{1},\mathbf{H}} \cap \mathcal{F}_0 }\left ( \mathbb{E}\left [f\left (\hat{\mathbf{w} }_T  \right )   \right ]-\inf_{\mathbf{w}} f\left (\mathbf{w}   \right ) \right )  \ge \frac{1}{8} \left ( \frac{k^*_1\sigma^2}{T}+\sum_{j=k^*_1+1}^d\lambda _jc ^2 \right ) ,
\end{equation}
where $k^*_1=\max \left \{ k: \lambda _k \ge \frac{\sigma ^2}{Tc^2} \right \} $. When $\sigma^2$ is regarded as constant, comparing \eqref{eq: linf up} and \eqref{eq: linf low} concludes that SGD attains a sharp rate ignoring polylogarithmic factors, leading to the following optimality theorem for SGD.
\begin{thm}\label{thm:SGD l infty}
    For any Hessian matrix $\mathbf{H}$, ignoring polylogarithmic factors, if the optimum $\mathbf{w}_*$ has $\ell_{\infty}$ constraints in the Hessian eigenspace, and $\sigma^2$ is regarded as constant, vanilla SGD with a decaying step size attains the min-max optimal rate.

    % The vanilla SGD with decaying step size can reach optimality when the optimum of $f$ has $\ell_{\infty}$ constraints with respect to the standard orthonormal basis formed by the eigenvectors of $\mathbf{H}$, and the noise $\sigma^2$ is at a constant level.
\end{thm}

Theorem~\ref{thm:SGD l infty} demonstrates that under $\ell_{\infty}$ constraint of the optimum, SGD attains the optimal rate for any Hessian matrix $\mathbf{H}$ ignoring polylogarithmic factors. Given the implementation of Algorithm~\ref{alg} relies only on the largest Hessian eigenvalue and is independent of the remaining ones, Algorithm~\ref{alg} attains an instance optimality that adapts to finder eigenvalue structure. 

\noindent \textbf{Comparison with Minimum-norm Solution of Linear Regression} \quad
Since GD for linear regression converges to the minimum-norm solution \citep{gunasekar2017implicit}, we now compare the excess risk of the minimum-norm solution with the output of SGD. Specifically, we analyze the output of Algorithm~\ref{alg} with momentum $\beta=0$, for the optimization problem defined in \eqref{eq: objective}, where $\mathbf{H}\left (  \bm{\xi}\right )= \mathbf{x} \mathbf{x} ^{\top }$ and $\mathbf{b}\left ( \bm{\xi} \right )=\mathbf{x}y$ with $\mathbf{x}$ as the covariate vector and $y$ as the response in linear regression.  As per the excess risk upper bound of the minimum $\ell_2$ norm solution $\mathbf{\hat{w} }_T^{\mathrm{min} }$ given by \citet{Bartlett_2020}, with probability at least $1-\delta$, 
\begin{equation}\nonumber
    \begin{aligned}
        \mathbb{E} f\left ( \mathbf{\hat{w} }_T^{\mathrm{min} }  \right ) -f\left ( \mathbf{w}_*  \right ) \le &c_4\left ( \left \| \mathbf{w}_*  \right \| ^2\lambda _1\mathrm{max}\left \{ \sqrt{\frac{\sum _{i>0}\lambda _i}{T\lambda _1} },\frac{\sum _{i>0}\lambda _i}{T\lambda _1},\sqrt{\frac{\mathrm{log\left ( 1/\delta  \right ) } }{T} }    \right \}   \right ) \\
        &+c_5\mathrm{log}\left ( 1/\delta  \right ) \sigma ^2\left ( \frac{k^*}{T} +\frac{T\sum _{i>k^*}\lambda _i^2}{\left ( \sum _{i>k^*}\lambda _i \right )^2 }  \right ) ,
    \end{aligned}
\end{equation}
where $c_4,c_5>1$ are constants, and $k^*=\mathrm{min}\left \{ k\ge 0:\frac{\sum_{i>k}\lambda _i}{\lambda _{k+1}} \ge c_6T \right \}$ for some constant $c_6>1$. The upper bound in \citet{Bartlett_2020} includes the term $\left \| \mathbf{w}_*  \right \| ^2$, which becomes $dc^2$ when $\left \| \mathbf{w}_*  \right \| _{\infty}=c$ indicating a dependence on the problem's dimension. In contrast, our upper bound in~\eqref{eq: linf up}, under the condition $\left \| \mathbf{w}_*  \right \| _{\infty}=c$, can be independent of $d$ and achieve optimality for a range of eigenvalues $\lambda_i$,  such as $\tilde{\mathcal{O}}  \left ( T^{-1+1/a} \right ) $ when $\lambda_i=i^{-a}$, or , $\tilde{\mathcal{O}}  \left ( T^{-1} \right ) $ when $\lambda_i=e^{-i}$, for $1\le i\le d$, and $a>1$.

\section{Proof of  Main Theorems}
In this section, we provide the proofs of Theorem~\ref{SHB UP} and Theorem~\ref{Low}. The detailed proofs for all related lemmas are provided in the Appendix.

\subsection{Proof of Theorem~\ref{SHB UP}}
We first introduce the iteration of $\begin{bmatrix}
\mathbf{w}_{t}-\mathbf{w}_* ,  & \mathbf{w}_{t-1}-\mathbf{w}_*
\end{bmatrix}^{\top}$, as considered in the proof.
Using the iterative process of $\begin{bmatrix}
\mathbf{w}_t,  &\mathbf{u}_t
\end{bmatrix}^{\top}$ in Algorithm~\ref{alg}, we then rewrite the iteration of $\mathbf{w}_t$ as
\begin{equation}\label{w diedai}
\begin{aligned}
 \mathbf{w}_{t+1}-\mathbf{w}_* &=\mathbf{w}_{t}-\mathbf{w}_*-\mathbf{u}_{t+1}
\\ &=\mathbf{w}_{t}-\mathbf{w}_*-\beta\mathbf{u}_t-\eta_t\widehat{\nabla }  f\left (\mathbf{ w}_t  \right ) 
\\ &=\left [ \left ( 1+\beta  \right ) \mathbf{I}-\eta_t\mathbf{H}   \right ] \left ( \mathbf{w}_{t}-\mathbf{w}_* \right ) -\beta \left ( \mathbf{w}_{t-1}-\mathbf{w}_*  \right ) +\eta_t \bm{\zeta } _t. 
\end{aligned}
\end{equation}
We adopt the notation introduced in \eqref{eq:ww}, with $\mathbf{w} _0=\mathbf{w}_{-1}=0 $ and further define $\tilde{\mathbf{H} } =\begin{bmatrix}
 \mathbf{H}  & \mathbf{O} \\
 \mathbf{O}  & \mathbf{O} 
\end{bmatrix}$. 
Substituting the above notation into \eqref{w diedai}, the iteration of $\tilde{\mathbf{w}} _t$ satisfies
\begin{equation}\nonumber
    \tilde{\mathbf{w}} _t =\mathbf{A}_{t-1} \tilde{\mathbf{w}} _{t-1}+\eta_{t-1}\tilde{\bm{\zeta } }_{t-1}.
\end{equation}
The proof of Theorem~\ref{SHB UP} relies on the widely used bias-variance decomposition technique~\citep{bach2013non,jain2017markov,ge2019step,wu2022last,pan2023accelerated}, which decomposes the iteration $ \tilde{\mathbf{w}} _t $  into the bias $ \tilde{\mathbf{w}} _t ^{b} $ and the variance $ \tilde{\mathbf{w}} _t ^{v} $, defined recursively as
\begin{equation}\nonumber
\begin{aligned}
     &\tilde{\mathbf{w}} _t ^{b}=\mathbf{A} _{t-1}\tilde{\mathbf{w}} _{t-1}^{b},\enspace  \tilde{\mathbf{w}} _0 ^{b}= \tilde{\mathbf{w}} _0 ;\\
     &\tilde{\mathbf{w}} _t ^{v}=\mathbf{A} _{t-1}\tilde{\mathbf{w}} _{t-1} ^{v}+\eta_{t-1}\tilde{\bm{\zeta } }_{t-1},\enspace  \tilde{\mathbf{w}} _0 ^{v}= \mathbf{0} ;\\
     &\tilde{\mathbf{w}} _t=\tilde{\mathbf{w}} _t ^{b} +\tilde{\mathbf{w}} _t ^{v} .
     \end{aligned}
\end{equation}
Let $\mathbf{B}_t =\mathbb{E} \left [  \tilde{\mathbf{w}} _t ^{b}  \left (  \tilde{\mathbf{w}} _t ^{b}  \right )^{\top }  \right ] $ and $\mathbf{C}_t =\mathbb{E} \left [  \tilde{\mathbf{w}} _t ^{v} \left (  \tilde{\mathbf{w}} _t ^{v} \right )  ^{\top } \right ] $, then 
 the iterations of $\mathbf{B}_t$ and $\mathbf{C}_t$ satisfy
\begin{equation}\label{BC}
    \begin{aligned}
       & \mathbf{B}_t=\mathbf{A}_{t-1}\mathbf{B}_{t-1}\mathbf{A}_{t-1}^{\top},\enspace  \mathbf{B}_0=\tilde{\mathbf{w} }_0\left (   \tilde{\mathbf{w} }_0 \right ) ^{\top },   
        \\ & \mathbf{C} _{t}=\mathbf{A}_{t-1} \mathbf{C} _{t-1}\mathbf{A} _{t-1}^{\top } +\eta_{t-1}^2 \mathbb{E} \tilde{ \bm{\zeta }}_{t-1} \left ( \tilde{ \bm{\zeta } }_{t-1} \right ) ^{\top}, \enspace \mathbf{C}_0=\mathbf{0}.  
    \end{aligned}
\end{equation}
Therefore, given the quadratic objective $f\left ( \mathbf{w}  \right )$ in \eqref{eq: objective}, the output of Algorithm~\ref{alg} satisfies 

\begin{equation}\label{SHB decomposition}
    \begin{aligned}
        \mathbb{E} \left [ f\left ( \mathbf{w} _T \right )  \right ] -f\left ( \mathbf{\mathbf{w} }^*  \right ) 
&=\frac{1}{2}  \mathbb{E } \left (\mathbf{w} _T -w^*  \right )^{\top}\mathbf{H}   \left (\mathbf{w} _T -w^*  \right )\\
&= \frac{1}{2} \left \langle \tilde{\mathbf{H} },\mathbb{E} \left [\left (\tilde{\mathbf{w}} _T ^{b} +\tilde{\mathbf{w}} _T ^{v}  \right )\left (\tilde{\mathbf{w}} _T ^{b} +\tilde{\mathbf{w}} _T ^{v}  \right ) ^{\top} \right ]   \right \rangle \\
&\overset{\left ( a \right ) }{\le}   \underbrace{\left \langle \tilde{\mathbf{H} },\mathbf{B}_T   \right \rangle}_{\mathrm{Bias} } +\underbrace{\left \langle \tilde{\mathbf{H} },\mathbf{C}_T   \right \rangle}_{\mathrm{Variance}}  ,
    \end{aligned}
\end{equation}
where $\left ( a \right )$ holds by 
\begin{equation}\label{aaT}
   \mathbb{E} \left [\left (\tilde{\mathbf{w}} _T ^{b} +\tilde{\mathbf{w}} _T ^{v}  \right )\left (\tilde{\mathbf{w}} _T ^{b} +\tilde{\mathbf{w}} _T ^{v}  \right ) ^{\top} \right ]\preceq 2\left ( \mathbb{E}\left [\tilde{\mathbf{w}} _T ^{b}\left ( \tilde{\mathbf{w}} _T ^{b} \right ) ^{\top }  \right ] +\mathbb{E}\left [\tilde{\mathbf{w}} _T ^{v}\left ( \tilde{\mathbf{w}} _T ^{v} \right ) ^{\top }  \right ]\right )    .
\end{equation}
The proof of Theorem~\ref{SHB UP} follows from Lemma~\ref{SHB Var} and Lemma~\ref{SHB Bias}. These two lemmas provide the upper bounds for the $\mathrm{Variance}$ term and the $\mathrm{Bias}$ term, respectively.
\setcounter{lemma}{0} 
\renewcommand{\thelemma}{B.\arabic{lemma}}
 \begin{lemma} \label{SHB Var}
 Under the required conditions and the parameter settings as in Theorem~\ref{SHB UP}, $\mathbf{C}_T$ as defined in \eqref{BC} satisfies
    \begin{equation}\nonumber
        \left \langle \tilde{\mathbf{H} },\mathbf{C}_T   \right \rangle \le C_2\sigma ^2 \left ( \frac{\left ( \lambda _1^2\eta_0^2+A^2 \right )k^*}{T} +\eta_0^2 \sum_{j=k^*+1}^{d}\lambda _j^2T\frac{1}{\left ( 1-\sqrt[]{\beta }  \right )^2} \right ),
    \end{equation}
    where $k^*=\mathrm{max}\left \{ k:\lambda _k\eta_0\ge \frac{ 1-\sqrt{\beta}  }{T}  \right \}  $, and $A$ is defined in Theorem \ref{SHB UP} with magnitude of $\widetilde{O}(1)$.
\end{lemma}

\begin{lemma}\label{SHB Bias}
  Under the required conditions and the parameter settings as in Theorem~\ref{SHB UP},  $\mathbf{B}_t$ as defined in \eqref{BC} satisfies 
    \begin{equation}\nonumber
        \left \langle \tilde{\mathbf{H} },\mathbf{B}_T   \right \rangle\le  C_1\left ( \sum_{j=1}^{k^*}\frac{\left ( \mathrm{log}_2 T \right ) ^2A\left ( 1-\sqrt[]{\beta }  \right ) }{\eta_0T}\left ( \mathbf{w} _*^{\left ( j \right ) } \right ) ^2+\sum_{j=k^*+1}^{d}\lambda _j\left ( \mathrm{log}_2 T \right ) ^2\left ( \mathbf{w} _*^{\left ( j \right ) } \right ) ^2 \right ),
    \end{equation}
    where $k^*=\mathrm{max}\left \{ k:\lambda _k\eta_0\ge \frac{ 1-\sqrt{\beta}  }{T} \right \} $, and $A$ is defined in Theorem \ref{SHB UP} with magnitude of $\widetilde{O}(1)$.
\end{lemma}

The proofs of Lemma~\ref{SHB Var} and Lemma~\ref{SHB Bias} are provided at the end of this section.
We now provide the proof of Theorem~\ref{SHB UP}.
\renewcommand{\proofname}{Proof of Theorem~\ref{SHB UP}}
\begin{proof}
The error of the output from Algorithm~\ref{alg} can be bounded above by the sum of the $\mathrm{Variance}$ term and the $\mathrm{Bias}$ term, as follows from Lemma~\ref{SHB Var} and Lemma~\ref{SHB Bias}:
\begin{equation}\nonumber
    \begin{aligned}
      \mathbb{E} \left [ f\left ( \mathbf{w} _T \right )  \right ] -f\left ( \mathbf{w }_*  \right ) \le \underbrace{\left \langle \tilde{\mathbf{H} },\mathbf{B}_T   \right \rangle}_{\mathrm{Bias} } +\underbrace{\left \langle \tilde{\mathbf{H} },\mathbf{C}_T   \right \rangle}_{\mathrm{Variance}} .
        \end{aligned}
    \end{equation}
By Lemma~\ref{SHB Var} and Lemma~\ref{SHB Bias}, we can bound 
 the $\mathrm{Variance}$ term and the $\mathrm{Bias}$ term from above respectively.
\end{proof}
\renewcommand{\proofname}{Proof}
The proofs of Lemma~\ref{SHB Var} and Lemma~\ref{SHB Bias} rely on the properties of the momentum matrix $\mathbf{A}_t^{\left ( j \right )}$ defined as follows,
\begin{equation}\label{mom mat}
   \mathbf{A}_t^{\left ( j \right )} =\begin{bmatrix}
1+\beta -\eta _t\lambda _j  & -\beta \\
 1 &0
\end{bmatrix} \in \mathbb{R} ^{2\times 2},\ \  \forall\  1\le j\le d\ \ \text{and}\ \ 0\le t\le T-1 .
\end{equation}
The necessary properties of the momentum matrix are provided in  Appendix~\ref{sec: pro mat} and Appendix~\ref{sec:PPM2}. 
The proof of Lemma~\ref{SHB Var} is divided into  Lemma~\ref{SHB VZK} and Lemma~\ref{SHB VJS}. Lemma~\ref{SHB VZK} shows that the $\mathrm{Variance}$ term can be bounded from above by the product of momentum matrix. Lemma~\ref{SHB VJS} provides the exact upper bound for each coordinate.
\setcounter{lemma}{9} 
\renewcommand{\thelemma}{B.\arabic{lemma}}
\begin{lemma} \label{SHB VZK}
With $\mathbf{A}_i^{\left ( j \right )}$ defined in \eqref{mom mat}, the $\mathrm{Variance}$ term in Theorem~\ref{SHB UP} can be bounded from above by 
    \begin{equation}\nonumber
        \left \langle \tilde{\mathbf{H} },\mathbf{C}_T   \right \rangle \le \sigma ^2\sum_{j=1}^{d} \sum_{t=0}^{T-1} \lambda _j^2\eta_t^2\left \| \mathbf{A}_{T-1}^{\left ( j \right )} \cdots \mathbf{A}_{t+1}^{\left ( j \right )}  \right \| ^2.
    \end{equation}
\end{lemma}

\begin{lemma}\label{SHB VJS}
    With $\mathbf{A}_i^{\left ( j \right )}$ defined in \eqref{mom mat}, $A=256\mathrm{log}_2 T\cdot\mathrm{ln}T $ and $T\ge 16$, we have:
    \begin{itemize}
        \item For $1\leq j\le k^*$, 
        \begin{equation}\nonumber
            \lambda _j^2\sum_{t=0}^{T-1}\eta_t^2\left \| \mathbf{A}_{T-1}^{\left ( j \right )} \cdots \mathbf{A}_{t+1}^{\left ( j \right )} \right \| ^2\le \frac{\left ( 3^{6}\lambda _1^2\eta_0^2+256A^2 \right )}{T} .
        \end{equation}
        \item For $ j> k^*$,
        \begin{equation}\nonumber
            \lambda _j^2\sum_{t=0}^{T-1}\eta_t^2\left \| \mathbf{A}_{T-1}^{\left ( j \right )} \cdots \mathbf{A}_{t+1}^{\left ( j \right )}  \right \| ^2\le 64\eta_0^2\lambda _j^2T\frac{1}{\left ( 1-\sqrt[]{\beta }  \right )^2 }.
        \end{equation}
    \end{itemize}
    Here, $k^*=\mathrm{max}\left \{ k:\lambda _k\eta_0\ge \frac{ 1-\sqrt{\beta}  }{T}  \right \} $.
\end{lemma}
The detailed proofs of Lemma~\ref{SHB VZK} and Lemma~\ref{SHB VJS} are provided in Appendix~\ref{SEC var}. We now give the proof sketch of Lemma~\ref{SHB VJS}.
The analysis in Lemma~\ref{SHB VJS} is divided into three parts based on the value of $\eta_{t+1}\lambda _j$. Assume that $t+1$ belongs to the $\ell$-th stage, meaning $K\left ( \ell-1 \right ) \le t+1\le K\ell-1$. 
\begin{itemize}
    \item In the first part, we consider the case where $\eta_{t+1}\lambda _j$ satisfies $\eta_{t+1}\lambda _j>4\left ( 1-\sqrt{\beta }  \right ) ^2$. In this part, the eigenvalues of $\mathbf{A}_s^{\left ( j \right )}$ are complex for $s$ in the $\ell+1$-th stage, and $\left \|  \mathbf{A}_{Kl }^{\left ( j \right )} \right \| ^{2K}$ provides exponential contraction. Specifically, we have
    \begin{equation}\nonumber
        \begin{aligned}
            &\lambda _j^2\eta_t^2\left \| \mathbf{A}_{T-1}^{\left ( j \right )} \cdots \mathbf{A}_{t+1}^{\left ( j \right )} \right \| ^2\\
\le &\lambda _1^2\eta _0^2\underbrace{ \left \| \mathbf{A}_{T-1}^{\left ( j \right )}\cdots  \mathbf{A}_{K\ell^*}^{\left ( j \right )} \right \| ^2}_{\text{real, Lemma~\ref{dandiao}} }\underbrace{ \left \|  \mathbf{A}_{K\left ( \ell^*-1 \right ) }^{\left ( j \right )} \right \| ^{2K}\cdots }_{\text{complex, Lemma~\ref{rho complex}} }\underbrace{ \left \|  \mathbf{A}_{K\ell }^{\left ( j \right )} \right \| ^{2K}}_{\text{complex, Lemma~\ref{Fvs2}} } \underbrace{\left \|  \mathbf{A}_{t+1 }^{\left ( j \right )} \right \| ^{2\left ( K-t \right ) }}_{\text{complex, Lemma~\ref{Fvs2}}  }.
        \end{aligned}
    \end{equation}
    \item In the second part, we consider the case where $\eta_{t+1}\lambda _j$ satisfies $\frac{A\left ( 1-\sqrt{\beta} \right ) }{2T}\le \eta_{t+1}\lambda_j\le  4\left ( 1-\sqrt{\beta} \right )^2 $. In this part, the eigenvalues of $\mathbf{A}_s^{\left ( j \right )}$ are real for $s$ from the $\ell+1$-th stage to the $n$-th stage, and $\left \|  \mathbf{A}_{Kl }^{\left ( j \right )} \right \| ^{2K}$ provides exponential contraction.  Specifically, we have
    \begin{equation}\nonumber
        \begin{aligned}
            \lambda _j^2\eta_t^2\left \| \mathbf{A}_{T-1}^{\left ( j \right )} \cdots \mathbf{A}_{t+1}^{\left ( j \right )} \right \| ^2
\le \lambda _1^2\eta _0^2\underbrace{ \left \| \mathbf{A}_{T-1}^{\left ( j \right )}\cdots  \mathbf{A}_{K\left ( \ell+1 \right ) }^{\left ( j \right )} \right \| ^2}_{\text{real, Lemma~\ref{dandiao}} }\underbrace{ \left \|  \mathbf{A}_{K\ell }^{\left ( j \right )} \right \| ^{2K}}_{\text{real, Lemma~\ref{Fvs2}} } \underbrace{\left \|  \mathbf{A}_{t+1 }^{\left ( j \right )} \right \| ^{2\left ( K-t \right ) }}_{\text{arbitrary, Lemma~\ref{Fvs2}}  }.
        \end{aligned}
    \end{equation}
    \item In the third part, we consider the case where  $\eta_{t+1}\lambda _j$ satisfies $\eta_{t+1}\lambda_j< \frac{A\left ( 1-\sqrt{\beta} \right ) }{2T}$. In this part, the eigenvalues of $\mathbf{A}_s^{\left ( j \right )}$ are real for $s\ge t+1$, and $\eta_{t}\lambda _j$ is already sufficiently small to offset the expansion caused by $\left \| \mathbf{A}_{T-1}^{\left ( j \right )} \cdots \mathbf{A}_{t+1}^{\left ( j \right )}  \right \| ^2$. Specifically, we have
    \begin{equation}\nonumber
        \begin{aligned}
            \lambda _j^2\eta_t^2\underbrace{ \left \| \mathbf{A}_{T-1}^{\left ( j \right )} \cdots \mathbf{A}_{t+1}^{\left ( j \right )} \right \| ^2}_{\text{real, Lemma~\ref{dandiao}} }\le \lambda _j^2\eta_t^2\frac{16}{\Delta _{T,j}^2} \left ( \rho \left ( \mathbf{A}_T^{\left ( j \right )}  \right ) \right ) ^{2\left ( T-t -1\right ) }.
        \end{aligned}
    \end{equation}
\end{itemize}

The proof of Lemma~\ref{SHB Bias} is divided into  Lemma~\ref{SHB BZK} and Lemma~\ref{SHB BJS1}. 
 Lemma~\ref{SHB BZK} shows that the $\mathrm{Bias}$ term can be bounded from above by $\sum_{j=1}^{d}\lambda _j\left (  \mathbf{A}_{T-1}^{\left ( j \right )}\cdots \mathbf{A}_{0}^{\left ( j \right )}  \begin{bmatrix}
1 \\1

\end{bmatrix}  \right ) ^2_1\left ( \mathbf{w} _*^{\left ( j \right ) }\right ) ^2$. Lemma~\ref{SHB BJS1} further provides the exact upper bound for each coordinate.
\begin{lemma}\label{SHB BZK}
With $\mathbf{A}_i^{\left ( j \right )}$ defined in \eqref{mom mat}, the $\mathrm{Bias}$ term in Theorem~\ref{SHB UP} can be bounded from above by 
    \begin{equation}\nonumber
        \left \langle \tilde{\mathbf{H} },\mathbf{B}_T   \right \rangle= \sum_{j=1}^{d}\lambda _j\left (  \mathbf{A}_{T-1}^{\left ( j \right )}\cdots \mathbf{A}_{0}^{\left ( j \right )}  \begin{bmatrix}
1 \\1

\end{bmatrix}  \right ) ^2_1\left ( \mathbf{w} _*^{\left ( j \right ) }\right ) ^2.
    \end{equation}
\end{lemma}

\begin{lemma}\label{SHB BJS1}
    With $\mathbf{A}_i^{\left ( j \right )}$ defined in \eqref{mom mat}, $A=256\mathrm{log}_2 T\cdot\mathrm{ln}T $ and $T\ge 16$, we have:
    \begin{itemize}
        \item For $1\le j\le k^*$, 
        \begin{equation}\nonumber
           \lambda _j\left (  \mathbf{A}_{T-1}^{\left ( j \right )}\cdots \mathbf{A}_{0}^{\left ( j \right )} \begin{bmatrix}
1 \\1

\end{bmatrix}  \right ) ^2_1\le 
\frac{2^{10}\left ( \mathrm{log}_2 T \right ) ^2A\left ( 1-\sqrt[]{\beta }  \right ) }{\eta_0T}.
        \end{equation}
        \item For $ j> k^*$, 
        \begin{equation}\nonumber
            \lambda _j\left ( \mathbf{A}_{T-1}^{\left ( j \right )}\cdots \mathbf{A}_{0}^{\left ( j \right )}  \begin{bmatrix}
1 \\1

\end{bmatrix}  \right ) ^2_1\le 2^{10}\lambda _j\left ( \mathrm{log}_2 T \right ) ^2.
        \end{equation}
    \end{itemize}
    Here, $k^*=\mathrm{max} \left \{k:\lambda _k\eta_0\ge \frac{ 1-\sqrt{\beta}  }{T}  \right \} $.
\end{lemma}
The detailed proofs of Lemma~\ref{SHB BZK} and Lemma~\ref{SHB BJS1} are provided in Appendix~\ref{SEC bias}. Below, we provide a proof sketch of Lemma~\ref{SHB BJS1}.
The analysis in Lemma~\ref{SHB BJS1} is divided into three parts based on the value of $\eta_{0}\lambda _j$.
\begin{itemize}
    \item In the first part, we consider the case where $\eta_0\lambda_j$ satisfies $ \eta_0\lambda_j>\left ( 1-\sqrt{\beta }  \right ) ^2$. In this part, the eigenvalues of $\mathbf{A}_s^{\left ( j \right )}$ are complex for $s$ in the first stage, and $\left \| \mathbf{A}_0^{\left ( j \right )}  \right \| ^{2K}$ provides exponential contraction. Specifically, we have
\begin{equation}\nonumber
    \begin{aligned}
         &\lambda _j\left ( \mathbf{A}_{T-1}^{\left ( j \right )}\cdots \mathbf{A}_{0}^{\left ( j \right )}  \begin{bmatrix}
1 \\1

\end{bmatrix}  \right ) ^2_1
\le 2\lambda _1\underbrace{\left \| \mathbf{A}_{T-1}^{\left ( j \right )}\cdots \mathbf{A}_{K\ell}^{\left ( j \right )}  \right \| ^2}_{\text{real, Lemma~\ref{dandiao} } }
\underbrace{\left \|  \mathbf{A}_{\left ( K-1 \right )\ell }^{\left ( j \right )} \right \|^{2K}\cdots}_{\text{complex, Lemma~\ref{rho complex}} } \underbrace{\left \| \mathbf{A}_0^{\left ( j \right )}  \right \| ^{2K}}_{\text{complex, Lemma~\ref{Fvs2} } }.
    \end{aligned}
\end{equation}
    \item In the second part, we consider the case where $\eta_0\lambda_j$ satisfies $ \frac{A\left ( 1-\sqrt[]{\beta }  \right ) }{2T} \le \eta_0\lambda _j\le \left ( 1-\sqrt[]{\beta }  \right )^2  $. In this part, the eigenvalues of $\mathbf{A}_s^{\left ( j \right )}$ are real for $0\le s \le T-1$, while $\left \| \mathbf{A}_0^{\left ( j \right )}  \right \| ^{2K}$ provides exponential contraction. Specifically, we have
    \begin{equation}\nonumber
        \begin{aligned}
            \lambda _j\left ( \mathbf{A}_{T-1}^{\left ( j \right )}\cdots \mathbf{A}_{0}^{\left ( j \right )}  \begin{bmatrix}
1 \\1

\end{bmatrix}  \right ) ^2_1
\le 2\lambda _1\underbrace{\left \| \mathbf{A}_{T-1}^{\left ( j \right )}\cdots \mathbf{A}_{K}^{\left ( j \right )}  \right \| ^2}_{\text{real, Lemma~\ref{dandiao} } }
 \underbrace{\left \| \mathbf{A}_0^{\left ( j \right )}  \right \| ^{2K}}_{\text{real, Lemma~\ref{Fvs2} } }.  
        \end{aligned}
    \end{equation}
    \item In the third part, we consider the case where $\eta_0\lambda_j$ satisfies $\eta_0\lambda_j<\frac{A\left ( 1-\sqrt[]{\beta }  \right ) }{2T}$. In this part, the eigenvalues of $\mathbf{A}_s^{\left ( j \right )}$ are real for $0\le s \le T-1$, while $\lambda_j$ is small enough and  $\left | \left ( \mathbf{A}_{T-1}^{\left ( j \right )}\cdots\mathbf{A}_{0}^{\left ( j \right )} \begin{bmatrix}
1 \\1
\end{bmatrix} \right ) _1 \right |$ is at most on the order of $\mathrm{log}_2T$. Specifically, we have
\begin{equation}\nonumber
    \begin{aligned}
        \underbrace{\left | \left ( \mathbf{A}_{T-1}^{\left ( j \right )}\cdots\mathbf{A}_{0}^{\left ( j \right )}  \begin{bmatrix}
1 \\1
\end{bmatrix} \right ) _1 \right | } _{\text{Lemma~\ref{a b} and Lemma~\ref{bias diedai} } }\lesssim  \mathrm{log}_2 T .
    \end{aligned}
\end{equation}
\end{itemize}

\subsection{Proof of Theorem~\ref{Low}}
In this section, we present the proof of Theorem~\ref{Low}. To derive the min-max error of algorithm class $\mathcal{A} _T $ on the function class $\mathcal{F}_{\bar{\mathbf{w} },\mathbf{H}}\cap\mathcal{F}_0$, 
we construct a ``well-separated"  function class $\mathcal{F}_{\mathbf{H}} \left (\mathsf{W}\right )\subseteq  \mathcal{F}_{\bar{\mathbf{w} },\mathbf{H}}\cap\mathcal{F}_0$ 
 as defined in \eqref{eq:ws f}. We then show that the uniform distribution on $ \mathcal{F}_{\mathbf{H}} \left ( \mathsf{W}\right )$ is a hard instance distribution for stochastic quadratic optimization. 
Recall that the min-max error measures the performance of the best algorithm $\mathrm{A}_{T} \in\mathcal{A} _{T}$ where the performance is required to be uniformly good over the function class $\mathcal{F}_{\bar{\mathbf{w}},\mathbf{H}}\cap\mathcal{F}_0$.  
Since the ``well-separated"  function class $ \mathcal{F}_{\mathbf{H}} \left ( \mathsf{W}\right )$ is a subset of $\mathcal{F}_{\bar{\mathbf{w}},\mathbf{H}}\cap\mathcal{F}_0$, 
 the min-max error can be bounded from below by the infimum of the average error over the uniform distribution on the ``well-separated" function class $ \mathcal{F}_{\mathbf{H}} \left ( \mathsf{W}\right )$. 
Lemma~\ref{Bay} refines this by showing that the error of the output of algorithm $\mathrm{A}_{T}$ can be expressed in a coordinate-separable form, allowing the min-max error to be bounded from below by the infimum of the sum of the average errors in each coordinate.

 The structure of the ``well-separated" function class $\mathcal{F}_{\mathbf{H}} \left( \mathsf{W} \right)$ allows us to establish a lower bound on the average errors in each coordinate for any algorithm $\mathrm{A}_T \in \mathcal{A}_T$. $ \mathcal{F}_{\mathbf{H}} \left ( \mathsf{W}\right )$ consists of all functions $\bar{F}_{\mathbf{w}_{*}}\left ( \mathbf{w},\bar{\bm{\xi }}  \right ) =\frac{1}{2}\mathbf{w}^{\top}\mathbf{H}\mathbf{w} -\left ( \mathbf{H}\mathbf{w}_{*}+\bar{\bm{\xi }}\right ) ^{\top}\mathbf{w}$, where $\mathbf{w}_*\in \mathsf{W}$ and $\bar{\bm{\xi }}\sim  \mathcal{N}\left ( 0,\sigma^2\mathbf{H}  \right )$. For simplicity, denote $\bar{F}_{\mathbf{w}_{*}}\left ( \cdot,\bar{\bm{\xi }}  \right )$ as $\bar{F}_{\mathbf{w}_{*}}$. The candidate parameter set $\mathsf{W}$ can be divided into pairs $\left ( \mathbf{u}_0,\mathbf{u}_1   \right )$ where $\mathbf{u}_0$ and $\mathbf{u}_1$  differ only in the $k$-th coordinate. 
Lemma~\ref{Bay2} will show that the infimum of the average error can be bounded from below by the sum of the infimum of two-point test error of $\left ( \bar{F}_{\mathbf{u}_0},\mathbb{P}_{\bar{\bm{\xi }}} \right )\in \mathcal{F}_{\mathbf{H}} \left ( \mathsf{W}\right )$ and $\left ( \bar{F}_{\mathbf{u}_1},\mathbb{P}_{\bar{\bm{\xi }}} \right )\in \mathcal{F}_{\mathbf{H}} \left ( \mathsf{W}\right )$ in the $k$-th coordinate.
The term ``well-separated" refers to the fact that for any $\mathbf{u}_0,\ \mathbf{u}_1 \in \mathsf{W}$  that differ only in the $k$-th coordinate,  $\left ( \bar{F}_{\mathbf{u}_0},\mathbb{P}_{\bar{\bm{\xi }}} \right )$ is separated from $\left ( \bar{F}_{\mathbf{u}_1},\mathbb{P}_{\bar{\bm{\xi }}} \right )$ in the $k$-th coordinate. 
Finally,  Lemma~\ref{low dimen}, derived from the consequence of the informational constrained risk inequality~\citep{ brown1996constrained},  asserts that for any algorithm $\mathrm{A}_T\in \mathcal{A}_T$ and any pair $\left ( \mathbf{u}_0,\mathbf{u}_1   \right )$, if $\hat{\mathbf{w} }_T^{\left ( k \right ) }$ accurately estimates $\mathbf{u} _0^{\left ( k \right ) }$, then there exists a lower bound on the estimation error of $\hat{\mathbf{w} }_T^{\left ( k \right ) }$  with respect to $\mathbf{u} _1^{\left ( k \right ) }$.

The ``well-separated" function class $ \mathcal{F}_{\mathbf{H}} \left ( \mathsf{W}\right )$ is defined as follows.
For the $k$-th coordinate, define $ w^{\left ( k \right ) }$ based on $\bar{w} ^{\left ( k \right ) }$ in $\mathcal{F}_{\bar{\mathbf{w} },\mathbf{H}}$,
\begin{equation}\label{wk}
    w ^{\left ( k \right ) }=\left\{\begin{matrix}\frac{\sigma }{\left ( T\lambda _k  \right )^{\frac{1}{2} } },&\ \text{if} \ & \lambda _k\left ( \bar{ w} ^{\left ( k \right ) }  \right ) ^2\ge  \frac{\sigma ^2}{T} ,
 \\\bar{ w} ^{\left ( k \right )},\ & \text{if}\ & \lambda _k\left ( \bar{ w} ^{\left ( k \right ) }  \right ) ^2< \frac{\sigma ^2}{T} .
\end{matrix}\right.
\end{equation}
In this definition, $ w^{\left ( k \right ) }$ satisfies $0\le w ^{\left ( k \right ) }\le \bar{ w} ^{\left ( k \right )}$. The candidate parameters set $\mathsf{W}$ comprises all vertices of hypercube $\mathcal{V} =\left \{ 0,w^{\left ( k \right ) } \right \} _{k=1}^d$, and is defined as 
\begin{equation}\nonumber
    \mathsf{W} =\left \{ \mathbf{w} : \mathbf{w}\in \mathbb{R}^d,\quad    \mathbf{w} ^{\left ( k \right ) }=0\quad   \mathrm{or} \quad   w ^{\left ( k \right ) },\quad \forall  1\le k \le d \right \} .
\end{equation}
The ``well-separated" function class $ \mathcal{F}_{\mathbf{H}} \left ( \mathsf{W}\right )$ consists of all functions $\bar{F}_{\mathbf{w}_{*}}\left ( \mathbf{w},\bar{\bm{\xi }}  \right ) =\frac{1}{2}\mathbf{w}^{\top}\mathbf{H}\mathbf{w} -\left ( \mathbf{H}\mathbf{w}_{*}+\bar{\bm{\xi }}\right ) ^{\top}\mathbf{w}$, where $\mathbf{w}_*\in \mathsf{W}$ and $\bar{\bm{\xi }}\sim  \mathcal{N}\left ( 0,\sigma^2\mathbf{H}  \right )$, defined as,
\begin{equation}\label{eq:ws f}
    \mathcal{F}_{\mathbf{H}} \left (\mathsf{W}\right )=\begin{Bmatrix}
\left ( \bar{F}_{\mathbf{w}_{*}}\left ( \cdot,\bar{\bm{\xi }}  \right ),\mathbb{P}_{\bar{\bm{\xi }}} \right ): &\bar{F}_{\mathbf{w}_{*}}\left ( \mathbf{w},\bar{\bm{\xi }}  \right )=\frac{1}{2}\mathbf{w}^{\top}\mathbf{H}\mathbf{w} -\left ( \mathbf{H}\mathbf{w}_{*}+\bar{\bm{\xi }}\right ) ^{\top}\mathbf{w}, \\ &\bar{\bm{\xi }}\sim  \mathcal{N}\left ( 0,\sigma^2\mathbf{H}  \right ),\mathbf{w}_*\in \mathsf{W} 
\end{Bmatrix}.
\end{equation}
Since $0\le w ^{\left ( k \right ) }\le \bar{w} ^{\left ( k \right )}$ and $\mathrm{Var}_{\bar{\bm{\xi }}\sim \mathbb{P}_{\bar{\bm{\xi }}}}\left(\nabla \bar{F}_{\mathbf{w}_{*}}\right) \preceq \sigma^2\mathbf{H}$, the ``well-separated" function class $\mathcal{F}_{\mathbf{H}} \left ( \mathsf{W}\right )$ belongs to function class $\mathcal{F}_{\bar{\mathbf{w} },\mathbf{H}}\cap\mathcal{F}_0$. 
By the definition of $\mathcal{F}_{\mathbf{H}} \left (\mathsf{W}\right )$, the mapping from $\mathbf{w}_*\in \mathsf{W}$ to $\left ( \bar{F}_{\mathbf{w}_{*}},\mathbb{P}_{\bar{\bm{\xi }}} \right )\in \mathcal{F}_{\mathbf{H}} \left ( \mathsf{W}\right )$ is bijective.
Therefore, if $\tilde{\mathbf{w} }$ follows the uniform distribution on $\mathsf{W}$, then $\left ( \bar{F}_{\tilde{\mathbf{w} }},\mathbb{P}_{\bar{\bm{\xi }}} \right )$ follows the uniform distribution on $\mathcal{F}_{\mathbf{H}} \left (\mathsf{W}\right )$.

Lemma~\ref{Bay}  shows that the error has a coordinate-separable form.
\setcounter{lemma}{0} 
\renewcommand{\thelemma}{A.\arabic{lemma}}
\begin{lemma}
Let $\bm{\pi }$ be the uniform distribution on $\mathsf{W} $.  For any $T\in\mathbb{N}^+$ and positive semidefinite matrix $\mathbf{H}$, we have,
    \begin{equation}\label{a1}
    \begin{aligned}
       &\inf_{\mathrm{A}_{T} \in\mathcal{A} _{T}}\sup_{\left ( F\left ( \cdot ,\bm{\xi} \right ),\mathbb{P}_{\bm{\xi }} \right )\in \mathcal{F}_{\bar{\mathbf{w} },\mathbf{H}} \cap \mathcal{F}_0 }\left( \mathbb{E}\left [f\left (\hat{\mathbf{w} }_T  \right )   \right ]-\inf_{\mathbf{w}}f\left (\mathbf{w}  \right )\right) \\
       \ge &\inf_{\mathrm{A}_{T} \in\mathcal{A} _{T}}\frac{1}{2} \mathbb{E} _{\tilde{\mathbf{w} }\sim \bm{\pi}  } \sum_{k=1}^{d }  \lambda _k \mathbb{E}_{\tilde{\mathbf{w} }}\left |\hat{\mathbf{w} }_T^{\left ( k \right ) }    -\tilde{\mathbf{w} }^{\left ( k \right ) }  \right | ^2,
       \end{aligned}
    \end{equation}
    where the expectation is taken over the $T$ independent observations  $\left \{ \left ( \mathbf{H} \left ( \bm{\xi }_t \right ) ,\mathbf{b} \left ( \bm{\xi }_t  \right )  \right ) \right \} _{t=0}^{T-1} $.   $\mathbb{E}_{\tilde{\mathbf{w} }}$ denotes the expectation taken over the observations generated by $\left ( \bar{F}_{\tilde{\mathbf{w} }},\mathbb{P}_{\bar{\bm{\xi }}} \right )\in\mathcal{F}_{\mathbf{H}}  \left ( \mathsf{W}  \right )$.
    \label{Bay}
\end{lemma}
Notably, Lemma~\ref{Bay} demonstrates that the uniform distribution on the ``well-separated" function class $ \mathcal{F}_{\mathbf{H}} \left ( \mathsf{W}\right )$ is a hard instance distribution for stochastic quadratic optimization.  
This implies that each of our lower bounds is obtained by the average error on  $\mathcal{F}_{\mathbf{H}} \left (\mathsf{W}\right )$. Following Yao's min-max principle~\citep{yao1977probabilistic}, 
we show in Lemma~\ref{lem: ran vs det} that there is no loss of generality in assuming all algorithms  $\mathrm{A}_{T} \in\mathcal{A} _{T}$ are deterministic.
\begin{lemma}
    Let $\mathcal{A}_T^{det}$ be the collection of deterministic algorithms and $\bm{\pi}$ be the uniform distribution on $\mathsf{W}$. For any $T\in\mathbb{N}^+$ and positive semidefinite matrix $\mathbf{H}$, we have,
    \begin{equation}\nonumber
        \begin{aligned}
          \inf_{\mathrm{A}_{T} \in\mathcal{A} _{T}}\sup_{\left ( F\left ( \cdot ,\bm{\xi} \right ),\mathbb{P}_{\bm{\xi }} \right )\in \mathcal{F}_{\bar{\mathbf{w} },\mathbf{H}} \cap \mathcal{F}_0 }\left ( \mathbb{E}\left [f\left (\hat{\mathbf{w} }_T  \right )   \right ]-\inf_{\mathbf{w}} f\left (\mathbf{w}   \right ) \right) 
          \ge \inf_{\mathrm{A}_{T} \in\mathcal{A} _{T}^{det}}\mathbb{E} _{\tilde{\mathbf{w} }\sim \bm{\pi}  } \mathbb{E} _{\tilde{\mathbf{w} }}\left ( \left [f\left (\hat{\mathbf{w} }_T  \right )   \right ]-\inf_{\mathbf{w}} f\left (\mathbf{w}   \right ) \right),
        \end{aligned}
    \end{equation}
    where the expectation is taken over the $T$ independent observations  $\left \{ \left ( \mathbf{H} \left ( \bm{\xi }_t \right ) ,\mathbf{b} \left ( \bm{\xi }_t  \right )  \right ) \right \} _{t=0}^{T-1} $. $\mathbb{E}_{\tilde{\mathbf{w} }}$ denotes the expectation taken over the observations generated by $\left ( \bar{F}_{\tilde{\mathbf{w} }},\mathbb{P}_{\bar{\bm{\xi }}} \right )\in\mathcal{F}_{\mathbf{H}}  \left ( \mathsf{W}  \right )$.
    \label{lem: ran vs det}
\end{lemma}
Therefore, if we prove the lower bound for average error of deterministic algorithm on $\mathcal{F}_{\mathbf{H}} \left (\mathsf{W}\right )$, this yields  the lower bound for algorithm with any additional randomness.  In the following, we restrict our focus to the algorithms $\mathrm{A}_T$ in the deterministic function class.

Next, we define the set of all pairs $\mathbf{u}_0,\ \mathbf{u}_1 \in \mathsf{W}$ that differ only in the $k$-th coordinate. 
Let $\mathsf{W}_0^{\left ( k \right ) }$ be the set of vectors in $\mathsf{W}$ where the $k$-th coordinate is $0$, and $\mathsf{W}_1^{\left ( k \right ) }$ be the set of vectors in $\mathsf{W}$ where the $k$-th coordinate is $w ^{\left ( k \right )}$, defined as,
\begin{equation}\nonumber
    \begin{aligned}
        \mathsf{W}_0^{\left ( k \right ) }=\left \{ \mathbf{w}:\mathbf{w}^{\left ( k \right ) }=0 ,\  \mathbf{w}\in \mathsf{W}  \right \},\ \mathsf{W}_1^{\left ( k \right ) }=\left \{ \mathbf{w}:\mathbf{w}^{\left ( k \right ) }=w ^{\left ( k \right )} ,\  \mathbf{w}\in \mathsf{W}  \right \}.
    \end{aligned}
\end{equation}
For any $\mathbf{u}_0\in \mathsf{W}_0^{\left ( k \right ) } $, 
define $\mathbf{u}_1$ based on $\mathbf{u}_0$ such that  $\mathbf{u}_1^{\left ( j \right )}=\mathbf{u}_0^{\left ( j \right ) }$, for all  $j\ne k$ and $\mathbf{u}_1^{\left ( k \right ) }=w ^{\left ( k \right )}$. This construction ensures that $\mathbf{u}_0 $ and $\mathbf{u}_1$  differ only in the $k$-th coordinate. Given that $\tilde{\mathbf{w} }\sim \bm{\pi} $ is the uniform distribution on $\mathsf{W} $, it follows that 
\begin{equation}\nonumber
    \mathbb{P} \left ( \tilde{\mathbf{w} }=\mathbf{u}_0  \right ) =
\mathbb{P} \left ( \tilde{\mathbf{w} }=\mathbf{u}_1  \right )=\frac{1}{2^d}.
\end{equation}
Define $\mathsf{W} _{0,1}^{\left ( k \right ) }$ as the set of 
all pairs $\mathbf{u}_0\in \mathsf{W}_0^{\left ( k \right ) }$ and $\mathbf{u}_1\in \mathsf{W}_1^{\left ( k \right ) }$ where $\mathbf{u}_0$ and $\mathbf{u}_1$ %and $\mathbf{u}_0$ and $\mathbf{u}_1$
 differ only in the $k$-th coordinate, defined as,
\begin{equation}\nonumber
    \mathsf{W} _{0,1}^{\left ( k \right ) }=\left \{ \left ( \mathbf{u}_0,\mathbf{u}_1   \right ):\mathbf{u}_0\in \mathsf{W}_0^{\left ( k \right ) },\mathbf{u}_1\in \mathsf{W}_1^{\left ( k \right ) } , \mathbf{u}_0 \text{ and }\mathbf{u}_1 \text{  differ only in the $k$-th coordinate} \right \} .
\end{equation}

Lemma~\ref{Bay2} shows that the infimum of the average error on $ \mathcal{F}_{\mathbf{H}} \left ( \mathsf{W}\right )$ can be bounded from below by the sum of the infimum of the two-point testing error for $\left ( \bar{F}_{\mathbf{u}_0},\mathbb{P}_{\bar{\bm{\xi }}} \right )$ and $\left ( \bar{F}_{\mathbf{u}_1},\mathbb{P}_{\bar{\bm{\xi }}} \right )$ in the $k$-th coordinate, where $\left ( \mathbf{u}_0,\mathbf{u}_1   \right )\in \mathsf{W} _{0,1}^{\left ( k \right ) }$. 
\begin{lemma}\label{Bay2}
Let $\bm{\pi }$ be the uniform distribution on $\mathsf{W} $ and $\tilde{\mathbf{w} }\sim \bm{\pi }$.  For any $T\in\mathbb{N}^+$ and positive semidefinite matrix $\mathbf{H}$, we have 
 \begin{equation}\label{Bgeu}
     \begin{aligned}
         & \inf_{\mathrm{A}_{T} \in\mathcal{A} _{T}}\frac{1}{2} \mathbb{E} _{\tilde{\mathbf{w} }\sim \bm{\pi}  } \sum_{k=1}^{d }  \lambda _k \mathbb{E}_{\tilde{\mathbf{w} }}\left |\hat{\mathbf{w} }_T^{\left ( k \right ) }    -\tilde{\mathbf{w} }^{\left ( k \right ) }  \right | ^2\\ \ge  &\frac{1}{2}\sum_{k=1}^{d }\inf_{\mathrm{A}_{T} \in\mathcal{A} _{T}}\underset{\left ( \mathbf{u}_0,\mathbf{u}_1   \right )\in \mathsf{W} _{0,1}^{\left ( k \right ) } }{\mathrm{inf} }\lambda _k\left ( \mathbb{E}_{\mathbf{u}_0} \left | \hat{\mathbf{w} }_T^{\left ( k \right ) }  -\mathbf{u}_0^{\left ( k \right ) } \right |^2  
+\mathbb{E}_{\mathbf{u}_1} \left |  \hat{\mathbf{w} }_T^{\left ( k \right ) } -\mathbf{u}_1^{\left ( k \right ) } \right |^2 \right ),
     \end{aligned}
 \end{equation}
where the expectation is taken over $T$ independent observations $\left \{ \left ( \mathbf{H} \left ( \bm{\xi }_t \right ) ,\mathbf{b} \left ( \bm{\xi }_t  \right )  \right ) \right \} _{t=0}^{T-1} $.  $\mathbb{E}_{\left [ \cdot  \right ] }$ denotes the expectation when the observations are generated by $\left ( \bar{F}_{\left [ \cdot  \right ]},\mathbb{P}_{\bar{\bm{\xi }}} \right )\in\mathcal{F}_{\mathbf{H}}  \left ( \mathsf{W}  \right )$.
\end{lemma}

Lemma~\ref{low dimen} directly follows the informational constrained risk inequality~\citep{ brown1996constrained}, and offers a lower bound on the two-point testing error for $\left ( \bar{F}_{\mathbf{u}_0},\mathbb{P}_{\bar{\bm{\xi }}} \right )$ and $\left ( \bar{F}_{\mathbf{u}_1},\mathbb{P}_{\bar{\bm{\xi }}} \right )$ in the $k$-th coordinate for any algorithm $\mathrm{A}_T\in \mathcal{A}_T$ and any pair $\left ( \mathbf{u}_0,\mathbf{u}_1   \right )\in \mathsf{W} _{0,1}^{\left ( k \right ) } $. 

\begin{lemma}\label{low dimen}
    For any $T\in\mathbb{N}^+$, positive semidefinite matrix $\mathbf{H}$, $\mathrm{A}_{T} \in\mathcal{A} _{T}$ and $\left ( \mathbf{u}_0,\mathbf{u}_1   \right )\in  \mathsf{W} _{0,1}^{\left ( k \right ) } $, if the estimation error of $\hat{\mathbf{w} }_T^{\left ( k \right ) }$  with respect to $\mathbf{u}_0^{\left ( k \right ) }$ satisfies 
\begin{equation}\nonumber
    \mathbb{E} _{\mathbf{u}_0}\left |  \hat{\mathbf{w} }_T^{\left ( k \right ) }-\mathbf{u}_0^{\left ( k \right ) }   \right |^2\le \frac{1}{4} \left ( w ^{\left ( k \right ) } \right )^2,
\end{equation}
then the estimation error of $\hat{\mathbf{w} }_T^{\left ( k \right ) }$  with respect to $\mathbf{u}_1^{\left ( k \right ) }$ satisfies 
\begin{equation}\nonumber
    \mathbb{E} _{\mathbf{u}_1}\left |  \hat{\mathbf{w} }_T^{\left ( k \right ) } -\mathbf{u}_1^{\left ( k \right ) } \right |^2\ge \frac{1}{4} \left ( w ^{\left ( k \right ) } \right )^2,
\end{equation}
 where the expectation is taken over the observations  $\left \{ \left ( \mathbf{H} \left ( \bm{\xi }_t \right ) ,\mathbf{b} \left ( \bm{\xi }_t  \right )  \right ) \right \} _{t=0}^{T-1}  $.   $\mathbb{E}_{\left [ \cdot  \right ] }$ denotes the expectation when observations are generated by $\left ( F_{\left [ \cdot  \right ]},\mathbb{P}_{\bm{\xi }} \right )\in\mathcal{F}_{\mathbf{H}}  \left ( \mathsf{W}  \right )$.
\end{lemma}
The detailed proofs of Lemma~\ref{Bay}, Lemma~\ref{lem: ran vs det},  Lemma~\ref{Bay2}, and Lemma~\ref{low dimen} are provided in  Appendix~\ref{sec:Lemma123}.  
Theorem~\ref{Low} follows from these lemmas.

\renewcommand{\proofname}{Proof of Theorem~\ref{Low}}
\begin{proof}
By combining \eqref{a1} and \eqref{Bgeu}, and applying Lemma~\ref{low dimen} to bound the sum of  $\mathbb{E}_{\mathbf{u}_0} \left | \hat{\mathbf{w} }_T^{\left ( k \right ) }  -\mathbf{u}_0^{\left ( k \right ) } \right |^2  
$ and $\mathbb{E}_{\mathbf{u}_1} \left |  \hat{\mathbf{w} }_T^{\left ( k \right ) } -\mathbf{u}_1^{\left ( k \right ) } \right |^2$ from below for any algorithm  $\mathrm{A}_{T} \in \mathcal{A}_{T}  $ and $\left ( \mathbf{u}_0,\mathbf{u}_1   \right )\in \mathsf{W} _{0,1}^{\left ( k \right ) }$, we obtain 
\begin{equation}\nonumber
    \begin{aligned}
         &\inf_{\mathrm{A}_{T} \in\mathcal{A} _{T}}\sup_{\left ( F\left ( \cdot ,\bm{\xi} \right ),\mathbb{P}_{\bm{\xi }} \right )\in \mathcal{F}_{\bar{\mathbf{w} },\mathbf{H}} \cap \mathcal{F}_0 }\mathbb{E}\left [f\left (\hat{\mathbf{w} }_T  \right )   \right ]- \inf_{\mathbf{w}} f\left (\mathbf{w} \right )\\
         \ge &\frac{1}{2}\sum_{k=1}^{d }\inf_{\mathrm{A}_{T} \in\mathcal{A} _{T}}\underset{\left ( \mathbf{u}_0,\mathbf{u}_1   \right )\in \mathsf{W} _{0,1}^{\left ( k \right ) } }{\mathrm{inf} }\lambda _k\left ( \mathbb{E}_{\mathbf{u}_0} \left | \hat{\mathbf{w} }_T^{\left ( k \right ) }  -\mathbf{u}_0^{\left ( k \right ) } \right |^2  
+\mathbb{E}_{\mathbf{u}_1} \left |  \hat{\mathbf{w} }_T^{\left ( k \right ) } -\mathbf{u}_1^{\left ( k \right ) } \right |^2 \right )\\
\ge & \frac{1}{8}\sum_{k=1}^{d } \left ( w ^{\left ( k \right ) } \right )^2=\frac{1}{8} \left ( \frac{\left |  \mathcal{I} \right |  \sigma^2}{T}+\sum_{i\notin \mathcal{I}}^d\lambda _i\left ( \bar{ w} ^{\left ( i \right ) }  \right ) ^2  \right ),
    \end{aligned}
\end{equation}
where $\mathcal{I}=\left \{ i: i\in \mathbb{Z}^{+},\   \lambda _i\left ( \bar{ w} ^{\left ( i \right ) }  \right ) ^2\ge \frac{\sigma ^2}{T}\right \}$.
 \end{proof}
\renewcommand{\proofname}{Proof}

\begin{figure}[H]
\centering

\subfigure[\scriptsize{$\lambda_i=i^{-4}$, $\lambda_i\left ( \mathbf{w}_*^{\left ( i \right ) }  \right ) ^{2}=i^{-1.5}$}]{
\begin{minipage}[t]{0.32\linewidth}
\centering
\includegraphics[width=2 in]{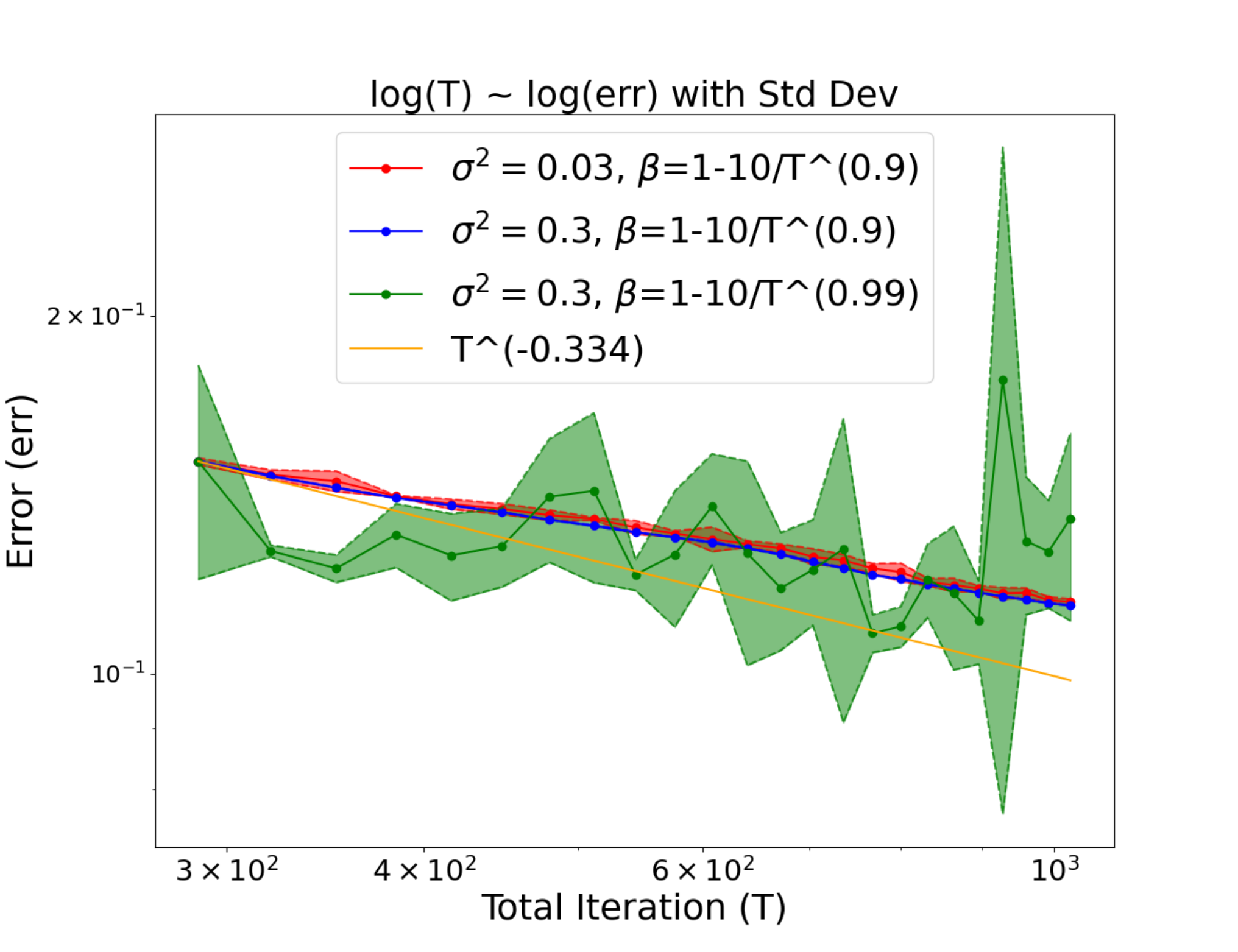}
\end{minipage}%
}%
\subfigure[\scriptsize{$\lambda_i=i^{-3}$, $\lambda_i\left ( \mathbf{w}_*^{\left ( i \right ) }  \right ) ^{2}=i^{-2}$}]{
\begin{minipage}[t]{0.32\linewidth}
\centering
\includegraphics[width=2 in]{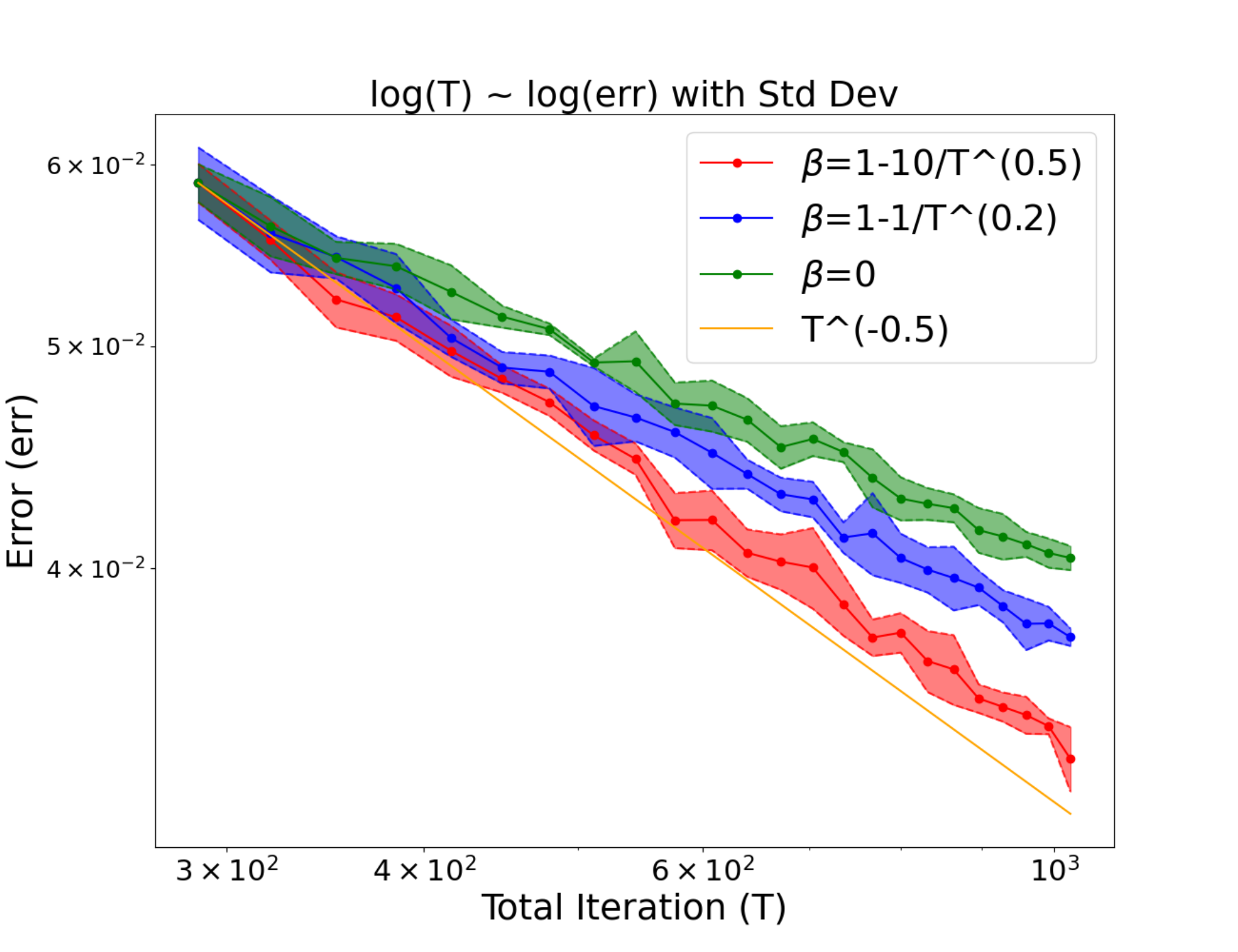}
\end{minipage}%
}%
\subfigure[\scriptsize{$\lambda_i=i^{-a}$, $\mathbf{w}_*^{\left ( i \right ) } =1$}]{
\begin{minipage}[t]{0.32\linewidth}
\centering
\includegraphics[width=2 in]{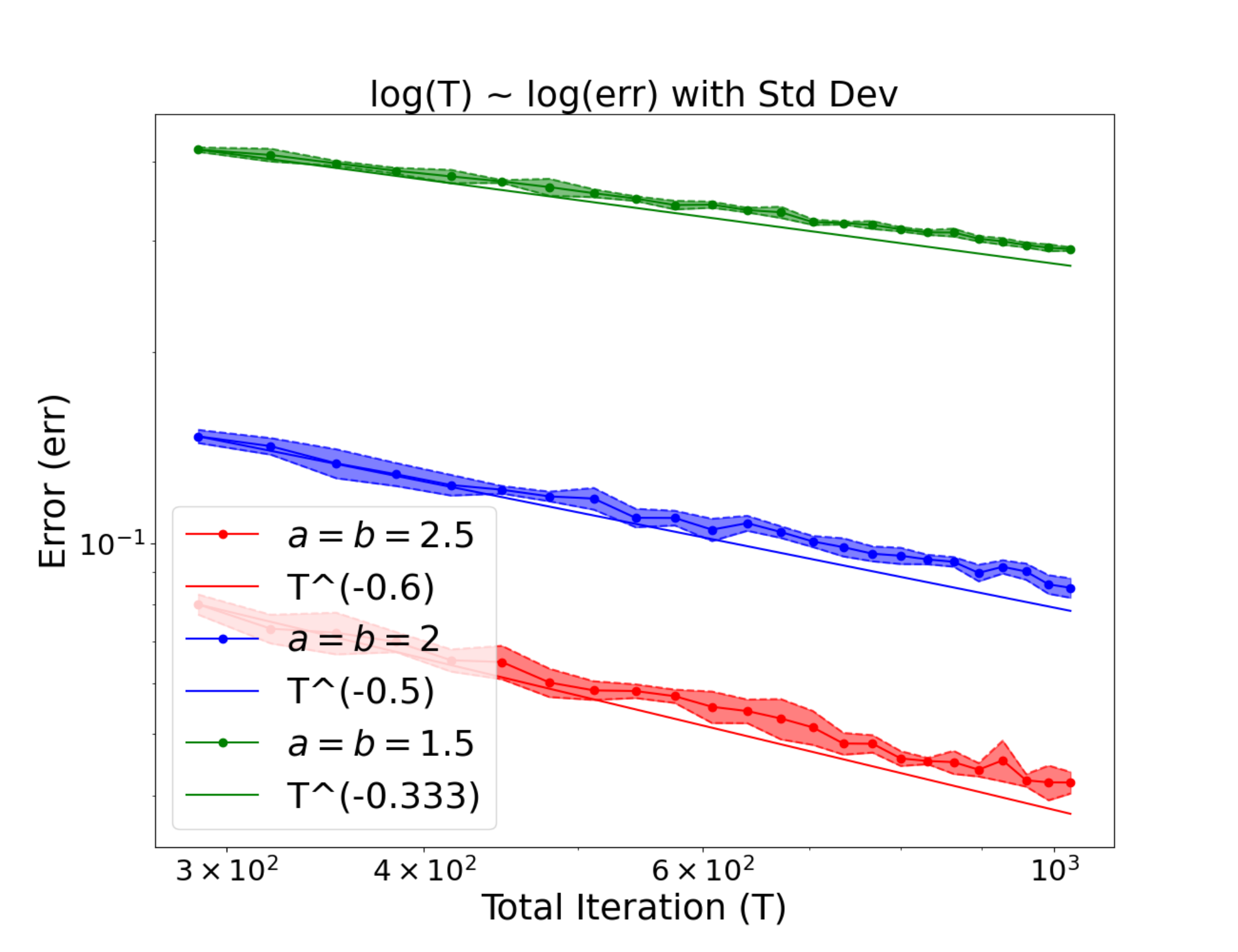}
\end{minipage}%
}%

\vspace{0.5cm} % 在上下两排之间添加适当的间隔

\subfigure[\scriptsize{$\lambda_i=i^{-1}\mathrm{log}^{-c}\left ( i+1 \right ) $, $\mathbf{w}_*^{\left ( i \right ) } =1$}]{
\begin{minipage}[t]{0.32\linewidth}
\centering
\includegraphics[width=2 in]{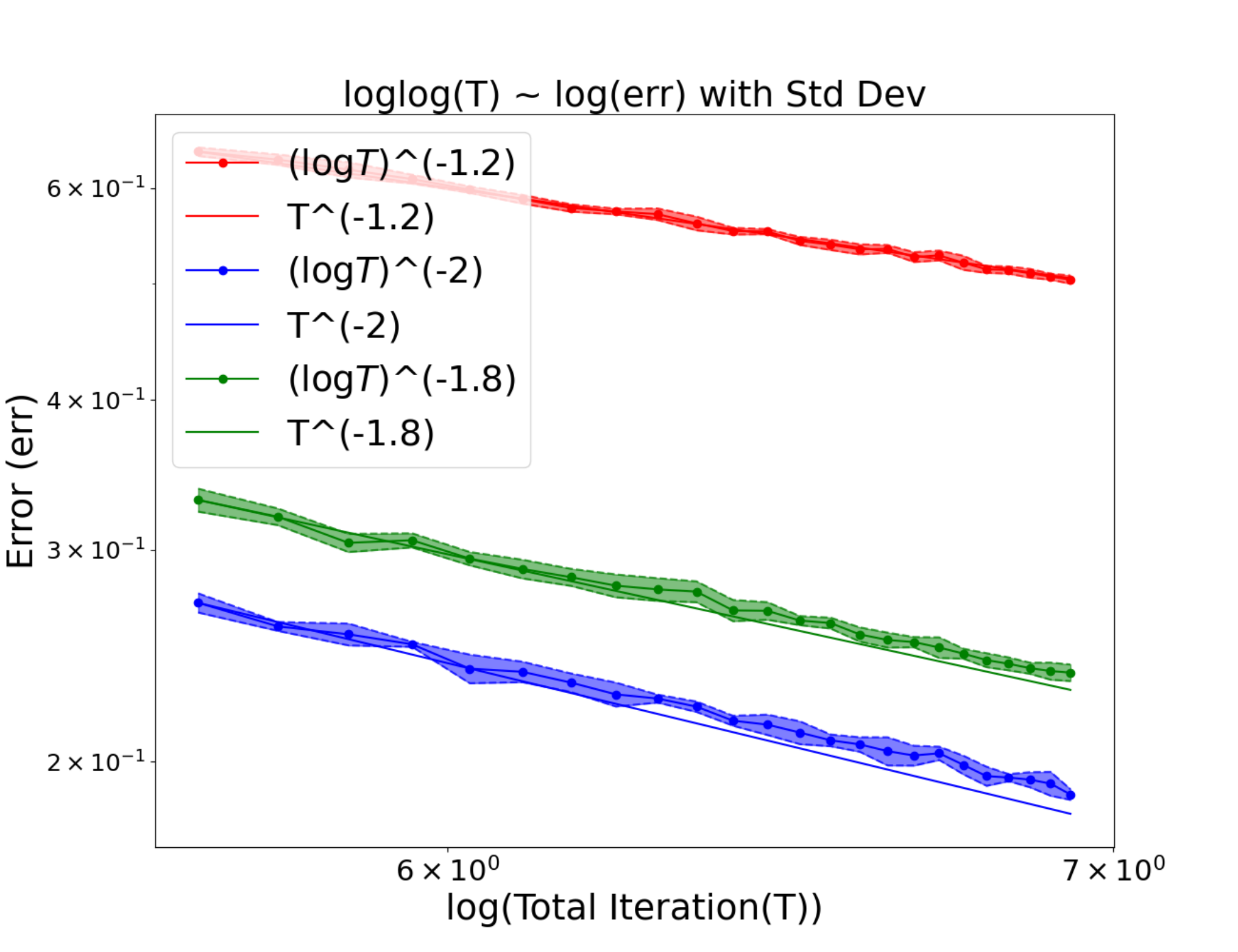}
\end{minipage}%
}%
\subfigure[\scriptsize{$\lambda_i=i^{-1.5}$, $\lambda_i\left ( \mathbf{w}_*^{\left ( i \right ) }  \right ) ^{2}=i^{-3}$}]{
\begin{minipage}[t]{0.32\linewidth}
\centering
\includegraphics[width=2 in]{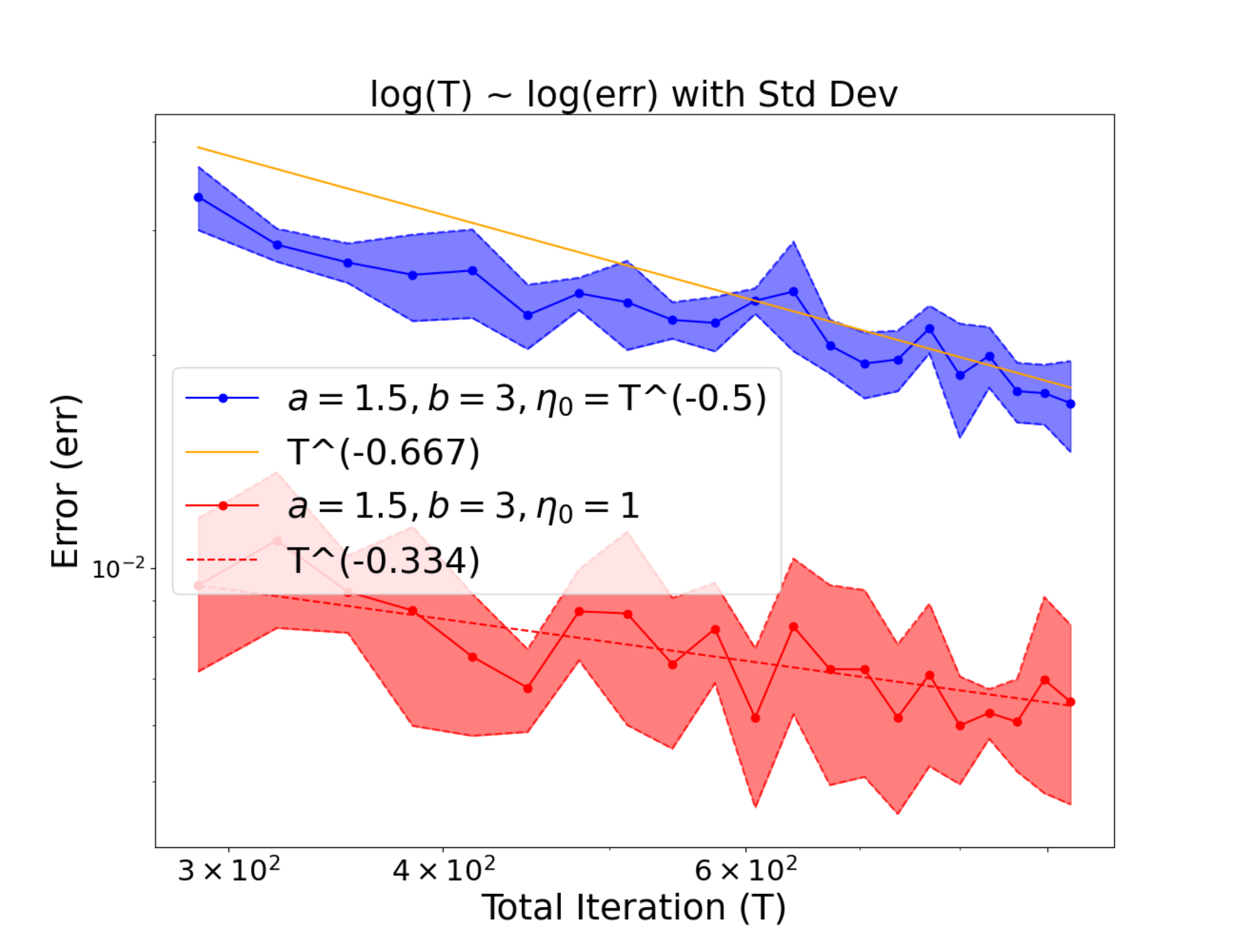}
\end{minipage}%
}%
\subfigure[\scriptsize{$\lambda_i=i^{-1.25}$, $\lambda_i\left ( \mathbf{w}_*^{\left ( i \right ) }  \right ) ^{2}=i^{-3.75}$}]{
\begin{minipage}[t]{0.32\linewidth}
\centering
\includegraphics[width=2 in]{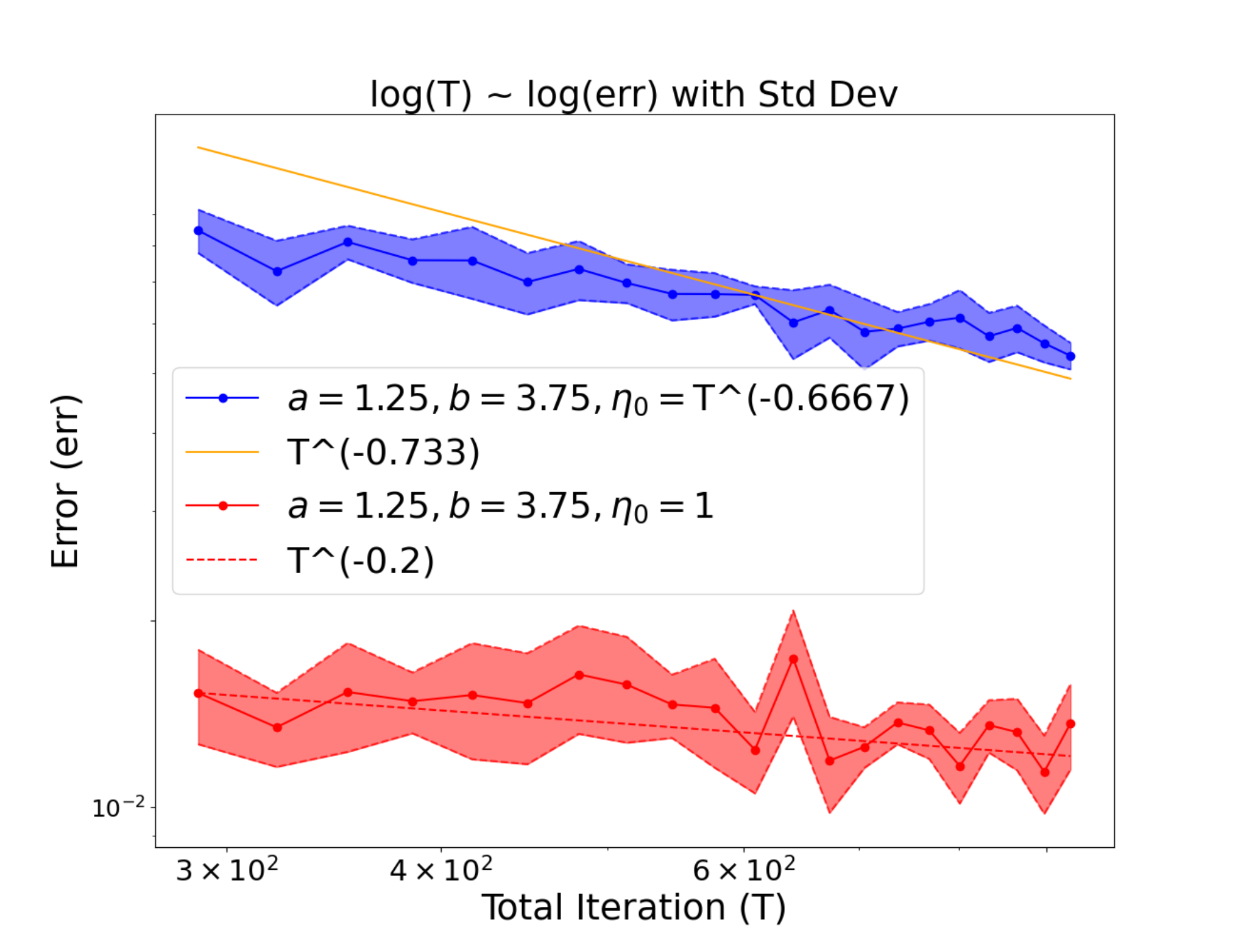}
\end{minipage}%
}%

\centering
\caption{The colored dotted curves represent the last-iterate error, while the solid lines represent the comparison $\frac{D}{T^{\alpha}}$, with $\alpha$ specified in the figure and hand-tuned constants $D$ to adjusted fit the data for each plot. The shaded regions, in corresponding colors, show the areas within one standard deviation.}
\label{experiment}
\end{figure}

\section{Experiments}
In this section, we conduct two sets of experiments. First, we artificially generate stochastic quadratic optimization problems to validate our theoretical findings. Next, we evaluate the performance of the two-layer ReLU-activated neural network in the NTK regime~\citep{arora2019fine}, trained using SGD and SHB, on the CIFAR-2 dataset~\citep{krizhevsky2009learning}.
\subsection{Stochastic Quadratic Optimization}
In this section, we conduct simulations in a high-dimensional setting ($d=30,000$). We artificially generate stochastic quadratic optimization problems as defined in \eqref{eq: objective}, varying the Hessian $\mathbf{H}$ and optimum $\mathbf{w}_*$ to explore different regularities.
We analyze the last-iterate error of SGD and its variants, with exponentially decaying step size, various momentum $\beta$, and initial step size $\eta_0$. 
In each simulation, given a total of $T$ iteration, the algorithm accesses $T$ independent data points $\left \{ \left ( \mathbf{H},\mathbf{H}\mathbf{w}_* +\bm{\xi}   \right )  \right \}_{i=0}^{T-1} 
$, where $\bm{\xi}\sim \mathcal{N}\left ( \mathbf{0},\sigma ^2\mathbf{H}   \right )  $.
Unless otherwise specified, the noise is set to $\sigma^2=0.3$. We repeat each experiment five times and report the average test results.
In the following, we detail the specific experimental settings and present the results obtained for each scenario.

\begin{itemize}
    \item 
    \textbf{Figure~\ref{experiment} (a) Last-iterate error of SHB with an initial step size $\eta_0=0.1$ and momentum $\beta=0.9$  and $0.99$ under polynomially decaying Hessian eigenvalues $\lambda_i=i^{-4}$, the source condition $\lambda_i\left ( \mathbf{w}_*^{\left ( i \right ) }  \right ) ^{2}=i^{-1.5}$, and   noise levels $\sigma=0.3$ and $0.03$:} For SHB with momentum $\beta = 0.9$, performance is similar in both noise settings ($\sigma = 0.3$ and $\sigma = 0.03$), with convergence rates slower than the optimal in both cases. This suggests that in this scenario, the bias term poses the primary challenge to optimization. However, increasing the momentum $\beta$ to 0.99 to accelerate the bias term convergence leads to significant variance, indicating that large momentum can result in exploding variance in this scenario.
 \item
\textbf{Figure~\ref{experiment} (b) Last-iterate error of SHB with an initial step size $\eta_0=0.1$ and momentum $\beta=1-10T^{-0.5}$, $1-T^{-0.2}$, and $0$ under polynomially decaying Hessian eigenvalues $\lambda_i=i^{-3}$ and the source condition $\lambda_i\left ( \mathbf{w}_*^{\left ( i \right ) }  \right ) ^{2}=i^{-2}$:} 
SHB with $\beta=\frac{10}{T^{0.5}}$, as defined in \eqref{parmeter:SHB}, achieves the optimal rate $T^{-\frac{1}{3}}$, whereas SHB with other momentum converges at slower rates.
 \item
\textbf{Figure~\ref{experiment} (c) Last-iterate error of SGD with an initial step size $\eta_0=0.1$ under varying polynomially decaying Hessian eigenvalues $\lambda_i=i^{-a}$ and $\ell_{\infty}$ constraints optimum:} SGD with an initial step size $\eta_0=0.1$ achieves the optimal rate $T^{-1+\frac{1}{a}}$. 
 \item
\textbf{Figure~\ref{experiment} (d) Last-iterate error of SGD with an initial step size $\eta_0=0.1$ under varying decaying Hessian eigenvalues $\lambda_i=i^{-1}\mathrm{log}^{-c}\left ( i+1 \right ) $ and $\ell_{\infty}$ constraints optimum:}  SGD with an initial step size $\eta_0=0.1$ achieves the optimal rate $\left ( \mathrm{log}T  \right ) ^{c}$.
 \item
\textbf{Figure~\ref{experiment} (e) Last-iterate error of SGD with an initial step size $\eta_0=T^{-0.5}$ and $\eta_0=1$  under polynomially decaying Hessian eigenvalues $\lambda_i=i^{-1.5}$ and the source condition $\lambda_i\left ( \mathbf{w}_*^{\left ( i \right ) }  \right ) ^{2}=i^{-3}$:} SGD with an initial step size $\eta_0=T^{-0.5}$, as defined in \eqref{parmeter:SGD},  achieves the optimal rate $T^{-\frac{2}{3}}$. In contrast, the convergence rate of SGD with an initial step size $\eta_0=1$ is closer to the upper bound $T^{-\frac{1}{3}}$.
 \item
\textbf{Figure~\ref{experiment} (f) Last-iterate error of SGD with an initial step size $\eta_0=T^{-\frac{2}{3}}$  and $\eta_0=1$ under polynomially decaying Hessian eigenvalues $\lambda_i=i^{-1.25}$ and the source condition $\lambda_i\left ( \mathbf{w}_*^{\left ( i \right ) }  \right ) ^{2}=i^{-3.75}$:} SGD with an initial step size $\eta_0=T^{-\frac{2}{3}}$, as defined in \eqref{parmeter:SGD}, achieves the  optimal rate $T^{-0.733}$. In contrast, the convergence rate of SGD with an initial step size $\eta_0=1$ is closer to the upper bound $T^{-0.2}$.
\end{itemize}
\subsection{Linear Regression on CIFAR-2}
\begin{figure}[H]
\centering

\subfigure[]{
\begin{minipage}[t]{0.42\linewidth}
\centering
\includegraphics[width=2.7 in]{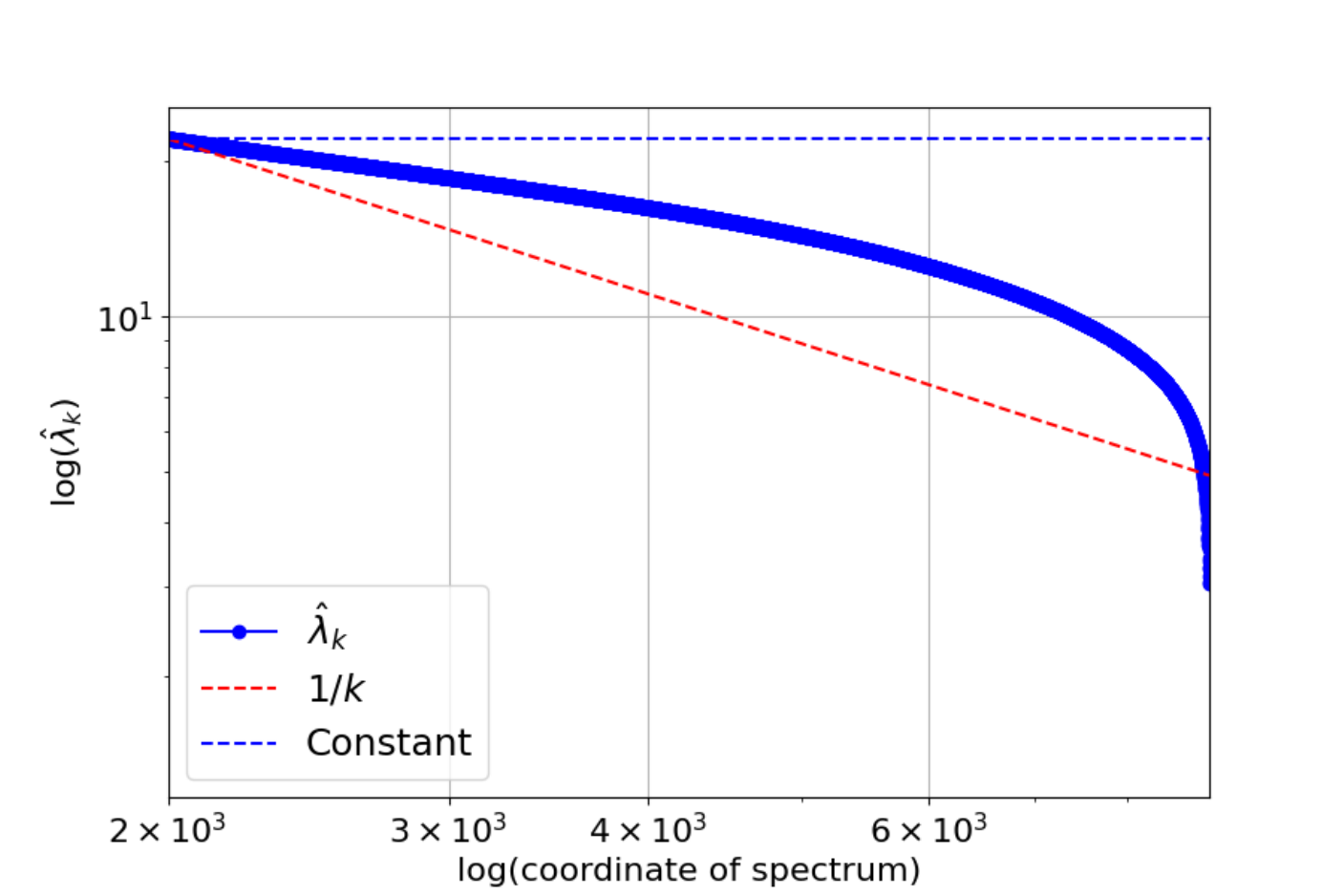}
%\caption{fig1}
\end{minipage}%
}%
\subfigure[]{
\begin{minipage}[t]{0.42\linewidth}
\centering
\includegraphics[width=2.7 in]{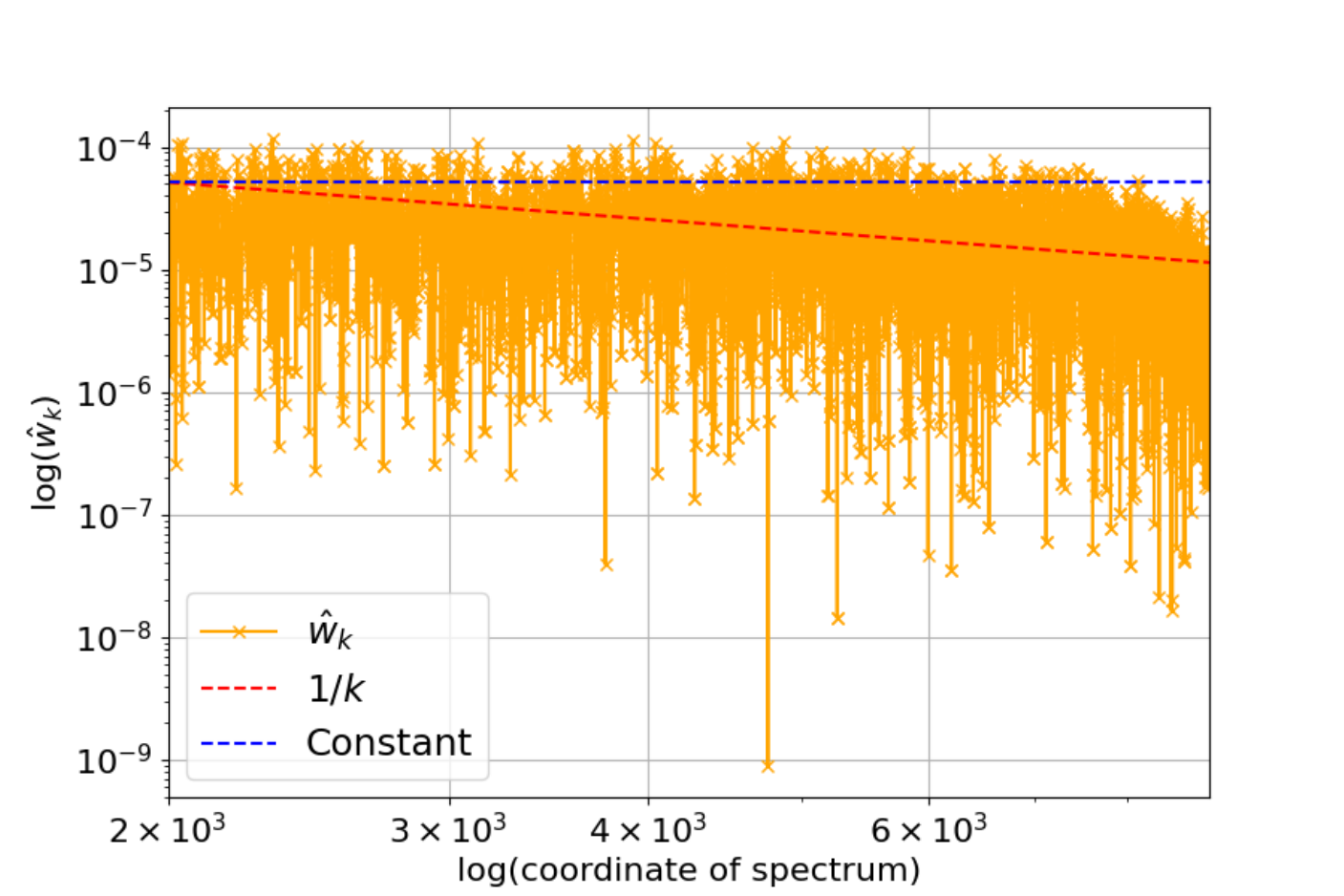}
%\caption{fig2}
\end{minipage}%
}%

\subfigure[]{
\begin{minipage}[t]{0.42\linewidth}
\centering
\includegraphics[width=2.7 in]{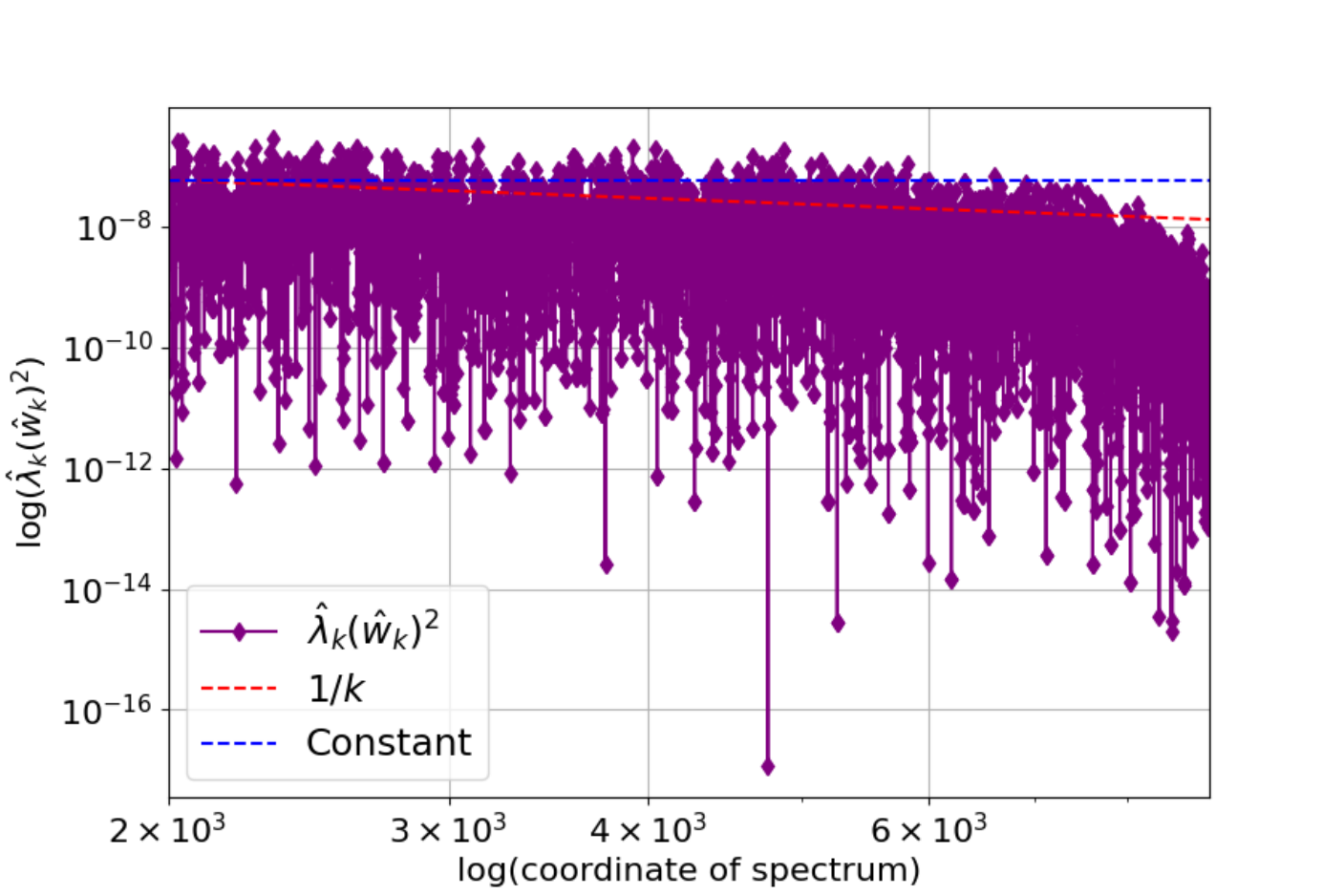}
%\caption{fig2}
\end{minipage}
}%
\subfigure[]{
\begin{minipage}[t]{0.42\linewidth}
\centering
\includegraphics[width=2.7 in]{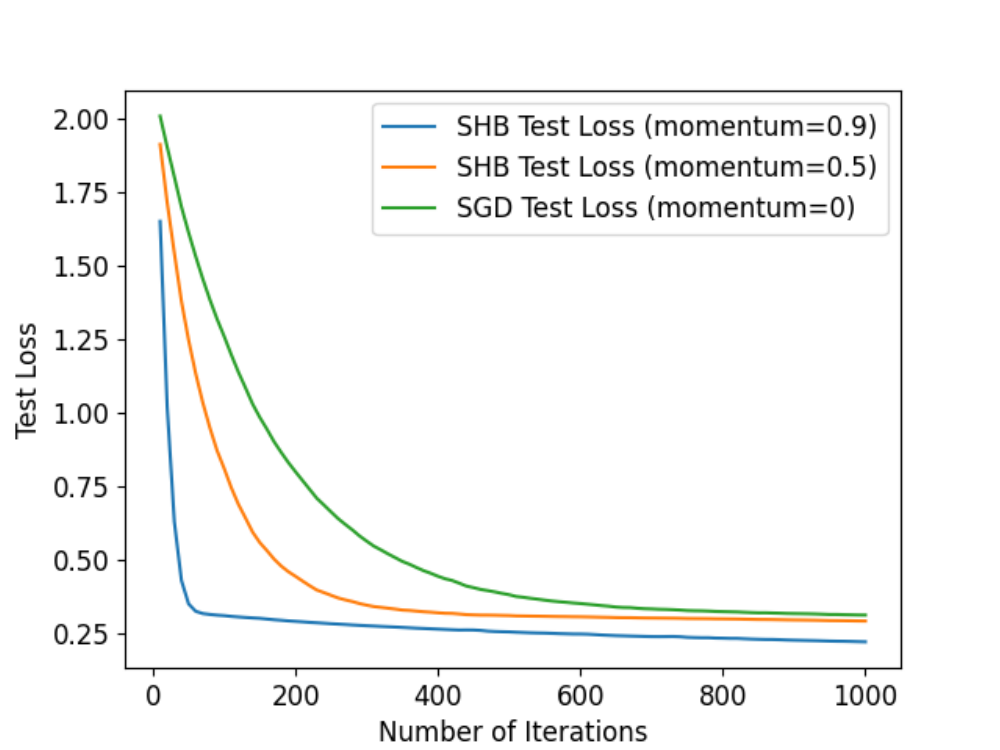}
%\caption{fig2}
\end{minipage}
}%

\centering
\caption{ (a) The values $\hat{\lambda } _j$ from the 2000th to the 9000th terms. (b) The values $\hat{w} _j$ from the 2000th to the 9000th terms. (c) The values $\hat{\lambda } _j\left ( \hat{w} _j \right ) ^2$ from the 2000th to the 9000th terms. (d) The test loss of SGD and SHB with momentum $\beta=0.5$ and $\beta=0.9$.}
\label{CIFAR}
\end{figure}
We use the CIFAR-2 dataset, which consists of the first two classes in CIFAR-10, containing 10,000 training samples and 2,000 test samples. The data is preprocessed such that in each sample $\left ( \mathbf{x} _i,y_i \right )$, the image $\mathbf{x} _i$  is normalized to have unit norm, and the label $y_i$ is set to either +1 or -1 representing the two classes.
We adopt the two-layer ReLU activated neural network architecture as defined in \citet{arora2019fine},  which operates in the neural tangent kernel (NTK) regime as the width of the neural network tends to infinity. Specifically, we consider a two-layer ReLU activated neural network with $m$ neurons in the hidden layer: $ f_{\mathbf{W} ,\mathbf{a} }(\mathbf{x} )=\frac{1}{\sqrt{m} } \sum_{r=1}^{m}a_r\sigma \left ( \mathbf{w} _r^{\top}\mathbf{x}  \right ) $, 
%\begin{equation}\nonumber
    %f_{\mathbf{W} ,\mathbf{a} }(\mathbf{x} )=\frac{1}{\sqrt{m} } \sum_{r=1}^{m}a_r\sigma \left ( \mathbf{w} _r^{\top}\mathbf{x}  \right ) 
%\end{equation}
where $\mathbf{x}\in \mathbb{R}^d$ is the input, $\mathbf{w}_1,\dots ,\mathbf{w}_m\in \mathbb{R}^d$ are the weight vectors in the first layer, $a_1,\dots ,a_m\in \mathbb{R}$ are the weights in the second layer. The training dynamics of this neural network can be characterized by the neural tangent kernel (NTK), $k\left ( \mathbf{x} ,\mathbf{y}  \right ) =\left \langle \mathbf{x} ,\mathbf{y} \right \rangle\left ( \pi -\mathrm{arccos}\left ( \left \langle \mathbf{x} ,\mathbf{y} \right \rangle \right )   \right )/ 2\pi ,$
\iffalse
\begin{equation}\nonumber
    k\left ( \mathbf{x} ,\mathbf{y}  \right ) =\frac{\left \langle \mathbf{x} ,\mathbf{y} \right \rangle\left ( \pi -\mathrm{arccos}\left ( \left \langle \mathbf{x} ,\mathbf{y} \right \rangle \right )   \right )  }{2\pi },
\end{equation}
\fi 
 when the parameters are initialized as $\mathbf{w}_r\left ( 0 \right )  \sim \mathcal{N}\left ( \mathbf{0},\mathbf{I}   \right )  $,  $a_r\sim \mathrm{unif} \left ( \left \{ -1,1 \right \}  \right )$ and the the width of the neural network $m\to \infty $.
 We express the spectral decomposition of the kernel $k\left ( \mathbf{x} ,\mathbf{y}  \right )$ as $k\left ( \mathbf{x} ,\mathbf{y}  \right )=\sum_{j=1}^{\infty }\lambda _j\phi_j\left ( \mathbf{x}   \right ) \phi_j\left ( \mathbf{y}   \right )$, where $\lambda_j$ are arranged in descending order. The data $\left ( \mathbf{x} _i,y_i \right ) $ is assumed to satisfy the model $y_i=f_*\left ( \mathbf{x} _i \right ) +\epsilon _i$, where $f_*$ belongs to the reproducing kernel Hilbert space associated with the NTK $k\left ( \mathbf{x} ,\mathbf{y}  \right )$, and $\epsilon _i$ is noise independent of the data. Let $f_*\left ( \mathbf{x}  \right )$ be expressed in terms of the eigenvectors of the NTK as $f_*\left ( \mathbf{x}  \right ) =\sum_{j=1}^{\infty }w_j\phi\left ( \mathbf{x}  \right )$. We randomly select 9000 samples $\left \{ \left ( \mathbf{x}_i,y_i  \right )  \right \} _{i=1}^{5000}$ from the training set. Define $\mathbf{y}=
\begin{bmatrix}
  y_1, & \cdots  &, y_n
\end{bmatrix}^{\top }$ and the matrix $\mathbf{K}$ such that $K_{i,j}=k\left ( \mathbf{x}_i,\mathbf{x}_j  \right ) $ for all $ 1\le i, j\le 5000$. We compute the spectral decomposition of $\mathbf{K}=\mathbf{K}=\sum_{j=1}^{5000}\hat{\lambda }_j\hat{\Phi} _j\left (  \hat{\Phi} _j  \right )^{\top } $,  then use $\hat{\lambda }_j$ as an estimator of $\lambda_j$, and $\hat{w} _j=\frac{1}{n}\left (  \hat{\Phi} _j  \right )^{\top } \mathbf{y}$ as an estimator of $w_j$. 
The values $\hat{\lambda } _j$, $\hat{w} _j$, and $\hat{\lambda } _j\left ( \hat{w} _j \right ) ^2$ from the 2000th to the 9000th terms are plotted in Figure~\ref{CIFAR} (a), (b), and (c). It can be observed that the distributions of $\hat{w} _j$ and $\hat{\lambda } _j\left ( \hat{w} _j \right ) ^2$ are relatively dense. This observation suggests that adding momentum could potentially improve its generalization ability, especially in low-noise scenarios. We verify this by training a 1024-width two-layer ReLU-activated neural network under the conditions set by \citet{arora2019fine}, using both SGD and SHB with momentum $\beta=0.5$ and $\beta=0.9$ employing exponentially decaying step size schedule with a decay rate of $\frac{1}{4}$. We set the batch size to 10 and the epoch to 1 to approximate online data acquisition. As shown in Figure~\ref{CIFAR} (d),  adding momentum can accelerate SGD and achieve a better performance.

\section{Conclusion}
In this paper, we study the optimality of (Accelerated) SGD for
high-dimensional quadratic optimization. We show for the target function class that has suitable power-law decays, SGD and its variants can achieve the min-max optimality under a wide regime. The future work concerns the case $a>2b$. We conjecture that more computational costs are essentially required under this regime.

\newpage
\bibliographystyle{plainnat}
\bibliography{main}

\newpage

\appendix
\section{Deferred Proof of Theorem~\ref{Low}}

\subsection{Proof of Lemmas~\ref{Bay}, \ref{lem: ran vs det}, \ref{Bay2}, and \ref{low dimen}}\label{sec:Lemma123}
We begin by proving Lemma~\ref{Bay}.
\renewcommand{\proofname}{Proof of Lemma~\ref{Bay}}
 \begin{proof}
Recall that for any quadratic objective $ f\left (\mathbf{w}  \right )=\mathbb{E}_{\bm{\xi} }\left [ F\left ( \mathbf{w} ,\bm{\xi} \right )  \right ]$. 
 $\mathbf{w}_{*}$ denotes the optimum of $f(\mathbf{w}) = \mathbb{E}_{\bm\xi}[F(\mathbf{w}, \bm\xi)]$.
For any algorithm $\mathrm{A} _T\in \mathcal{A} _T$ and a pair $\left ( F\left ( \cdot ,\bm{\xi} \right ),\mathbb{P}_{\bm{\xi }} \right )\in \mathcal{F}_{\bar{\mathbf{w} },\mathbf{H}} \cap \mathcal{F}_0$, the error of the algorithm's output can be expressed in a coordinate-separable form as follows, 
\begin{equation}\nonumber
    \begin{aligned}
        \mathbb{E}\left [f\left (\hat{\mathbf{w} }_T  \right )   \right ]-\inf_{\mathbf{w} }f\left (\mathbf{w} \right )
=\mathbb{E}\left [ \frac{1}{2}\left (\hat{\mathbf{w} }_T-\mathbf{w}_{*} \right ) ^{\top } \mathbf{H} \left ( \hat{\mathbf{w} }_T-\mathbf{w}_{*} \right ) \right ] 
=\frac{1}{2}\sum_{k=1}^d  \lambda _k \mathbb{E}\left | \left ( \hat{\mathbf{w} }_T \right )^{\left ( k \right ) }    -\mathbf{w}_{*}^{\left ( k \right )}   \right | ^2.
    \end{aligned}
\end{equation}
Since  $\mathcal{F}_{\mathbf{H}}  \left ( \mathsf{W}  \right )\subseteq \mathcal{F}_{\bar{\mathbf{w} },\mathbf{H}}\cap\mathcal{F}_0 $ and $\bm{\pi }$ is the uniform distribution on $\mathsf{W} $, for any algorithm  $\mathrm{A}_{T} \in\mathcal{A} _{T}$, we have 
 \begin{equation}\label{supwe}
 \begin{aligned}
     \sup_{\left ( F\left ( \cdot ,\bm{\xi} \right ),\mathbb{P}_{\bm{\xi }} \right )\in \mathcal{F}_{\bar{\mathbf{w} },\mathbf{H}} \cap \mathcal{F}_0  }\sum_{k=1}^d  \lambda _k \mathbb{E}\left | \hat{\mathbf{w} }_T^{\left ( k \right ) }    -\mathbf{w}_{*}^{\left ( k \right )}   \right | ^2
     \ge &
     \sup_{\left ( \bar{F}_{\mathbf{w}},\mathbb{P}_{\bar{\bm{\xi }}} \right )\in \mathcal{F}_{\mathbf{H}}  \left ( \mathsf{W}  \right )}
    \sum_{k=1}^d  \lambda _k \mathbb{E}_{\mathbf{w}}\left | \hat{\mathbf{w} }_T^{\left ( k \right ) }    -\mathbf{w}^{\left ( k \right )}   \right | ^2\\
    & \ge \mathbb{E} _{\tilde{\mathbf{w} }\sim \bm{\pi}  } \sum_{k=1}^{d }  \lambda _k \mathbb{E}_{\tilde{\mathbf{w} }}\left |  \hat{\mathbf{w} }_T^{\left ( k \right ) }    -\tilde{\mathbf{w} }^{\left ( k \right ) }  \right | ^2.
    \end{aligned}
 \end{equation}
Consequently, taking infimum of \eqref{supwe} with respect to $\mathrm{A}_{T} \in\mathcal{A} _{T}$, we obtain that the min-max error on function class $\mathcal{F}_{\bar{\mathbf{w} },\mathbf{H}}\cap\mathcal{F}_0$ can be bounded from below by the infimum of the average error with respect to $\left ( \bar{F}_{\tilde{\mathbf{w} }},\mathbb{P}_{\bar{\bm{\xi }}} \right )$.
    \end{proof}
\renewcommand{\proofname}{Proof}
Then we prove Lemma~\ref{lem: ran vs det}.
\renewcommand{\proofname}{Proof of Lemma~\ref{lem: ran vs det}}
\begin{proof}
Let $\bm{\bar{\zeta}  }_{1:T}$ denote the joint distribution of $\bm{\bar{\zeta}  }_{1}, \dots ,\bm{\bar{\zeta}  }_{T}$.
Since  $\mathcal{F}_{\mathbf{H}}  \left ( \mathsf{W}  \right )\subseteq \mathcal{F}_{\bar{\mathbf{w} },\mathbf{H}}\cap\mathcal{F}_0 $, for any algorithm  $\mathrm{A}_{T} \in\mathcal{A} _{T}$,  we have 
\begin{equation}\nonumber
    \begin{aligned}
        \sup_{\left ( F\left ( \cdot ,\bm{\xi} \right ),\mathbb{P}_{\bm{\xi }} \right )\in \mathcal{F}_{\bar{\mathbf{w} },\mathbf{H}} \cap \mathcal{F}_0 }\left ( \mathbb{E}\left [f\left (\hat{\mathbf{w} }_T  \right )   \right ]-\inf_{\mathbf{w}} f\left (\mathbf{w}   \right ) \right)\ge \sup_{\left ( \bar{F}_{\mathbf{w}},\mathbb{P}_{\bar{\bm{\xi }}} \right )\in \mathcal{F}_{\mathbf{H}}  \left ( \mathsf{W}  \right )}   \mathbb{E} _{\mathbf{w},\bm{\bar{\zeta}  }_{1:T}}\left ( \left [f\left (\hat{\mathbf{w} }_T  \right )   \right ]-\inf_{\mathbf{w}} f\left (\mathbf{w}   \right ) \right),
    \end{aligned}
\end{equation}
where $\mathbb{E} _{\mathbf{w},\bm{\bar{\zeta}  }_{1:T}}$ denotes the expectation taken over the randomness in $\mathrm{A}_T$ and  the $T$ independent observations $\left \{ \left ( \mathbf{H} \left ( \bm{\xi }_t \right ) ,\mathbf{b} \left ( \bm{\xi }_t  \right )  \right ) \right \} _{t=0}^{T-1}$ generated by $\left ( \bar{F}_{\mathbf{w}},\mathbb{P}_{\bar{\bm{\xi }}} \right )\in\mathcal{F}_{\mathbf{H}}  \left ( \mathsf{W}  \right )$.  Let $\mathcal{A}_T^{det}$ be the collection of deterministic algorithms and $\bm{\pi}$ be the uniform distribution on $\mathsf{W}$, according to the Yao's min-max principle~\citep{yao1977probabilistic},  
\begin{equation}\nonumber
    \begin{aligned}
      \sup_{\left ( \bar{F}_{\mathbf{w}},\mathbb{P}_{\bar{\bm{\xi }}} \right )\in \mathcal{F}_{\mathbf{H}}  \left ( \mathsf{W}  \right )}   \mathbb{E} _{ \mathbf{w},\bm{\bar \zeta }}\left ( \left [f\left (\hat{\mathbf{w} }_T  \right )   \right ]-\inf_{\mathbf{w}} f\left (\mathbf{w}   \right ) \right)
      \ge &\mathbb{E} _{\tilde{\mathbf{w} }\sim \bm{\pi}  } \mathbb{E} _{\tilde{\mathbf{w} },\bm{\bar{\zeta}  }_{1:T}}\left ( \left [f\left (\hat{\mathbf{w} }_T  \right )   \right ]-\inf_{\mathbf{w}} f\left (\mathbf{w}   \right ) \right)\\
      \ge &\inf_{\bm{\zeta  }_{1:T}}\mathbb{E} _{\tilde{\mathbf{w} }\sim \bm{\pi}  }  \mathbb{E} _{\tilde{\mathbf{w} }}\left [ \left [f\left (\hat{\mathbf{w} }_T  \right )   \right ]-\inf_{\mathbf{w}} f\left (\mathbf{w}   \right ) | \bm{\bar{\zeta}  }_{1:T}= \bm{\zeta  }_{1:T}\right ] \\
=&\inf_{\mathrm{A}_{T} \in\mathcal{A} _{T}^{det}}\mathbb{E} _{\tilde{\mathbf{w} }\sim \bm{\pi}  } \mathbb{E} _{\tilde{\mathbf{w} }}\left ( \left [f\left (\hat{\mathbf{w} }_T  \right )   \right ]-\inf_{\mathbf{w}} f\left (\mathbf{w}   \right ) \right). 
    \end{aligned}
\end{equation}
\end{proof}
\renewcommand{\proofname}{Proof}
Then we prove Lemma~\ref{Bay2}.
\renewcommand{\proofname}{Proof of Lemma~\ref{Bay2}}
\begin{proof}
Recall that $\mathsf{W} $ can be divided into pairs $\left ( \mathbf{u}_0,\mathbf{u}_1   \right )\in  \mathsf{W} _{0,1}^{\left ( k \right ) }$. 
Since  $\bm{\pi} $ is the uniform distribution on $\mathsf{W} $, for any $\mathrm{A}_{T} \in\mathcal{A} _{T}$,  the average error can be written as the sum of two-point test errors as follows,
\begin{equation}\nonumber
    \begin{aligned}
          &\mathbb{E} _{\tilde{\mathbf{w} }\sim \bm{\pi}  } \sum_{k=1}^{d }  \lambda _k \mathbb{E}_{\tilde{\mathbf{w} }}\left | \hat{\mathbf{w} }_T^{\left ( k \right ) }    -\tilde{\mathbf{w} }^{\left ( k \right ) }  \right | ^2 \\
=&\frac{1}{2^{d}}\sum_{k=1}^{d }\sum_{\left ( \mathbf{u}_0,\mathbf{u}_1   \right )\in \mathsf{W} _{0,1}^{\left ( k \right ) }}\lambda _k\left ( \mathbb{E}_{\mathbf{u}_0} \left | \hat{\mathbf{w} }_T^{\left ( k \right ) }  -\mathbf{u}_0^{\left ( k \right ) } \right |^2
+\mathbb{E}_{\mathbf{u}_1} \left |  \hat{\mathbf{w} }_T^{\left ( k \right ) } -\mathbf{u}_1^{\left ( k \right ) } \right |^2 \right ) \\
\ge &\frac{1}{2}\sum_{k=1}^{d }\underset{\left ( \mathbf{u}_0,\mathbf{u}_1   \right )\in \mathsf{W} _{0,1}^{\left ( k \right ) } }{\mathrm{inf} }\lambda _k\left ( \mathbb{E}_{\mathbf{u}_0} \left |   \hat{\mathbf{w} }_T^{\left ( k \right ) }  -\mathbf{u}_0^{\left ( k \right ) } \right |^2
+\mathbb{E}_{\mathbf{u}_1} \left |  \hat{\mathbf{w} }_T^{\left ( k \right ) } -\mathbf{u}_1^{\left ( k \right ) } \right |^2 \right ).
    \end{aligned}
\end{equation}
Taking the infimum on each coordinate of the right-hand side of the above inequality with respect to algorithm $\mathrm{A}_{T} \in \mathcal{A}_{T}  $, we have 
\begin{equation}\nonumber
    \begin{aligned}
        &\frac{1}{2}\sum_{k=1}^{d }\underset{\left ( \mathbf{u}_0,\mathbf{u}_1   \right )\in \mathsf{W} _{0,1}^{\left ( k \right ) } }{\mathrm{inf} }\lambda _k\left ( \mathbb{E}_{\mathbf{u}_0} \left |  \hat{\mathbf{w} }_T^{\left ( k \right ) }  -\mathbf{u}_0^{\left ( k \right ) } \right |^2
+\mathbb{E}_{\mathbf{u}_1} \left | \hat{\mathbf{w} }_T^{\left ( k \right ) } -\mathbf{u}_1^{\left ( k \right ) } \right |^2 \right )\\
\ge &\frac{1}{2}\sum_{k=1}^{d }\inf_{\mathrm{A}_{T} \in\mathcal{A} _{T} }\underset{\left ( \mathbf{u}_0,\mathbf{u}_1   \right )\in \mathsf{W} _{0,1}^{\left ( k \right ) } }{\mathrm{inf} }\lambda _k\left ( \mathbb{E}_{\mathbf{u}_0} \left | \hat{\mathbf{w} }_T^{\left ( k \right ) }  -\mathbf{u}_0^{\left ( k \right ) } \right |^2
+\mathbb{E}_{\mathbf{u}_1} \left |  \hat{\mathbf{w} }_T^{\left ( k \right ) } -\mathbf{u}_1^{\left ( k \right ) } \right |^2 \right ).
    \end{aligned}
\end{equation}
Consequently, taking the infimum on the average error with respect to algorithm $\mathrm{A}_{T} \in \mathcal{A}_{T}  $, we can proof Lemma~\ref{Bay2}.
\end{proof}
\renewcommand{\proofname}{Proof}
The proof of Lemma~\ref{low dimen} requires the following lemma which calculates $ \mathbb{E} _{\mathbf{u}_0}\left (\frac{\mathrm{d} \mathbb{P}_{\mathbf{u}_1} }{\mathrm{d}  \mathbb{P}_{\mathbf{u}_0}}  \right ) ^2$ .
\setcounter{lemma}{4} 
\renewcommand{\thelemma}{A.\arabic{lemma}}
\begin{lemma}\label{chi}
For any $T\in\mathbb{N}^+$, positive semidefinite matrix $\mathbf{H}$, and $\left ( \mathbf{u}_0,\mathbf{u}_1   \right )\in \mathsf{W} _{0,1}^{\left ( k \right ) } $, let $\mathbb{P}_{\mathbf{u}_0}$ and $\mathbb{P}_{\mathbf{u}_1}$ be the joint distributions of  the observed data points $\left \{ \left ( \mathbf{H} \left ( \bm{\xi }_t \right ) ,\mathbf{b} \left ( \bm{\xi }_t  \right )  \right ) \right \} _{t=0}^{T-1} $ that are generated by $\left ( \bar{F}_{\mathbf{u}_0},\mathbb{P}_{\bar{\bm{\xi }}} \right )$ and $\left ( \bar{F}_{\mathbf{u}_1},\mathbb{P}_{\bar{\bm{\xi }}} \right )$ belonging to  $\mathcal{F}_{\mathbf{H}}  \left ( \mathsf{W}  \right )$. We have 
    \begin{equation}\nonumber
        \mathbb{E} _{\mathbf{u}_0}\left (\frac{\mathrm{d} \mathbb{P}_{\mathbf{u}_1} }{\mathrm{d}  \mathbb{P}_{\mathbf{u}_0}}  \right ) ^2=\mathrm{exp}\left ( \frac{T\lambda _k\left ( w ^{\left ( k \right ) } \right )  ^2 }{\sigma ^2}  \right ) ,
    \end{equation}
    where the expectation is taken over the observations  $\left \{ \left ( \mathbf{H} \left ( \bm{\xi }_t \right ) ,\mathbf{b} \left ( \bm{\xi }_t  \right )  \right ) \right \} _{t=0}^{T-1}$.  $\mathbb{E}_{\left [ \cdot  \right ] }$ denote the expectation when the observations are generated by $\left ( F_{\left [ \cdot  \right ]},\mathbb{P}_{\bm{\xi }} \right )\in\mathcal{F} _{\mathbf{H}} \left ( \mathsf{W}  \right )$.
\end{lemma}
\renewcommand{\proofname}{Proof of Lemma~\ref{chi}}
\begin{proof}
 The observed data points $\left \{ \left ( \mathbf{H} \left ( \bm{\xi }_t \right ) ,\mathbf{b} \left ( \bm{\xi }_t  \right )  \right ) \right \} _{t=0}^{T-1}$ generated by $\left ( \bar{F}_{\mathbf{w}_{*}},\mathbb{P}_{\bar{\bm{\xi }}} \right )\in\mathcal{F}_{\mathbf{H}}  \left ( \mathsf{W}  \right )$ satisfy 
    \begin{equation}\nonumber
        \mathbf{H} \left ( \bm{\xi }_t \right )=\mathbf{H},\ \mathbf{b} \left ( \bm{\xi }_t  \right ) =\mathbf{H}\mathbf{w}_*+\bm{\xi }_t ,\ \bm{\xi }_t \overset{\mathrm{i.i.d} }{\sim}   \mathcal{N}\left ( 0,\sigma^2\mathbf{H}  \right ).
    \end{equation}
For any $0\le t\le T-1$, the stochastic gradient $\left ( \mathbf{H} \left ( \bm{\xi }_t \right ) ,\mathbf{b} \left ( \bm{\xi }_t  \right )  \right ) $ follows the Gaussian distribution $\mathcal{N}\left ( \mathbf{H}  \mathbf{w}_* ,\sigma^2\mathbf{H}  \right )$. Therefore, the probability density functions of the $i$-th coordinate of the stochastic gradient $\left ( \mathbf{H} \left ( \bm{\xi }_t \right ) ,\mathbf{b} \left ( \bm{\xi }_t  \right )  \right ) $ for $\mathbf{w}_*=\mathbf{u}_0$ or $\mathbf{w}_*=\mathbf{u}_1$ have the following forms
\begin{equation}\nonumber
    \begin{aligned}
        \mathrm{p}  _{\mathbf{u}_j } \left (\mathbf{x} _t^{\left ( i \right ) } \right ) 
=\frac{1}{\sqrt{2\pi\lambda _i }\sigma  }\mathrm{exp}\left ( -\frac{\left (\mathbf{x} _t^{\left ( i \right ) }-\lambda _i\left ( \mathbf{u }_j^{\left ( i \right )}  \right ) \right ) ^2}{2\lambda _i\sigma ^2 }  \right ),
    \end{aligned}
\end{equation}
where $j=0$ or $ j=1$ denote $\mathbf{u}_0$ and $\mathbf{u}_1$ respectively.
Since the observed data points $\left ( \mathbf{H} \left ( \bm{\xi }_t \right ) ,\mathbf{b} \left ( \bm{\xi }_t  \right )  \right )_{t=0}^{T-1} $ are independent, the probability density functions of joint distribution for $\mathbf{w}_*=\mathbf{u}_0$ or $\mathbf{w}_*=\mathbf{u}_1$ can be represented by
\begin{equation}\nonumber
    \begin{aligned}
        &\mathrm{d} \mathbb{P}_{\mathbf{u}_j} =\prod_{t=0}^{T-1}\prod_{i=1}^{d}\mathrm{p}  _{\mathbf{u}_j } \left (\mathbf{x} _t^{\left ( i \right ) } \right ) \mathrm{d} \left ( \mathbf{x} _t^{\left ( i \right ) }  \right ) ,
    \end{aligned}
\end{equation}
where $j=0$ or $ j=1$ denote $\mathbf{u}_0$ and $\mathbf{u}_1$ respectively.
Then we calculate $\mathbb{E} _{\mathbf{u}_0}\left (\frac{\mathrm{d} \mathbb{P}_{\mathbf{u}_1} }{\mathrm{d}  \mathbb{P}_{\mathbf{u}_0}}  \right ) ^2$,
\begin{equation}\nonumber
    \begin{aligned}
     &\mathbb{E} _{\mathbf{u}_0}\left (\frac{\mathrm{d} \mathbb{P}_{\mathbf{u}_1} }{\mathrm{d}  \mathbb{P}_{\mathbf{u}_0}}  \right ) ^2
     =\prod_{t=0}^{T-1}\prod_{i=1}^{d}\int \frac{\left ( \mathrm{p}  _{\mathbf{u}_1 } \left (\mathbf{x} _t^{\left ( i \right ) }\right ) \right ) ^2 }{\mathrm{p}  _{\mathbf{u}_0 } \left (\mathbf{x} _t^{\left ( i \right ) } \right )}\mathrm{d} \left ( \mathbf{x} _t^{\left ( i \right ) }  \right ) \\ 
=&\prod_{t=0}^{T-1} \prod_{i=1}^{d}\int  \frac{1}{\sqrt{2\pi \lambda _i}\sigma  }\mathrm{exp}\left (-\frac{\left (  \mathbf{x} _t^{\left ( i \right ) }-\lambda _i\left ( \mathbf{u }_1^{\left ( i \right )}  \right ) \right ) ^2}{\lambda _i\sigma ^2 } +\frac{\left (  \mathbf{x} _t^{\left ( i \right ) }-\lambda _i\left ( \mathbf{u }_0^{\left ( i \right )}  \right ) \right ) ^2 }{2\lambda _i\sigma ^2 }  \right ) \mathrm{d} \left ( \mathbf{x} _t^{\left ( i \right ) } \right )  .
    \end{aligned}
\end{equation}
We separate the above equation into two parts 
 $i \ne k$ and $i=k$ as follows,
\begin{equation}\label{eq:i,k}
    \begin{aligned}
        &\prod_{t=0}^{T-1} \prod_{i=1}^{d}\int  \frac{1}{\sqrt{2\pi \lambda _i}\sigma  }\mathrm{exp}\left (-\frac{\left (  \mathbf{x} _t^{\left ( i \right ) }-\lambda _i\left ( \mathbf{u }_1^{\left ( i \right )}  \right ) \right ) ^2}{\lambda _i\sigma ^2 } +\frac{\left (  \mathbf{x} _t^{\left ( i \right ) }-\lambda _i\left ( \mathbf{u }_0^{\left ( i \right )}  \right ) \right ) ^2 }{2\lambda _i\sigma ^2 }  \right ) \mathrm{d} \left ( \mathbf{x} _t^{\left ( i \right ) } \right )  \\
=&\prod_{t=0}^{T-1} \prod_{i\ne k} \int  \frac{1}{\sqrt{2\pi \lambda _i}\sigma  }\mathrm{exp}\left (-\frac{\left (  \mathbf{x} _t^{\left ( i \right ) }-\lambda _i\left ( \mathbf{u }_1^{\left ( i \right )}  \right ) \right ) ^2}{\lambda _i\sigma ^2 } +\frac{\left (  \mathbf{x} _t^{\left ( i \right ) }-\lambda _i\left ( \mathbf{u }_0^{\left ( i \right )}  \right ) \right ) ^2 }{2\lambda _i\sigma ^2 }  \right ) \mathrm{d} \left ( \mathbf{x} _t^{\left ( i \right ) } \right ) 
\\ &\cdot \prod_{t=0}^{T-1}  \int  \frac{1}{\sqrt{2\pi \lambda _k}\sigma  }\mathrm{exp}\left (-\frac{\left (  \mathbf{x} _t^{\left ( k \right ) }-\lambda _k\left ( \mathbf{u }_1^{\left ( k \right )}  \right ) \right ) ^2}{\lambda _k\sigma ^2 } +\frac{\left (  \mathbf{x} _t^{\left ( k \right ) }-\lambda _k\left ( \mathbf{u }_0^{\left ( k \right )}  \right ) \right ) ^2 }{2\lambda _k\sigma ^2 }  \right ) \mathrm{d} \left ( \mathbf{x} _t^{\left ( k \right ) } \right )    .
    \end{aligned}
\end{equation}
For $i\ne k$, we have $\mathbf{u}_0^{\left ( i \right ) } =\mathbf{u}_1^{\left ( i \right ) }$.  For any $0\le t\le T-1$, the corresponding part in the above equation is an integral of the Gaussian distribution probability density function which is equal to 1 as follows
\begin{equation}\label{eq:i}
    \begin{aligned}
        &\int  \frac{1}{\sqrt{2\pi \lambda _i}\sigma  }\mathrm{exp}\left (-\frac{\left (  \mathbf{x} _t^{\left ( i \right ) }-\lambda _i\left ( \mathbf{u }_1^{\left ( i \right )}  \right ) \right ) ^2}{\lambda _i\sigma ^2 } +\frac{\left (  \mathbf{x} _t^{\left ( i \right ) }-\lambda _i\left ( \mathbf{u }_0^{\left ( i \right )}  \right ) \right ) ^2 }{2\lambda _i\sigma ^2 }  \right ) \mathrm{d} \left ( \mathbf{x} _t^{\left ( i \right ) } \right )   \\
        =&\int  \frac{1}{\sqrt{2\pi \lambda _i}\sigma  }\mathrm{exp}\left ( -\frac{\mathbf{x} _t^{\left ( i \right ) }-\lambda _i\left (\mathbf{u }_1^{\left ( i \right )}  \right )  }{2\lambda _i\sigma ^2 }  \right ) \mathrm{d} \left ( \mathbf{x} _t^{\left ( i \right ) } \right )  \\
        =&1.
    \end{aligned}
\end{equation}
For $i=k$, we have $\mathbf{u}_0^{\left ( k \right ) }=0$ and $\mathbf{u}_1^{\left ( k \right ) }=w ^{\left ( k \right )}$. This implies
\begin{equation}\nonumber
    2\left (\mathbf{x} _t^{\left ( k \right ) }-\lambda _k\left ( \mathbf{u }_1^{\left ( k \right )}  \right ) \right ) ^2-\left (\mathbf{x} _t^{\left ( k \right ) }-\lambda _k\left ( \mathbf{u }_0^{\left ( k \right )}  \right ) \right ) ^2=\left ( \mathbf{x} _t^{\left ( k \right ) }-2\lambda _k\left (w ^{\left ( k \right )}  \right )  \right )^2 -2\lambda _k^2\left (w ^{\left ( k \right )}   \right )^2.
\end{equation}
For any $0\le t\le T-1$, we can calculate the corresponding part in \eqref{eq:i,k}  as
\begin{equation}\label{eq:k}
    \begin{aligned}
   &\int  \frac{1}{\sqrt{2\pi \lambda _k}\sigma  }\mathrm{exp}\left (-\frac{\left (  \mathbf{x} _t^{\left ( k \right ) }-\lambda _k\left ( \mathbf{u }_1^{\left ( k \right )}  \right ) \right ) ^2}{\lambda _k\sigma ^2 } +\frac{\left (  \mathbf{x} _t^{\left ( k \right ) }-\lambda _k\left ( \mathbf{u }_0^{\left ( k \right )}  \right ) \right ) ^2 }{2\lambda _k\sigma ^2 }  \right ) \mathrm{d} \left ( \mathbf{x} _t^{\left ( k \right ) } \right )\\
        =&\int  \frac{1}{\sqrt{2\pi \lambda _k}\sigma  }\mathrm{exp}\left ( -\frac{\left (  \mathbf{x} _t^{\left ( k \right ) }-2\lambda _k\left (w ^{\left ( k \right )}  \right ) \right )^2}{2\lambda _k\sigma ^2 } \right ) \mathrm{d} \left ( \mathbf{x} _t^{\left ( k \right ) } \right )  \mathrm{exp}\left ( \frac{\lambda _k\left ( w ^{\left ( k \right )}    \right )^2}{\sigma ^2}  \right )\\
        =&\mathrm{exp}\left ( \frac{\lambda _k\left (  w ^{\left ( k \right )}   \right )^2}{\sigma ^2}  \right ).
    \end{aligned}
\end{equation}
Plugging \eqref{eq:i} and \eqref{eq:k} into \eqref{eq:i,k}, we obtain $\mathbb{E} _{\mathbf{u}_0}\left (\frac{\mathrm{d} \mathbb{P}_{\mathbf{u}_1} }{\mathrm{d}  \mathbb{P}_{\mathbf{u}_0}}  \right ) ^2
        = \mathrm{exp}\left ( \frac{T\lambda _k\left ( w ^{\left ( k \right ) } \right )  ^2 }{\sigma ^2}  \right )$.
\end{proof}
\renewcommand{\proofname}{Proof}
Then we prove Lemma~\ref{low dimen}.
\renewcommand{\proofname}{Proof of Lemma~\ref{low dimen}}
\begin{proof}
The proof of Lemma~\ref{low dimen} follows the informational constrained risk inequality~\citep{brown1996constrained}.  For completeness, we provide a proof here.
For any  $\left ( \mathbf{u}_0,\mathbf{u}_1   \right )\in \mathsf{W} _{0,1}^{\left ( k \right ) } $, 
to establish a lower bound on the estimation error of $\hat{\mathbf{w} }_T^{\left ( k \right ) }$  with respect to $\mathbf{u}_1^{\left ( k \right ) }$ given that the estimation error of $\hat{\mathbf{w} }_T^{\left ( k \right ) }$  with respect to $\mathbf{u}_0^{\left ( k \right ) }$ is less than  $\frac{1}{4} \left ( w ^{\left ( k \right ) } \right )^2 $, we formulate this as the following constrained optimization problem, 
\begin{equation}\label{constrain opt}
\begin{aligned}
    \min_{\hat{\mathbf{w} }_T^{\left ( k \right ) }}\  \mathbb{E}_{\mathbf{u}_1}\left (  \hat{\mathbf{w} }_T^{\left ( k \right ) } -\mathbf{u}_1^{\left ( k \right ) } \right )^2, \quad
    \text{subject to}\  \mathbb{E}_{\mathbf{u}_0}\left ( \hat{\mathbf{w} }_T^{\left ( k \right ) }-\mathbf{u}_0^{\left ( k \right ) }   \right )^2\le \frac{1}{4} \left ( w ^{\left ( k \right ) } \right )^2  .
    \end{aligned}
\end{equation}
 The expectation in above equation can be written in the following form as
 \begin{equation}\nonumber
     \begin{aligned}
         &\mathbb{E}_{\mathbf{u}_1}\left (  \hat{\mathbf{w} }_T^{\left ( k \right ) }  -\mathbf{u}_1^{\left ( k \right ) } \right )^2=\int \left ( \hat{\mathbf{w} }_T^{\left ( k \right ) } -\mathbf{u}_1^{\left ( k \right ) } \right ) ^2\mathrm{d} \mathbb{P}_{\mathbf{u}_1}    ,\\
         &\mathbb{E}_{\mathbf{u}_0}\left (  \hat{\mathbf{w} }_T^{\left ( k \right ) }  -\mathbf{u}_0^{\left ( k \right ) } \right )^2=\int \left ( \hat{\mathbf{w} }_T^{\left ( k \right ) } -\mathbf{u}_0^{\left ( k \right ) } \right ) ^2\mathrm{d} \mathbb{P}_{\mathbf{u}_0}    .
     \end{aligned}
 \end{equation}
Applying Lagrange multipliers to the constrained optimization problem~\eqref{constrain opt}, the minimizer $\hat{\mathbf{w} }_T^{\left ( k \right ) }$ satisfies the following equation,
\begin{equation}\nonumber
  \begin{aligned}
   & 2\left ( \hat{\mathbf{w} }_T^{\left ( k \right ) }-\mathbf{u}_1^{\left ( k \right ) }  \right ) \mathrm{d} \mathbb{P}_{\mathbf{u}_1} +2\lambda \left (  \hat{\mathbf{w} }_T^{\left ( k \right ) }-\mathbf{u}_0^{\left ( k \right ) }  \right ) \mathrm{d} \mathbb{P}_{\mathbf{u}_0}.
 \end{aligned}
\end{equation}
Since $\mathbf{u}_0^{\left ( k \right ) }=0$ and $\mathbf{u}_1^{\left ( k \right ) }=w ^{\left ( k \right )}$, the above equation can be reduced to
\begin{equation}\label{eq:wtp}
     \left ( \hat{\mathbf{w} }_T^{\left ( k \right ) }- w ^{\left ( k \right ) }  \right ) \mathrm{d} \mathbb{P}_{\mathbf{u}_1} +\lambda  \left ( \hat{\mathbf{w} }_T \right )^{\left ( k \right ) } \mathrm{d} \mathbb{P}_{\mathbf{u}_0}=0.
\end{equation}
If $\lambda=0$, then $\hat{\mathbf{w} }_T^{\left ( k \right ) }=w ^{\left ( k \right ) }$, and
\begin{equation}\nonumber
    \left ( w ^{\left ( k \right ) }\right )   ^2=\mathbb{E} _{\mathbf{u}_0}\left ( \hat{\mathbf{w} }_T^{\left ( k \right ) } \right  )^2=\mathbb{E}_{\mathbf{u}_0}\left ( \hat{\mathbf{w} }_T^{\left ( k \right ) }-\mathbf{u}_0^{\left ( k \right ) }     \right ) ^2\le \frac{1}{4} \left (  w ^{\left ( k \right ) } \right )   ^2<\left ( w ^{\left ( k \right ) } \right )   ^2.
\end{equation}
The above equation leads to a contradiction. This implies that $\lambda>0$. By \eqref{eq:wtp},   $\hat{\mathbf{w} }_T^{\left ( k \right ) }$ can be expressed as
 \begin{equation}\label{atfk}
     \hat{\mathbf{w} }_T^{\left ( k \right ) }=\frac{ w ^{\left ( k \right ) }\mathrm{d} \mathbb{P}_{\mathbf{u}_1} }{\mathrm{d} \mathbb{P}_{\mathbf{u}_1}+\mathrm{d} \mathbb{P}_{\mathbf{u}_0}\lambda } .
 \end{equation}
Also by Lagrange multipliers, $ \hat{\mathbf{w} }_T^{\left ( k \right ) }$ satisfies  the following equation,
\begin{equation}\label{e0}
    \mathbb{E}_{\mathbf{u}_0}\left (  \hat{\mathbf{w} }_T^{\left ( k \right ) }-\mathbf{u}_0^{\left ( k \right ) }     \right ) ^2=\frac{1}{4}\left (w ^{\left ( k \right ) }\right )   ^2.
\end{equation}
Substituting the value of $w ^{\left ( k \right ) }$ given in \eqref{wk} into Lemma~\ref{chi}, we have
 \begin{equation}\label{p0p1}
     \mathbb{E} _{\mathbf{u}_0}\left (\frac{\mathrm{d} \mathbb{P}_{\mathbf{u}_1} }{\mathrm{d}  \mathbb{P}_{\mathbf{u}_0}}  \right ) ^2=\mathrm{exp}\left ( \frac{T\lambda _k\left ( w ^{\left ( k \right ) } \right )  ^2 }{\sigma ^2}  \right )  \le 1.
 \end{equation}
We can derive an upper bound of $\mathbb{E}_{\mathbf{u}_1} \left ( \frac{\mathrm{d} \mathbb{P}_{\mathbf{u}_1} }{\mathrm{d} \mathbb{P}_{\mathbf{u}_1}+\mathrm{d} \mathbb{P}_{\mathbf{u}_0}\lambda } \right )$ by the above equations as follows,
\begin{equation}\nonumber
    \begin{aligned}
     \frac{1}{4}\left (w ^{\left ( k \right ) }\right )   ^2\overset{\eqref{p0p1}}{\ge}  
         &\frac{1}{4}\left (w ^{\left ( k \right ) }\right )   ^2\mathbb{E}_{\mathbf{u}_0}\left (\frac{\mathrm{d} \mathbb{P}_{\mathbf{u}_1} }{\mathrm{d}  \mathbb{P}_{\mathbf{u}_0}}  \right ) ^2\\
\overset{\eqref{e0}}{=}   &\mathbb{E}_{\mathbf{u}_0}\left (  \hat{\mathbf{w} }_T^{\left ( k \right ) }-\mathbf{u}_0^{\left ( k \right ) }     \right ) ^2\mathbb{E}_{\mathbf{u}_0}\left (\frac{\mathrm{d} \mathbb{P}_{\mathbf{u}_1}}{\mathrm{d}  \mathbb{P}_{\mathbf{u}_0}}  \right ) ^2\\
\overset{\eqref{atfk}}{=}&\left (w ^{\left ( k \right ) }\right )   ^2\mathbb{E}_{\mathbf{u}_0}
\left ( \frac{ \mathrm{d} \mathbb{P}_{\mathbf{u}_1} }{\mathrm{d} \mathbb{P}_{\mathbf{u}_1}+\mathrm{d} \mathbb{P}_{\mathbf{u}_0}\lambda } \right )^2 \mathbb{E}_{\mathbf{u}_0}\left (\frac{\mathrm{d} \mathbb{P}_{\mathbf{u}_1}}{\mathrm{d}  \mathbb{P}_{\mathbf{u}_0}}  \right ) ^2\\
\overset{\left ( a \right ) }{\ge }  &\left (w ^{\left ( k \right ) }\right )   ^2\left ( \mathbb{E}_{\mathbf{u}_1}\left ( \frac{ \mathrm{d} \mathbb{P}_{\mathbf{u}_1} }{\mathrm{d} \mathbb{P}_{\mathbf{u}_1}+\mathrm{d} \mathbb{P}_{\mathbf{u}_0}\lambda } \right ) \right ) ^2,
    \end{aligned}
\end{equation}
where $\left ( a \right )$ holds by Cauchy-Schwarz inequality. 
This inequality implies 
\begin{equation}\label{a42}
    \mathbb{E}_{\mathbf{u}_1} \left ( \frac{\mathrm{d} \mathbb{P}_{\mathbf{u}_1} }{\mathrm{d} \mathbb{P}_{\mathbf{u}_1}+\mathrm{d} \mathbb{P}_{\mathbf{u}_0}\lambda } \right )\le \frac{1}{4}.
\end{equation}
Substituting the expression for $\hat{\mathbf{w} }_T^{\left ( k \right ) }$ from \eqref{atfk} into $\mathbb{E}_{\mathbf{u}_1}\left ( \hat{\mathbf{w} }_T^{\left ( k \right ) }-\mathbf{u}_1^{\left ( k \right )}   \right ) ^2$, we have 
\begin{equation}\nonumber
    \begin{aligned}
         \mathbb{E}_{\mathbf{u}_1}\left ( \hat{\mathbf{w} }_T^{\left ( k \right ) }-\mathbf{u}_1^{\left ( k \right )}   \right ) ^2
\overset{\eqref{atfk} }{=}&\mathbb{E}_{\mathbf{u}_1}\left ( \frac{ w ^{\left ( k \right ) }\mathrm{d} \mathbb{P}_{\mathbf{u}_1} }{\mathrm{d} \mathbb{P}_{\mathbf{u}_1}+\mathrm{d} \mathbb{P}_{\mathbf{u}_0}\lambda } -w ^{\left ( k \right ) } \right )^2 \\
&=\left ( w ^{\left ( k \right ) }  \right ) ^2\mathbb{E}_{\mathbf{u}_1}\left ( \frac{ \mathrm{d} \mathbb{P}_{\mathbf{u}_1} }{\mathrm{d} \mathbb{P}_{\mathbf{u}_1}+\mathrm{d} \mathbb{P}_{\mathbf{u}_0}\lambda } -1 \right )^2.
    \end{aligned}
\end{equation}
By Cauchy-Schwarz inequality and  $\mathbb{E}_{\mathbf{u}_1}\mathrm{d} \mathbb{P}_{\mathbf{u}_1}=1$, we have 
\begin{equation}\nonumber
    \begin{aligned}
    \mathbb{E}_{\mathbf{u}_1}\left ( \frac{ \mathrm{d} \mathbb{P}_{\mathbf{u}_1} }{\mathrm{d} \mathbb{P}_{\mathbf{u}_1}+\mathrm{d} \mathbb{P}_{\mathbf{u}_0}\lambda } -1 \right )^2
        =&\mathbb{E}_{\mathbf{u}_1}\left ( \frac{ \mathrm{d} \mathbb{P}_{\mathbf{u}_1} }{\mathrm{d} \mathbb{P}_{\mathbf{u}_1}+\mathrm{d} \mathbb{P}_{\mathbf{u}_0}\lambda } -1 \right )^2\mathbb{E}_{\mathbf{u}_1}\mathrm{d} \mathbb{P}_{\mathbf{u}_1}\\
\ge& \left ( \mathbb{E}_{\mathbf{u}_1}\left ( 1-\frac{ \mathrm{d} \mathbb{P}_{\mathbf{u}_1} }{\mathrm{d} \mathbb{P}_{\mathbf{u}_1}+\mathrm{d} \mathbb{P}_{\mathbf{u}_0}\lambda }  \right ) \right ) ^2\\
\overset{\eqref{a42}}{\ge }&\left ( 1-\frac{1}{4} \right ) ^2>\frac{1}{4}.
    \end{aligned}
\end{equation}
Therefore, we obtain that $\mathbb{E}_{\mathbf{u}_1}\left ( \hat{\mathbf{w} }_T^{\left ( k \right ) }-\mathbf{u}_1^{\left ( k \right )}   \right ) ^2>\frac{1}{4} \left ( w ^{\left ( k \right ) } \right ) ^2$.
\end{proof}
\renewcommand{\proofname}{Proof}
\section{Deferred Proof of Theorem~\ref{SHB UP}}
In this section, we present the key properties of the momentum matrix in Section~\ref{sec: pro mat} and Section~\ref{sec:PPM2}, which are essential for proving Theorem~\ref{SHB UP}. We then provide proofs of Lemma~\ref{SHB VZK} and Lemma~\ref{SHB VJS} in Section~\ref{SEC var}, 
 followed by the proofs of Lemma~\ref{SHB BZK} and Lemma~\ref{SHB BJS1} in Section~\ref{SEC bias}.
\subsection{Properties of Momentum Matrix}\label{sec: pro mat}
\setcounter{lemma}{2} 
\renewcommand{\thelemma}{B.\arabic{lemma}}
For $1\le j\le d$ and $0\le t\le T-1$, define the momentum matrix
\begin{equation}\nonumber
    \mathbf{A}_t^{\left ( j \right )} =\begin{bmatrix}
1+\beta -\eta _t\lambda _j  & -\beta \\
 1 &0
\end{bmatrix} \in \mathbb{R} ^{2\times 2}.
\end{equation}
In Lemma~\ref{SHB VZK} and Lemma~\ref{SHB BZK}, the $\mathrm{Variance}$ term and the $\mathrm{Bias}$ term can be bounded from above by the product of $\mathbf{A}_t^{\left ( j \right )} $. 
 As a consequence, the properties of $\mathbf{A}_t^{\left ( j \right )} $ play a crucial role in the proof of Theorem~\ref{SHB UP}.
 
The eigenvalues of $\mathbf{A}_t^{\left ( j \right )} $ are 
\begin{equation}\label{Aeigenvalue}
    \begin{aligned}
        &\gamma_{t,j,1}=\frac{1}{2}\left ( 1+\beta -\eta_t \lambda _j+\sqrt[]{\left ( 1+\beta -\eta_t \lambda _j \right )^2-4\beta  } \right ) ,\\
        &\gamma_{t,j,2}=\frac{1}{2}\left ( 1+\beta -\eta_t \lambda _j-\sqrt[]{\left ( 1+\beta -\eta_t \lambda _j \right )^2-4\beta  } \right ).
    \end{aligned}
\end{equation}
The following lemma can be derived from \eqref{Aeigenvalue}.
\begin{lemma}\label{discriminant}
    Let
    \begin{equation}\nonumber
        \Delta _{t,j}^2=\left ( 1+\beta -\eta_t \lambda _j \right )^2-4\beta=\left ( \left ( 1-\sqrt{\beta}  \right ) ^2-\eta_{t}\lambda _j \right ) \left ( \left ( 1+\sqrt{\beta}  \right ) ^2-\eta_{t}\lambda _j \right ).
    \end{equation}
  Then $\mathbf{A} _t^{\left ( j \right )}$ has real eigenvalues if $\Delta _{t,j}^2\ge 0$, and $\mathbf{A} _t^{\left ( j \right )}$ has complex eigenvalues if $\Delta _{t,j}^2< 0$.
\end{lemma}
We adopt Lemma~\ref{rho js} from \citet[Lemma~9]{pan2023accelerated} 
 which analyzes the spectral radius of the momentum matrix.
\begin{lemma}\label{rho js}
    Let $\rho \left ( \mathbf{A}_t^{\left ( j \right )}  \right ) $ be the spectral radius of $\mathbf{A}_t^{\left ( j \right )} $. If $\mathbf{A}_t^{\left ( j \right )} $ has real eigenvalues, then 
    \begin{equation}\nonumber
        \rho \left ( \mathbf{A}_t^{\left ( j \right )}  \right ) \le 1-\frac{\eta_t\lambda _j}{2} -\frac{\eta_t\lambda _j}{4\left ( 1-\sqrt[]{\beta }  \right ) } .
    \end{equation}
     If $\mathbf{A}_t^{\left ( j \right )} $ has complex eigenvalues, then 
     \begin{equation}\nonumber
         \rho \left ( \mathbf{A}_t^{\left ( j \right )}  \right ) =\sqrt[]{\beta } .
     \end{equation}
     For $0\le t\le T-1$, $1\le j\le d$, we have $\rho \left ( \mathbf{A}_t^{\left ( j \right )}  \right )<1$.
\end{lemma}
We adopt Lemma~\ref{Fvs2} and Lemma~\ref{dandiao} from  \citet[Lemma~6 and Lemma~7]{pan2023accelerated} with slightly modified, these two lemmas enable bounding $\left \|\mathbf{A}_{t+k}^{\left ( j \right )}\cdots \mathbf{A}_{t+1}^{\left ( j \right )} \right \| $ with $\rho \left ( \mathbf{A}_{t+k}^{\left ( j \right )} \right )$.

\begin{lemma} \label{Fvs2}
Given momentum matrices $\mathbf{A} _t^{\left ( j \right )}$ that are defined in (\ref{mom mat}), and $\beta \in \left [ 0,\left ( 1-\frac{A}{T}  \right )^2  \right ] $. For positive integer $k=1$, we have $ \left \| \left ( \mathbf{A} _t^{\left ( j \right )}  \right )^k \right \| _F\le 3$, and for any positive integer $k\ge 2$, we have
    \begin{equation}\nonumber
        \left \| \left ( \mathbf{A} _t^{\left ( j \right )}  \right )^k \right \| _F\le\min \left ( \frac{4}{\sqrt{\left | \left ( 1+\beta -\eta_t\lambda _j \right )^2-4\beta   \right | } }\left (  \rho\left ( \mathbf{A}_t^{\left ( j \right )}  \right ) \right )^{k-1},3k  \left (  \rho\left ( \mathbf{A}_t^{\left ( j \right )}  \right ) \right )^{k-2}  \right )  .
    \end{equation}
\end{lemma}
\begin{lemma}\label{dandiao}
    Given momentum matrices $\mathbf{A} _t^{\left ( j \right )}$ that are defined in (\ref{mom mat}), and $\beta \in \left [ 0,\left ( 1-\frac{A}{T}  \right )^2  \right ] $. If $\mathbf{A} _t^{\left ( j \right )}$ has real eigenvalues, which is equivalent to $\left ( 1+\beta-\eta_t\lambda _j \right ) ^2-4\beta \ge0$, we have
    \begin{equation}\nonumber
        \begin{aligned}
            \left \| \mathbf{A} _{t+k}^{\left ( j \right )}\cdots\mathbf{A} _{t+1}^{\left ( j \right )}  \right \| \le \left \| \mathbf{A} _{t+k}^{\left ( j \right )}\cdots\mathbf{A} _{t+1}^{\left ( j \right )}  \right \|_F
\le \left \|\left (  \mathbf{A} _{t+k}^{\left ( j \right )}  \right )^k \right \|_F.
        \end{aligned}
    \end{equation}
\end{lemma}

Lemma~\ref{rho js} and Lemma~\ref{Fvs2} imply the following lemma.
\begin{lemma}\label{rho complex}
    For $0\le \ell \le n-1$, and $j\le d$, if $\mathbf{A}_{K\ell}^{\left ( j \right )}$ has complex eigenvalues then 
    \begin{equation}\nonumber
        \left \| \left ( \mathbf{A}_{K\ell}^{\left ( j \right )}  \right ) ^K \right \| \le1.
    \end{equation}
\end{lemma}
\renewcommand{\proofname}{Proof of Lemma~\ref{rho complex}}
\begin{proof}
According to Lemma~\ref{rho js} and Lemma~\ref{Fvs2}, for $1\le j\le d$ and $0\le \ell\le n-1$, if $\mathbf{A}_{K\ell}^{\left ( j \right )}$ has complex eigenvalues in the $\ell$-th stage, then we have 
\begin{equation}\nonumber
    \left \| \left (  \mathbf{A}_{K\ell}^{\left ( j \right )} \right ) ^K \right \| \le 4K\left ( \rho \left (  \mathbf{A}_{K\ell}^{\left ( j \right )} \right ) \right ) ^K=4K\left ( \sqrt{\beta}  \right ) ^K\le \frac{4T}{\mathrm{log}_2T }\left ( 1-\frac{A}{T}  \right ) ^{\frac{T}{\mathrm{log}_2T }}\le1 .
\end{equation}
\end{proof}
\renewcommand{\proofname}{Proof}
\subsection{Properties of Product of Momentum Matrix}\label{sec:PPM2}
In the following, we provide an analysis of $\mathbf{A}_{T-1}^{\left ( j \right )}\cdots\mathbf{A}_{0}^{\left ( j \right )} \begin{bmatrix}
1 \\1

\end{bmatrix}$, under the condition that $\eta_0\lambda_j$ satisfies $\eta_0\lambda_j< \frac{A\left ( 1-\sqrt[]{\beta }  \right ) }{2T}$. This analysis is crucial for  bounding the $\mathrm{bias}$ term.
According to the step schedule method~\eqref{step schedule} and momentum matrix $ \mathbf{A}_t^{\left ( j \right )}$ defined in \eqref{mom mat}, 
\begin{equation}\nonumber
    \begin{aligned}
        \mathbf{A}_{T-1}^{\left ( j \right )}\cdots\mathbf{A}_{0}^{\left ( j \right )} \begin{bmatrix}
1 \\1

\end{bmatrix}
=\left ( \mathbf{A}_{\left ( n-1 \right )K }^{\left ( j \right )} \right )^{K} \cdots \left ( \mathbf{A}_{0 }^{\left ( j \right )} \right )^{K}\begin{bmatrix}
1 \\1

\end{bmatrix}.
    \end{aligned}
\end{equation}
Denote  $\mathbf{1}_2=\begin{bmatrix}
1 \\1

\end{bmatrix}$ and $\mathbf{e}_2=\begin{bmatrix}
0 \\1

\end{bmatrix}$. 
For $0\le s \le n-1$, we first consider the property of $\left ( \mathbf{A}_{sK }^{\left ( j \right )}\right )^{K}\mathbf{1}_2$ and decompose it over $\mathbf{1}_2$ and $\mathbf{e}_2$.
Let $\left ( \mathbf{A}_{sK }^{\left ( j \right )} \right )^{K}\mathbf{1}_2
=a_s\mathbf{1}_2
+b_s\mathbf{e}_2
$, then Lemma~\ref{a b} shows that $0\le a_s<1$ and $b_s$ is relatively small.
\begin{lemma}\label{a b}
    If $\eta_0\lambda_j< \frac{A\left ( 1-\sqrt[]{\beta }  \right ) }{2T}$,  for $ 0\le s\le n-1$, given $\left ( \mathbf{A}_{sK }^{\left ( j \right )} \right )^{K}\mathbf{1}_2
=a_s\mathbf{1}_2
+b_s\mathbf{e}_2
$, then we have $0\le a_s<1$ and $b_s\le 8\eta_0\lambda_j\frac{1}{1-\sqrt[]{\beta }}$.
\end{lemma}
\renewcommand{\proofname}{Proof of Lemma~\ref{a b}}
\begin{proof}
Since $\eta_0\lambda_j< \frac{A\left ( 1-\sqrt[]{\beta }  \right ) }{2T}$, according to the step schedule~\eqref{step schedule}, for $0\le t \le T-1$, we have
\begin{equation}\nonumber
    \begin{aligned}
        \eta_{t}\lambda _j&\le\eta_{0}\lambda _j< \frac{A\left ( 1-\sqrt[]{\beta }  \right ) }{2T} \le \frac{\left ( 1-\sqrt[]{\beta }  \right )^2 }{2} \le \frac{1}{2}.  
    \end{aligned}
\end{equation}
The above inequality implies that 
\begin{equation}\label{b91}
    \begin{aligned}
        \Delta _{t,j}^2&=\left ( \left ( 1-\sqrt{\beta}  \right ) ^2-\eta_{t}\lambda _j \right ) \left ( \left ( 1+\sqrt{\beta}  \right ) ^2-\eta_{t}\lambda _j \right ) \\
&>\frac{1}{4} \left ( 1-\sqrt[]{\beta }  \right )^2>0.
    \end{aligned}
\end{equation}
Therefore, for $0\le t \le T-1$, the eigenvalues of $\mathbf{A}_{t }^{\left ( j \right )} $ are real.
Let the eigenvalues of $\mathbf{A}_{sK }^{\left ( j \right )} $ are $\gamma _{s1}$ and $\gamma _{s2}$. By \eqref{Aeigenvalue}, $\gamma _{s1}$ and $\gamma _{s2}$ satisfy
\begin{equation}\label{eq:g1g2}
    \gamma _{s1}+\gamma _{s2}=1+\beta -\eta_{sK}\lambda _j>0,\ \gamma _{s1}\gamma _{s2}=\beta>0.
\end{equation}
The above inequality and Lemma~\ref{rho js} indicate $0<\gamma _{s2}<\gamma _{s1}<1$. Meanwhile, 
$\left ( \mathbf{A}_{sK }^{\left ( j \right )} \right )^{K}$ can be expressed by $\gamma _{s1}$ and $\gamma _{s2}$ as
\begin{equation}\label{AK}
    \left ( \mathbf{A}_{sK }^{\left ( j \right )} \right )^{K}=\begin{bmatrix}
 \frac{\gamma _{s1}^{K+1}-\gamma _{s2}^{K+1}}{\gamma _{s1}-\gamma _{s2}}  &-\beta \frac{\gamma _{s1}^{K}-\gamma _{s2}^{K}}{\gamma _{s1}-\gamma _{s2}} \\
 \frac{\gamma _{s1}^{K}-\gamma _{s2}^{K}}{\gamma _{s1}-\gamma _{s2}} &-\beta \frac{\gamma _{s1}^{K-1}-\gamma _{s2}^{K-1}}{\gamma _{s1}-\gamma _{s2}}
\end{bmatrix}.
\end{equation}
By the definition of $a_s$ and $b_s$, they can be calculated by
\begin{equation}\label{eq: ab}
    a_s= \left ( \left ( \mathbf{A}_{sK }^{\left ( j \right )} \right )^{K}\mathbf{1}_2   \right ) _1,\ b_s= \left ( \left ( \mathbf{A}_{sK }^{\left ( j \right )} \right )^{K}\mathbf{1}_2  \right ) _2-\left ( \left ( \mathbf{A}_{sK }^{\left ( j \right )} \right )^{K}\mathbf{1}_2  \right ) _1.
\end{equation}
Plugging \eqref{AK} into \eqref{eq: ab}, $a_s$ and $b_s$ can be expressed by $\gamma _{s1}$ and $\gamma _{s2}$. We express $a_s$ as
\begin{equation}\label{eq: as}
    \begin{aligned}
a_s=&\frac{\gamma _{s1}^{K+1}-\gamma _{s2}^{K+1}}{\gamma _{s1}-\gamma _{s2}}-\beta \frac{\gamma _{s1}^{K}-\gamma _{s2}^{K}}{\gamma _{s1}-\gamma _{s2}}\\
=&\left ( \sum_{i=0}^{K}\gamma _{s1}^{K-i}\gamma _{s2}^i  \right ) -\gamma _{s1}\gamma _{s2}\left ( \sum_{i=0}^{K-1}\gamma _{s1}^{K-i}\gamma _{s2}^i  \right )\\
=&\left ( 1-\gamma _{s2}\right ) \left ( \sum_{i=0}^{K}\gamma _{s1}^{K-i}\gamma _{s2}^i  \right ) +\gamma _{s2}^{K+1},
    \end{aligned}
\end{equation}
According to \eqref{eq: as} and $0<\gamma _{s2}<\gamma _{s1}<1$, we can derive $a_s>0$ and 
\begin{equation}\nonumber
\begin{aligned}
    a_s&=\left ( 1-\gamma _{s2}\right ) \left ( \sum_{i=0}^{K}\gamma _{s1}^{K-i}\gamma _{s2}^K  \right ) +\gamma _{s2}^{K+1}\\
   & <\left ( 1-\gamma _{s2}\right ) \left ( \sum_{i=0}^{K}\gamma _{s2}^i  \right ) +\gamma _{s2}^{K+1}
    =1-\gamma _{s2}^{K+1}+\gamma _{s2}^{K+1}
    =1.
     \end{aligned}
\end{equation}
Then we express $b_s$ as
\begin{equation}\nonumber
    \begin{aligned}
        b_s&=\frac{1}{\gamma _{s1}-\gamma _{s2}}\left ( -\left ( \gamma _{s1}^{K+1}-\gamma _{s2}^{K+1} \right ) + \beta\left (\gamma _{s1}^{K}-\gamma _{s2}^{K}  \right )+\left ( \gamma _{s1}^{K}-\gamma _{s2}^{K} \right ) -\beta\left (\gamma _{s1}^{K-1}-\gamma _{s2}^{K-1}  \right )\right )\\
&=-\left ( 1- \gamma _{s1}\right )\left ( 1- \gamma _{s2}\right )\frac{\gamma _{s1}^{K}-\gamma _{s2}^{K}}{\gamma _{s1}-\gamma _{s2}} \\
&\overset{\eqref{eq:g1g2}}{=} -\left ( 1-\left ( 1+\beta -\eta _{s K} \lambda _j\right ) +\beta  \right ) \frac{\gamma _{s1}^{K}-\gamma _{s2}^{K}}{\gamma _{s1}-\gamma _{s2}} 
=-\eta _{s K} \lambda _j\frac{\gamma _{s1}^{K}-\gamma _{s2}^{K}}{\gamma _{s1}-\gamma _{s2}} .
    \end{aligned}
\end{equation}
Plugging $0<\gamma _{s2}<\gamma _{s1}<1$ into the above equation, we can bound $ \left | b_s \right | $ as 
\begin{equation}\nonumber
    \begin{aligned}
       \left | b_s \right | 
 =\eta _{s K} \lambda _j\frac{\gamma _{s1}^{K}-\gamma _{s2}^{K}}{\gamma _{s1}-\gamma _{s2}}
\le 2\eta _0 \lambda _j\frac{1}{\gamma _{s1}-\gamma _{s2}}=2\eta _0 \lambda _j\frac{1}{\sqrt{\Delta _{sK,j}^2}}\overset{\eqref{b91} }{\le}  8\eta _0 \lambda _j\frac{1}{1-\sqrt[]{\beta } }.
    \end{aligned}
\end{equation}
\end{proof}
\renewcommand{\proofname}{Proof }
We then provide the property of $\left ( \mathbf{A}_{\left ( n-1 \right )K }^{\left ( j \right )} \right )^{K} \cdots \left ( \mathbf{A}_{0 }^{\left ( j \right )} \right )^{K}\mathbf{1}_2$ in Lemma~\ref{bias diedai}.
\begin{lemma}\label{bias diedai}
Given momentum matrices $\mathbf{A} _t^{\left ( j \right )}$ defined in (\ref{mom mat}), for $0\le s \le n-1$, let $\left ( \mathbf{A}_{sK }^{\left ( j \right )} \right )^{K}\mathbf{1}_2
=a_s\mathbf{1}_2
+b_s\mathbf{e}_2
$, and additionally define $a_{-1}=1$. Then $\mathbf{A}_{T-1}^{\left ( j \right )}\cdots\mathbf{A}_{0}^{\left ( j \right )}  \mathbf{1}_2$ can be represented as 
    \begin{equation}\nonumber
        \begin{aligned}
                & \mathbf{A}_{T-1}^{\left ( j \right )}\cdots\mathbf{A}_{0}^{\left ( j \right )}  \mathbf{1}_2
=\left ( \mathbf{A}_{\left ( n-1 \right )K }^{\left ( j \right )} \right )^{K} \cdots \left ( \mathbf{A}_{0 }^{\left ( j \right )} \right )^{K}\mathbf{1}_2\\
=&\left ( \prod_{i=0}^{n-1} a_{i} \right ) \mathbf{1}_2+
b_{n-1}\left ( \prod_{i=0}^{n-2} a_{i} \right ) \mathbf{e}_2+
\sum_{s=1}^{n-1}b_{s-1}\left ( \prod_{i=-1}^{s-2} a_{i} \right ) \left ( \mathbf{A}_{\left ( n-1 \right )K }^{\left ( j \right )} \right )^{K} \cdots \left ( \mathbf{A}_{sK }^{\left ( j \right )} \right )^{K}\mathbf{e}_2.
        \end{aligned}
    \end{equation}
\end{lemma}
\renewcommand{\proofname}{Proof of Lemma~\ref{bias diedai}}
\begin{proof}
    We prove the claimed result by induction. 
    
    For the case $t=1$, the claimed result holds by the following equation
    \begin{equation}\nonumber
        \begin{aligned}
            \left ( \mathbf{A}_{K }^{\left ( j \right )} \right )^{K}\left ( \mathbf{A}_{0 }^{\left ( j \right )} \right )^{K} \mathbf{1}_2
=a_{1} a_0\left ( \mathbf{A}_{K }^{\left ( j \right )} \right )^{K}\mathbf{1}_2+
b_{1}a_{0}\left ( \mathbf{A}_{K }^{\left ( j \right )} \right )^{K}\mathbf{e}_2+
b_{0}a_{-1}\left ( \mathbf{A}_{K }^{\left ( j \right )} \right )^{K}\left ( \mathbf{A}_{0}^{\left ( j \right )} \right )^{K} \mathbf{e}_2.
        \end{aligned}
    \end{equation}
  Assume that the following equality holds for the case $t-1$, 
    \begin{equation}\label{t-1}
    \begin{aligned}
       & \left ( \mathbf{A}_{\left ( t-1 \right )K }^{\left ( j \right )} \right )^{K} \cdots \left ( \mathbf{A}_{0 }^{\left ( j \right )} \right )^{K}\mathbf{1}_2\\
=&\left ( \prod_{i=0}^{t-1} a_{i} \right ) \mathbf{1}_2+
b_{t-1}\left ( \prod_{i=0}^{t-2} a_{i} \right ) \mathbf{e}_2+
\sum_{s=1}^{t-1}b_{s-1}\left ( \prod_{i=-1}^{s-2} a_{i} \right ) \left ( \mathbf{A}_{\left ( t-1 \right )K }^{\left ( j \right )} \right )^{K} \cdots \left ( \mathbf{A}_{sK }^{\left ( j \right )} \right )^{K}\mathbf{e}_2.
 \end{aligned}
    \end{equation}
Then consider the case $t$, by plugging \eqref{t-1} into case $t$, we have
\begin{equation}\nonumber
  \begin{aligned}
     & \left ( \mathbf{A}_{tK }^{\left ( j \right )} \right )^{K}\left ( \mathbf{A}_{\left ( t-1 \right )K }^{\left ( j \right )} \right )^{K} \cdots \left ( \mathbf{A}_{0 }^{\left ( j \right )} \right )^{K}\mathbf{1}_2\\
=&\left ( \prod_{i=0}^{t-1} a_{i} \right ) \left ( \mathbf{A}_{tK }^{\left ( j \right )} \right )^{K}\mathbf{1}_2+
b_{t-1}\left ( \prod_{i=0}^{t-2} a_{i} \right ) \left ( \mathbf{A}_{tK }^{\left ( j \right )} \right )^{K}\mathbf{e}_2\\
&+\sum_{s=1}^{t-1}b_{s-1}\left ( \prod_{i=-1}^{s-2} a_{i} \right ) \left ( \mathbf{A}_{tK }^{\left ( j \right )} \right )^{K}\left ( \mathbf{A}_{\left ( t-1 \right )K }^{\left ( j \right )} \right )^{K} \cdots \left ( \mathbf{A}_{sK }^{\left ( j \right )} \right )^{K}\mathbf{e}_2\\
=&\left ( \prod_{i=0}^{t} a_{i} \right ) \mathbf{1}_2+b_{t}\left ( \prod_{i=0}^{t-1} a_{i} \right ) \mathbf{e}_2
+\sum_{s=1}^{t}b_{s-1}\left ( \prod_{i=-1}^{s-2} a_{i} \right ) \left ( \mathbf{A}_{tK }^{\left ( j \right )} \right )^{K}\left ( \mathbf{A}_{\left ( t-1 \right )K }^{\left ( j \right )} \right )^{K} \cdots \left ( \mathbf{A}_{sK }^{\left ( j \right )} \right )^{K}\mathbf{e}_2.
  \end{aligned}
\end{equation}
Consequently, the equality holds for the case $t$ and hence the proof is completed by reduction.
\end{proof}
\renewcommand{\proofname}{Proof}

\subsection{Proof of Lemma~\ref{SHB VZK} and Lemma~\ref{SHB VJS}}\label{SEC var}
We now proceed to prove Lemma~\ref{SHB VZK}.
\renewcommand{\proofname}{Proof of Lemma~\ref{SHB VZK}}
\begin{proof}
Recall the definition in \eqref{eq:ww},
\begin{equation}\nonumber
    \begin{aligned}
        \tilde{\mathbf{H} } =\begin{bmatrix}
\mathbf{H}   & \mathbf{0} \\
  \mathbf{0} &\mathbf{0} 
\end{bmatrix},\quad \mathbf{A}_t  =\begin{bmatrix}
\left ( 1+\beta  \right )\mathbf{I}-\eta_t  \mathbf{H}   & -\beta\mathbf{I} \\
  \mathbf{I} &\mathbf{0} 
\end{bmatrix},\quad \mathbf{H} =\mathbf{V} \bm{\Sigma } \mathbf{V} ^{\top }.
    \end{aligned}
\end{equation}
Define the orthogonal matrices $\tilde{\mathbf{V} }$ and $\bm{\Pi }$ as follows, 
\begin{equation}\nonumber
    \begin{aligned}
         \tilde{\mathbf{V} } =\begin{bmatrix}
\mathbf{V}   & \mathbf{0} \\
  \mathbf{0} & \mathbf{V}
\end{bmatrix},\quad
\bm{\Pi } =\begin{bmatrix}
 \mathbf{e} _1 &  \mathbf{0}  &\cdots   &  \mathbf{e} _d &\mathbf{0} \\
 \mathbf{0}  & \mathbf{e} _1 &\cdots   & \mathbf{0} & \mathbf{e} _d
\end{bmatrix}, \quad
\bm{\Pi }^{\top }\bm{\Pi }=\tilde{\mathbf{V} } ^{\top }\tilde{\mathbf{V} }=\mathbf{I}_{2d\times 2d}.
    \end{aligned}
\end{equation}
 $\mathbf{e} _i\in \mathbb{R}^{d\times 1}$ is a standard unit vector, where the i-th component is 1 and all other components are 0. Therefore, we have $\bm{\Pi }^{\top }\bm{\Pi }=\tilde{\mathbf{V} } ^{\top }\tilde{\mathbf{V} }=\mathbf{I}_{2d\times 2d}$. 
 We show in the following that $\mathbf{A}_t $ and $ \tilde{\mathbf{H} }$ can be transformed into block diagonal matrices $\mathbf{A}_t ^{'}$ and $\tilde{\mathbf{H}}  ^{'}$ by a rearrangement of the coordinates via the eigenvalue decomposition of $\mathbf{H}$, 
\begin{equation}\label{eq: ort matrix}
    \begin{aligned}
 \mathbf{A}_t ^{'}\equiv \begin{bmatrix}
\mathbf{A}_t^{\left ( 1 \right )}   &  & \\
  & \ddots  & \\
  &  &\mathbf{A}_t^{\left ( d \right )}
\end{bmatrix}
=\bm{\Pi }^{\top }\tilde{\mathbf{V}} ^{\top } \mathbf{A}_t \tilde{\mathbf{V}}\bm{\Pi },\quad 
&\tilde{\mathbf{H}}  ^{'}\equiv\begin{bmatrix}
 \lambda _1 &  0&  &  & \\
 0 & 0 &  &  & \\
  &  & \ddots  &  & \\
  &  &  & \lambda _d & 0\\
  &  &  & 0 &0
\end{bmatrix}=\bm{\Pi }^{\top }\tilde{\mathbf{V}} ^{\top } \tilde{\mathbf{H}}  \tilde{\mathbf{V}}\bm{\Pi }.
    \end{aligned}
\end{equation}
By the recursive of $\mathbf{C}_t$ in \eqref{BC}, $\mathbf{C} _{T}$ can be unrolled until the start as 
\begin{equation}\nonumber
    \begin{aligned}
        \mathbf{C} _{T}=&\mathbf{A}_{T-1} \mathbf{C} _{T-1}\mathbf{A} _{T-1}^{\top } +\eta_{T-1}^2 \mathbb{E} \bm{ \tilde{\zeta }}_{T-1}\left ( \bm{ \tilde{\zeta } }_{T-1} \right ) ^{\top} \\
=&\sum_{t=0}^{T-1}\eta_t^2\mathbf{A}_{T-1}\cdots \mathbf{A}_{t+1}\mathbb{E}  \tilde{\zeta }_t \left (  \tilde{\zeta }_t \right )^{\top}   \mathbf{A}_{t+1}^{\top }\cdots \mathbf{A}_{T-1}^{\top }.
    \end{aligned}
\end{equation}
Substituting Assmuption~\ref{noise} into the above equation, we can bound $\mathbf{C} _{T}$ from above as
\begin{equation}\nonumber
    \begin{aligned}
        \mathbf{C} _{T}
&\overset{\eqref{noise}}{\preceq}  \sigma ^2\sum_{t=0}^{T-1}\eta_t^2\mathbf{A}_{T-1}\cdots \mathbf{A}_{t+1}  \tilde{\mathbf{H} }  \mathbf{A}_{t+1}^{\top }\cdots \mathbf{A}_{T-1}^{\top }\\
&=\sigma ^2\sum_{t=0}^{T-1}\eta_t^2\tilde{\mathbf{V} }\bm{\Pi }\mathbf{A}_{T-1}^{'} \cdots \mathbf{A}_{t+1}^{'}\tilde{\mathbf{H}}  ^{'}\left ( \mathbf{A}_{t+1}^{'} \right ) ^{\top }\cdots \left ( \mathbf{A}_{T-1}^{'} \right ) ^{\top }\bm{\Pi }^{\top }\tilde{\mathbf{V} }^{\top }.
    \end{aligned}
\end{equation}
Consequently, the $\mathrm{Variance}$ term can be bounded by 
\begin{equation}\nonumber
    \begin{aligned}
        \left \langle \tilde{\mathbf{H}} ,\mathbf{C}_T   \right \rangle 
&=\sigma ^2\sum_{t=0}^{T-1}\eta_t^2\left \langle \tilde{\mathbf{H}} ,\tilde{\mathbf{V} }\bm{\Pi }\mathbf{A}_{T-1}^{'} \cdots \mathbf{A}_{t+1}^{'}\tilde{\mathbf{H}}  ^{'}\left ( \mathbf{A}_{t+1}^{'} \right ) ^{\top }\cdots \left ( \mathbf{A}_{T-1}^{'} \right ) ^{\top }\bm{\Pi }^{\top }\tilde{\mathbf{V} }^{\top }   \right \rangle\\
&=  \sigma ^2\sum_{t=0}^{T-1}\eta_t^2\mathrm{tr} \left (\tilde{\mathbf{H}}  ^{'} \mathbf{A}_{T-1}^{'} \cdots \mathbf{A}_{t+1}^{'}\tilde{\mathbf{H}}  ^{'}\left ( \mathbf{A}_{t+1}^{'} \right ) ^{\top }\cdots \left ( \mathbf{A}_{T-1}^{'} \right ) ^{\top }\right )\\
&= \sigma ^2\sum _{j=1}^{d}\lambda _j^2\sum_{t=0}^{T-1}\eta_t^2\mathrm{tr}\left (\begin{bmatrix}
1  & 0\\
0  &0
\end{bmatrix}  \mathbf{A}_{T-1}^{\left ( j \right )} \cdots \mathbf{A}_{t+1}^{\left ( j \right )}\begin{bmatrix}
 1 &0 \\
0  &0
\end{bmatrix}\left ( \mathbf{A}_{t+1}^{\left ( j \right )} \right ) ^{\top }\cdots \left ( \mathbf{A}_{T-1}^{\left ( j \right )} \right ) ^{\top }\right )\\
&=\sigma ^2\sum _{j=1}^{d}\lambda _j^2\sum_{t=0}^{T-1}\eta_t^2\left ( \mathbf{A}_{T-1}^{\left ( j \right )} \cdots \mathbf{A}_{t+1}^{\left ( j \right )}\begin{bmatrix}
1 \\0

\end{bmatrix} \right ) _1^2 \\
&\le \sigma ^2\sum _{j=1}^{d}\lambda _j^2\sum_{t=0}^{T-1}\eta_t^2\left \| \mathbf{A}_{T-1}^{\left ( j \right )} \cdots \mathbf{A}_{t+1}^{\left ( j \right )} \right \| ^2.
    \end{aligned}
\end{equation}
\end{proof}
\renewcommand{\proofname}{Proof}
Then we prove Lemma~\ref{SHB VJS}.
\renewcommand{\proofname}{Proof of Lemma~\ref{SHB VJS}}
\begin{proof}
For $j\le k^*$, we have $\lambda _j\eta_0\ge \frac{ 1-\sqrt{\beta}  }{T}$.
The analysis is divided into three parts based on the value of $\eta_{t+1}\lambda _j$. Assume that $t+1$ belongs to the $\ell$-th stage, that is, $K\left ( \ell-1 \right ) \le t+1\le K\ell-1$. 

 \noindent In the first part, $\eta_{t+1}\lambda _j$ satisfies $\eta_{t+1}\lambda _j>4\left ( 1-\sqrt{\beta }  \right ) ^2$. According to the step schedule~\eqref{step schedule}, we have
\begin{equation}\nonumber
\eta _T\lambda _j\le \frac{\eta_0\lambda _1}{T^2} <\frac{A^2}{T^2}\le  \eta _{t+1}\lambda _j.
\end{equation}
The above inequality shows that $t+1$ does not belong to the last stage, that is $\ell\le n-1$, and
\begin{equation}\nonumber
    \begin{aligned}
        \Delta _{K\ell,j}^2=\left ( \left ( 1-\sqrt{\beta}  \right ) ^2-\eta_{t+1}\lambda _j/4 \right ) \left ( \left ( 1+\sqrt{\beta}  \right ) ^2-\eta_{t+1}\lambda _j/4 \right )<0.
    \end{aligned}
\end{equation}
This implies the eigenvalues of $\mathbf{A}_{Kl}^{\left ( j \right )}$ are complex. Let $\ell^*$ be the last stage such that for $K\left ( \ell^*-1 \right ) \le s\le K\ell^*-1$ the eigenvalues of $\mathbf{A}_s^{\left ( j \right )}$ are complex. From this, we obtain that
\begin{equation}\nonumber
    \begin{aligned}
         &\lambda _j^2\eta_t^2\left \| \mathbf{A}_{T-1}^{\left ( j \right )} \cdots \mathbf{A}_{t+1}^{\left ( j \right )} \right \| ^2\\
\le &\lambda_1^2\eta _0^2\underbrace{ \left \| \mathbf{A}_{T-1}^{\left ( j \right )}\cdots  \mathbf{A}_{K\ell^*}^{\left ( j \right )} \right \| ^2}_{\text{real, Lemma~\ref{dandiao}} }\underbrace{ \left \|  \mathbf{A}_{K\left ( \ell^*-1 \right ) }^{\left ( j \right )} \right \| ^{2K}\cdots }_{\text{complex, Lemma~\ref{rho complex}} }\underbrace{ \left \|  \mathbf{A}_{K\ell }^{\left ( j \right )} \right \| ^{2K}}_{\text{complex, Lemma~\ref{Fvs2}} } \underbrace{\left \|  \mathbf{A}_{t+1 }^{\left ( j \right )} \right \| ^{2\left ( K-t \right ) }}_{\text{complex, Lemma~\ref{Fvs2}}  }\\
\le &9\lambda _1^2 \eta _0^2 \left ( T-K\ell^* \right ) ^2\left ( \rho \left ( \mathbf{A}_{T-1}^{\left ( j \right )} \right ) ^{2\left ( T-K\ell^* -2\right ) } \right )\cdot 9K^2\left ( \sqrt{\beta }  \right ) ^{2K-4}\cdot 9 \left ( K-t \right ) ^2\left ( \sqrt{\beta }  \right )^{2\left ( K-t -2\right ) }.
    \end{aligned}
\end{equation}

Combining the above inequalities and $\beta \le \left ( 1-\frac{A}{T}  \right )^2$, we can bound $\lambda _j^2\eta_t^2\left \| \mathbf{A}_{T-1}^{\left ( j \right )} \cdots \mathbf{A}_{t+1}^{\left ( j \right )}   \right \| ^2$ from above as follows
\begin{equation}\label{var1}
\begin{aligned}
     &\lambda _j^2\eta_t^2\left \| \mathbf{A}_{T-1}^{\left ( j \right )} \cdots \mathbf{A}_{t+1}^{\left ( j \right )}   \right \| ^2
\le  3^{6}\lambda _1^2\eta _0^2T^6\left ( 1-\frac{A}{T}  \right ) ^{\frac{T}{\mathrm{log}_2T } }
\le  3^{6}\lambda _1^2\eta _0^2\frac{1}{T^2}.
\end{aligned}
\end{equation}

 \noindent In the second part,  $\eta_{t+1}\lambda _j$ satisfies $\frac{A\left ( 1-\sqrt{\beta} \right ) }{2T}\le \eta_{t+1}\lambda_j\le  4\left ( 1-\sqrt{\beta} \right )^2 $. 
According to the step schedule~\eqref{step schedule}, we have
\begin{equation}\nonumber
\eta _T\lambda _j\le \frac{\eta_0\lambda _1}{T^2} <\frac{A^2}{4T^2}\le  \frac{A\left ( 1-\sqrt{\beta }  \right ) }{4T}< \eta _{t+1}\lambda _j.
\end{equation}
The above inequality shows that $t+1$ does not belong to the last stage, that is, $\ell\le n-1$. For $K\ell\le s \le T-1$, 
\begin{equation}\nonumber
    \begin{aligned}
        \Delta _{s,j}^2&=\left ( \left ( 1-\sqrt{\beta}  \right ) ^2-\eta_{s}\lambda _j \right ) \left ( \left ( 1+\sqrt{\beta}  \right ) ^2-\eta_{s}\lambda _j \right )\\
&\ge\left ( \left ( 1-\sqrt{\beta}  \right ) ^2-\eta_{t+1}\lambda _j/4 \right ) \left ( \left ( 1+\sqrt{\beta}  \right ) ^2-\eta_{t+1}\lambda _j/4 \right )
\ge 0.
    \end{aligned}
\end{equation}
This implies the eigenvalues of $\mathbf{A}_s^{\left ( j \right )}$ are real for $K\ell\le s \le T-1$. Form this, we have
\begin{equation}\nonumber
    \begin{aligned}
         &\lambda _j^2\eta_t^2\left \| \mathbf{A}_{T-1}^{\left ( j \right )} \cdots \mathbf{A}_{t+1}^{\left ( j \right )} \right \| ^2\\
\le &\lambda _1^2\eta _0^2\underbrace{ \left \| \mathbf{A}_{T-1}^{\left ( j \right )}\cdots  \mathbf{A}_{K\left ( \ell+1 \right ) }^{\left ( j \right )} \right \| ^2}_{\text{real, Lemma~\ref{dandiao}} }\underbrace{ \left \|  \mathbf{A}_{K\ell }^{\left ( j \right )} \right \| ^{2K}}_{\text{real, Lemma~\ref{Fvs2}} } \underbrace{\left \|  \mathbf{A}_{t+1 }^{\left ( j \right )} \right \| ^{2\left ( K-t \right ) }}_{\text{arbitrary, Lemma~\ref{Fvs2}}  }\\
\le &9\left ( T-K\left ( \ell+1 \right )  \right ) ^2\left ( \rho \left ( \mathbf{A}_{T-1}^{\left ( j \right )} \right ) ^{2T-K\left ( \ell+1 \right )-4 } \right )\\
&\cdot 9 K^2\left ( \rho \left ( \mathbf{A}_{K\ell }^{\left ( j \right )} \right )  \right ) ^{2K-4}\cdot 9\left ( K-t \right ) ^2\left ( \rho \left (  \mathbf{A}_{t+1 }^{\left ( j \right )}\right )  \right )^{2\left ( K-t-2 \right ) }.
    \end{aligned}
\end{equation}
According to $ \eta_{t+1}\lambda_j\ge \frac{A\left ( 1-\sqrt{\beta} \right ) }{2T}$ and Lemma~\ref{rho js}, we have the following inequalities,
\begin{equation}\nonumber
\begin{aligned}
    &\rho \left ( \mathbf{A}_{T-1}^{\left ( j \right )} \right ) <1,\quad \rho \left (  \mathbf{A}_{t+1 }^{\left ( j \right )}\right )<1,\\
    &\rho \left ( \mathbf{A}_{K\ell }^{\left ( j \right )} \right )\le 1-\frac{\eta _{K\ell}\lambda _j}{4\left ( 1-\sqrt[]{\beta }  \right ) }=1-\frac{\eta _{t+1}\lambda _j}{16\left ( 1-\sqrt[]{\beta }  \right ) }\le 1-\frac{A}{32T}.
\end{aligned}
\end{equation}
Combining the above inequalities, we can bound $\lambda _j^2\eta_t^2\left \| \mathbf{A}_{T-1}^{\left ( j \right )} \cdots \mathbf{A}_{t+1}^{\left ( j \right )}   \right \| ^2$ from above as follows
\begin{equation}\label{var2}
    \begin{aligned}
        \lambda _j^2\eta_t^2\left \| \mathbf{A}_{T-1}^{\left ( j \right )} \cdots \mathbf{A}_{t+1}^{\left ( j \right )}   \right \| ^2
        \le 3^{6}  \lambda _1^2\eta _0^2T^6\left ( 1-\frac{A}{32T }  \right ) ^{\frac{T}{\mathrm{log}_2T } }
\le 3^{6} \lambda _1^2\eta _0^2\frac{1}{T^2}.
    \end{aligned}
\end{equation}

 \noindent In the third part,  $\eta_{t+1}\lambda _j$ satisfies $ \eta_{t+1}\lambda_j< \frac{A\left ( 1-\sqrt{\beta} \right ) }{2T}$.
According to the step schedule~\eqref{step schedule}, for $T\le s\le t+1$, we have
\begin{equation}\nonumber
    \begin{aligned}
        \eta_s\lambda _j\le\eta_{t+1}\lambda _j\le \frac{A\left ( 1-\sqrt[]{\beta }  \right ) }{2T} \le  \frac{\left ( 1-\sqrt[]{\beta }  \right )^2 }{2} \le \frac{1}{2}. 
    \end{aligned}
\end{equation}
The above inequality shows that
\begin{equation}\label{b112}
    \Delta _{s,j}^2=\left ( \left ( 1-\sqrt{\beta}  \right ) ^2-\eta_s\lambda _j \right ) \left ( \left ( 1+\sqrt{\beta}  \right ) ^2-\eta_s\lambda _j \right ) 
>\frac{1}{4} \left ( 1-\sqrt[]{\beta }  \right )^2>0.
\end{equation}
This implies the eigenvalues of $\mathbf{A}_s^{\left ( j \right )}$ are real for $t+1\le s \le T-1$. By   Lemma~\ref{dandiao}, we have 
\begin{equation}\label{b1122}
    \begin{aligned}
           \lambda _j^2\eta_t^2\underbrace{ \left \| \mathbf{A}_{T-1}^{\left ( j \right )} \cdots \mathbf{A}_{t+1}^{\left ( j \right )} \right \| ^2}_{\text{real, Lemma~\ref{dandiao}} }
      \le   \lambda _j^2\eta_t^2 \frac{16}{\Delta _{T,j}^2} \left ( \rho \left ( \mathbf{A}_T^{\left ( j \right )}  \right ) \right ) ^{2\left ( T-t -1\right ) }.
    \end{aligned}
\end{equation}
In this part, we have $ \eta_{t+1}\lambda_j< \frac{A\left ( 1-\sqrt{\beta} \right ) }{2T}$, which implies $ \eta_{t}\lambda_j< \frac{2A\left ( 1-\sqrt{\beta} \right ) }{T}$. By \eqref{b112}, we  obtain  $\frac{1}{\Delta _{T,j}^2}<\frac{4}{ \left ( 1-\sqrt[]{\beta }  \right )^2}$. According to Lemma~\ref{rho js}, the spectral radius of momentum matrix satisfies $ \rho \left ( \mathbf{A}_T^{\left ( j \right )}  \right )<1 $. Using \eqref{b1122}, we can bound $\lambda _j^2\eta_t^2\left \| \mathbf{A}_{T-1}^{\left ( j \right )} \cdots \mathbf{A}_{t+1}^{\left ( j \right )}   \right \| ^2$ from above as follows
\begin{equation}\label{var3}
    \begin{aligned}
        \lambda _j^2\eta_t^2\left \|\mathbf{A}_{T-1}^{\left ( j \right )} \cdots \mathbf{A}_{t+1}^{\left ( j \right )} \right \| ^2
\le  \left ( \frac{2A\left ( 1-\sqrt{\beta} \right ) }{T} \right ) ^2 \frac{64}{\left ( 1-\sqrt[]{\beta }  \right )^2}
=\frac{256A^2}{T^2} .
    \end{aligned}
\end{equation}
 
 \noindent According to  \eqref{var1}, \eqref{var2}, and \eqref{var3}, for $j\le k^*$ and $0\le t\le T-1$, we have 
\begin{equation}\nonumber
    \begin{aligned}
         \lambda _j^2\eta_t^2\left \|\mathbf{A}_{T-1}^{\left ( j \right )} \cdots \mathbf{A}_{t+1}^{\left ( j \right )} \right \| ^2\le \frac{\left ( 3^{6}\lambda _1^2\eta_0^2+256A^2 \right ) }{T^2} . 
    \end{aligned}
\end{equation}
Consequently, summing the above equation from we can obtain that 
\begin{equation}\nonumber
    \lambda _j^2\sum_{t=0}^{T-1}\eta_t^2\left \| \mathbf{A}_{T-1}^{\left ( j \right )} \cdots \mathbf{A}_{t+1}^{\left ( j \right )}  \right \| ^2\le \frac{\left ( 3^{6}\lambda _1^2\eta_0^2+256A^2 \right ) }{T}.
\end{equation}

For $j>k^{*}$, we have $\lambda_j\eta_0<\frac{1-\sqrt{\beta}}{T}$. According to the step schedule~\eqref{step schedule}, for $T\le s\le t+1$, we have
\begin{equation}
    \begin{aligned}
        \eta_s\lambda _j\le\eta_{0}\lambda _j\le \frac{\eta_0\left ( 1-\sqrt[]{\beta }  \right ) }{2T} \le \frac{\left ( 1-\sqrt[]{\beta }  \right )^2 }{2} \le \frac{1}{2}.
    \end{aligned}
\end{equation}
The above inequality shows that
\begin{equation}\label{b113}
    \begin{aligned}
\Delta _{s,j}^2=\left ( \left ( 1-\sqrt{\beta}  \right ) ^2-\eta_s\lambda _j \right ) \left ( \left ( 1+\sqrt{\beta}  \right ) ^2-\eta_s\lambda _j \right ) 
>\frac{1}{4} \left ( 1-\sqrt[]{\beta }  \right )^2>0.
    \end{aligned}
\end{equation}
This implies that the eigenvalues of $\mathbf{A}_s^{\left ( j \right )}$ are real for $t+1\le s \le T-1$, then 
\begin{equation}\nonumber
    \begin{aligned}
        \lambda _j^2\eta_t^2\underbrace{ \left \| \mathbf{A}_{T-1}^{\left ( j \right )} \cdots \mathbf{A}_{t+1}^{\left ( j \right )} \right \| ^2}_{\text{real, Lemma~\ref{dandiao}} }
\le  &\lambda _j^2\eta_0^2\frac{16}{\Delta _{T-1,j}^2} \left ( \rho \left ( \mathbf{A}_{T-1}^{\left ( j \right )}  \right )  \right )^{2\left ( T-t-1 \right )}\\
\overset{\left ( a  \right ) }{\le}   &64\eta_0^2\lambda _j^2\frac{1}{\left ( 1-\sqrt[]{\beta }  \right )^2 } ,
    \end{aligned}
\end{equation}
where  $\left ( a  \right )$ holds for \eqref{b113} and Lemma~\ref{rho js}.
Consequently, for $j>k^*$, summing the above equation from we can obtain that
\begin{equation}\nonumber
    \begin{aligned}
         \lambda _j^2\sum_{t=0}^{T-1}\eta_t^2\left \| \mathbf{A}_{T-1}^{\left ( j \right )} \cdots \mathbf{A}_{t+1}^{\left ( j \right )} \right \| ^2
         \le \sum_{t=0}^{T-1}64\eta_0^2\lambda _j^2\frac{1}{\left ( 1-\sqrt[]{\beta }  \right )^2 }
         =64\eta_0^2\lambda _j^2T\frac{1}{\left ( 1-\sqrt[]{\beta }  \right )^2 }.
    \end{aligned}
\end{equation}

\end{proof}
\renewcommand{\proofname}{Proof}

\subsection{Proof of Lemma~\ref{SHB BZK} and Lemma~\ref{SHB BJS1}}\label{SEC bias}
We now proceed to prove Lemma~\ref{SHB BZK}.
\renewcommand{\proofname}{Proof of Lemma~\ref{SHB BZK}}
\begin{proof}
    We continue to use the definition in Lemma~\ref{SHB VZK}, and denote 
    \begin{equation}
        \tilde{\mathbf{w} } _0^{'}=\bm{\Pi }^{\top }\tilde{\mathbf{V} } ^{\top }\tilde{\mathbf{w} } _0 =-\begin{bmatrix}
\mathbf{w} _*^{\left ( 1 \right ) }  & \mathbf{w} _*^{\left ( 1 \right ) } & \cdots  & \mathbf{w} _*^{\left ( d \right ) } &\mathbf{w} _*^{\left ( d \right ) }
\end{bmatrix}^{\top }.
    \end{equation}
 By the recursive of $\mathbf{B}_t$, $\mathbf{B}_t$ can be expressed in terms of $\tilde{\mathbf{V} } \bm{\Pi }$ and $\mathbf{A}_{t}^{'}$ as
    \begin{equation}\nonumber
        \begin{aligned}
            \mathbf{B} _T&=\mathbf{A }_{T-1} \mathbf{B} _{T-1}\mathbf{A }_{T-1}^{\top }\\
&=\mathbf{A }_{T-1}\cdots \mathbf{A }_{0}\tilde{\mathbf{w} } _0\left (  \tilde{\mathbf{w} } _0 \right ) ^{\top}\mathbf{A }_{0}^{\top }\cdots\mathbf{A }_{T-1}^{\top }\\
&=\tilde{\mathbf{V} } \bm{\Pi } \mathbf{A}_{T-1}^{'} \cdots \mathbf{A}_{0}^{'}\tilde{\mathbf{w} } _0^{'}\left ( \tilde{\mathbf{w} } _0^{'} \right ) ^{\top}\left ( \mathbf{A}_{0}^{'} \right )^{\top }  \cdots \left ( \mathbf{A}_{T-1}^{'} \right )^{\top } \bm{\Pi }^{\top }\tilde{\mathbf{V} }^{\top }.
        \end{aligned}
    \end{equation}
The $\mathrm{Bias}$ term can be bounded from above by 
\begin{equation}\nonumber
    \begin{aligned}
       \left \langle \tilde{\mathbf{H} } ,\mathbf{B}_t  \right \rangle
&= \left \langle \tilde{\mathbf{H} } ,\tilde{\mathbf{V} } \bm{\Pi } \mathbf{A}_{T-1}^{'} \cdots \mathbf{A}_{0}^{'} \tilde{\mathbf{w} } _0^{'}\left ( \tilde{\mathbf{w} } _0^{'} \right ) ^{\top}\left ( \mathbf{A}_{0}^{'} \right )^{\top }  \cdots \left ( \mathbf{A}_{T-1}^{'} \right )^{\top } \bm{\Pi }^{\top }\tilde{\mathbf{V} }^{\top }  \right \rangle\\
&=\mathrm{tr} \left( \tilde{\mathbf{H} }^{'} \mathbf{A}_{T-1}^{'} \cdots \mathbf{A}_{0}^{'} \tilde{\mathbf{w} } _0^{'}\left ( \tilde{\mathbf{w} } _0^{'} \right ) ^{\top}\left ( \mathbf{A}_{0}^{'} \right )^{\top }  \cdots \left ( \mathbf{A}_{T-1}^{'} \right )^{\top }\right ) \\
&=\sum _{j=1}^d\lambda _j\mathrm{tr} \left ( \begin{bmatrix}
1  & 0\\
0  &0
\end{bmatrix} \mathbf{A}_{T-1}^{\left ( j \right )}\cdots \mathbf{A}_{0}^j \begin{bmatrix}
  1&1 \\
  1&1
\end{bmatrix} \mathbf{A}_{0}^{\left ( j \right )}\cdots \mathbf{A}_{T-1}^{\left ( j \right )}\right ) \left ( \mathbf{w} _*^{\left ( j \right ) } \right ) ^2\\
&=\sum_{j=1}^{d}\lambda _j\left ( \mathbf{A}_{T-1}^{\left ( j \right )}\cdots \mathbf{A}_{0}^{\left ( j \right )} \begin{bmatrix}
1 \\1

\end{bmatrix}  \right ) ^2_1\left (  \mathbf{w} _*^{\left ( j \right ) } \right ) ^2.
    \end{aligned}
\end{equation}
\end{proof}
\renewcommand{\proofname}{Proof}
Then we prove Lemma~\ref{SHB BJS1}.
\renewcommand{\proofname}{Proof of Lemma~\ref{SHB BJS1}}
\begin{proof}
For $1\le j \le k^*$, we have $\eta_0\lambda _j\ge \frac{1-\sqrt{\beta}}{T}$. The analysis is divided into three parts based on the value of $\eta_{0}\lambda _j$.

 \noindent In the first part, $\eta_0\lambda_j$ satisfies $ \eta_0\lambda_j>\left ( 1-\sqrt{\beta }  \right ) ^2$. For $0\le s\le K-1$, according to the step schedule~\eqref{step schedule}, we have
\begin{equation}\nonumber
    \begin{aligned}
        \Delta _{s,j}^2
&=\left ( \left ( 1-\sqrt{\beta}  \right ) ^2-\eta_s\lambda _j \right ) \left ( \left ( 1+\sqrt{\beta}  \right ) ^2-\eta_s\lambda _j \right )\\
&=\left ( \left ( 1-\sqrt{\beta}  \right ) ^2-\eta_0\lambda _j \right ) \left ( \left ( 1+\sqrt{\beta}  \right ) ^2-\eta_0\lambda _j \right )<0.
    \end{aligned}
\end{equation}
This implies for $0\le s\le K-1$ the eigenvalues of $\mathbf{A}_s^{\left ( j \right )}$ are complex.
Let $\ell$ be the last stage such that for $K\left ( \ell-1 \right ) \le s\le K\ell -1$ the eigenvalues of $\mathbf{A}_s^{\left ( j \right )}$ are complex. Then we have
\begin{equation}\label{bi1}
    \begin{aligned}
         &\lambda _j\left ( \mathbf{A}_{T-1}^{\left ( j \right )}\cdots \mathbf{A}_{0}^{\left ( j \right )}  \begin{bmatrix}
1 \\1

\end{bmatrix}  \right ) ^2_1
\le 2\lambda _1\left \| \mathbf{A}_{T-1}^{\left ( j \right )}\cdots \mathbf{A}_{0}^{\left ( j \right )} \right \| ^2\\
\le &2\lambda _1\underbrace{\left \| \mathbf{A}_{T-1}^{\left ( j \right )}\cdots \mathbf{A}_{K\ell}^{\left ( j \right )}  \right \| ^2}_{\text{real, Lemma~\ref{dandiao} } }
\underbrace{\left \|  \mathbf{A}_{\left ( K-1 \right )\ell }^{\left ( j \right )} \right \|^{2K}\cdots}_{\text{complex, Lemma~\ref{rho complex}} } \underbrace{\left \| \mathbf{A}_0^{\left ( j \right )}  \right \| ^{2K}}_{\text{complex, Lemma~\ref{Fvs2} } }  \\
\le & 2\lambda _1 \cdot  9\left ( T-K\ell \right )^2\left ( \rho \left ( \mathbf{A}_{T-1}^{\left ( j \right )}  \right )  \right )  ^{2\left ( T-K\ell-2 \right )} \cdot  9K^2\left ( \rho \left ( \mathbf{A}_{0}^{\left ( j \right )}  \right )  \right )  ^{2K-4} \\
\overset{\left ( a \right )  }{\le }  &  182\lambda _1T^4\left (\sqrt{\beta } \right ) ^{\frac{2T}{\mathrm{log}_2T }-4 }
\le  \frac{182\lambda _1}{T^2},
    \end{aligned}
\end{equation}
where $\left ( a \right )$ holds for Lemma~\ref{rho js}.

 \noindent In the second part, $\eta_0\lambda_j$ satisfies $ \frac{A\left ( 1-\sqrt[]{\beta }  \right ) }{2T} \le \eta_0\lambda _j\le \left ( 1-\sqrt[]{\beta }  \right )^2  $. According to the step schedule~\eqref{step schedule}, for $0\le s\le T-1$, we have
\begin{equation}\nonumber
    \begin{aligned}
        \Delta _{s,j}^2
&=\left ( \left ( 1-\sqrt{\beta}  \right ) ^2-\eta_s\lambda _j \right ) \left ( \left ( 1+\sqrt{\beta}  \right ) ^2-\eta_s\lambda _j \right )\\
&\ge\left ( \left ( 1-\sqrt{\beta}  \right ) ^2-\eta_0\lambda _j \right ) \left ( \left ( 1+\sqrt{\beta}  \right ) ^2-\eta_0\lambda _j \right )
\ge 0.
    \end{aligned}
\end{equation}
This implies for $0\le s\le T-1$, the eigenvalues of $\mathbf{A}_s^{\left ( j \right )}$ are real and we have 
\begin{equation}\nonumber
    \begin{aligned}
        &\lambda _j\left ( \mathbf{A}_{T-1}^{\left ( j \right )}\cdots \mathbf{A}_{0}^{\left ( j \right )} \begin{bmatrix}
1 \\1

\end{bmatrix}  \right ) ^2_1
\le 2\lambda _1\left \| \mathbf{A}_{T-1}^{\left ( j \right )}\cdots \mathbf{A}_{0}^{\left ( j \right )} \right \| ^2\\
\le &2\lambda _1\underbrace{\left \| \mathbf{A}_{T-1}^{\left ( j \right )}\cdots \mathbf{A}_{K}^{\left ( j \right )}  \right \| ^2}_{\text{real, Lemma~\ref{dandiao} } }
 \underbrace{\left \| \mathbf{A}_0^{\left ( j \right )}  \right \| ^{2K}}_{\text{real, Lemma~\ref{Fvs2} } }\\
 \le &2\lambda _1 \cdot 9\left ( T-K \right )^2\left ( \rho \left ( \mathbf{A}_{T-1}^{\left ( j \right )}  \right )  \right )  ^{2\left ( T-K -2\right )}\cdot 9K^2\left ( \rho \left ( \mathbf{A}_{0}^{\left ( j \right )}  \right )  \right )  ^{2K-4}.
    \end{aligned}
\end{equation}
\iffalse
For $0\le s\le T-1$, the eigenvalues of $\mathbf{A}_s^{\left ( j \right )}$ are real. By Lemma~\ref{Fvs2} and Lemma~\ref{dandiao}, 
\begin{equation}\nonumber
    \begin{aligned}
       \left \| \mathbf{A}_{T-1}^{\left ( j \right )}\cdots \mathbf{A}_{K}^{\left ( j \right )}  \right \| ^2 \left \| \mathbf{A}_0^{\left ( j \right )}  \right \| ^{2K}
       \le  16\left ( T-K \right )^2\left ( \rho \left ( \mathbf{A}_{T-1}^{\left ( j \right )}  \right )  \right )  ^{2\left ( T-K -2\right )}16K^2\left ( \rho \left ( \mathbf{A}_{0}^{\left ( j \right )}  \right )  \right )  ^{2K-4}.
    \end{aligned}
\end{equation}
\fi
Since $ \frac{A\left ( 1-\sqrt[]{\beta }  \right ) }{2T} \le \eta_0\lambda _j$ and Lemma~\ref{rho js}, the spectral radius of momentum matrix satisfies $\rho \left ( \mathbf{A}_{T-1}^{\left ( j \right )}  \right )<1$ and 
\begin{equation}\nonumber
    \begin{aligned}
        \rho \left ( \mathbf{A}_{0}^{\left ( j \right )}  \right ) \le 1-\frac{\eta_0\lambda _j}{4\left ( 1-\sqrt{\beta }  \right ) } \le 1-\frac{A}{8T } .
    \end{aligned}
\end{equation}
Combining the above inequalities, we can bound $\lambda _j\left ( \mathbf{A}_{T-1}^{\left ( j \right )}\cdots \mathbf{A}_{0}^{\left ( j \right )} \begin{bmatrix}
1 \\1

\end{bmatrix}  \right ) ^2_1$ from above as follows
\begin{equation}\label{bi2}
    \begin{aligned}
       \lambda _j\left ( \mathbf{A}_{T-1}^{\left ( j \right )}\cdots \mathbf{A}_{0}^{\left ( j \right )} \begin{bmatrix}
1 \\1

\end{bmatrix}  \right ) ^2_1
\le  182\lambda _1T^4\left ( 1-\frac{A}{8T }  \right ) ^{\frac{T}{\mathrm{log}_2T } }
\le 182\lambda _1 \frac{1}{T^2}.
    \end{aligned}
\end{equation}

 \noindent In the third part, $\eta_0\lambda_j$ satisfies $\eta_0\lambda_j<\frac{A\left ( 1-\sqrt[]{\beta }  \right ) }{2T}\le \frac{\left ( 1-\sqrt[]{\beta }  \right )^2}{2}$. According to the step schedule~\eqref{step schedule}, for $0\le s\le T-1$, we have
\begin{equation}\nonumber
    \begin{aligned}
        \Delta _{s,j}^2
&=\left ( \left ( 1-\sqrt{\beta}  \right ) ^2-\eta_s\lambda _j \right ) \left ( \left ( 1+\sqrt{\beta}  \right ) ^2-\eta_s\lambda _j \right )\\
&\ge\left ( \left ( 1-\sqrt{\beta}  \right ) ^2-\eta_0\lambda _j \right ) \left ( \left ( 1+\sqrt{\beta}  \right ) ^2-\eta_0\lambda _j \right )
\ge 0.
    \end{aligned}
\end{equation}
 This implies that for $0\le s\le T-1$, the eigenvalues of $\mathbf{A}_s^{\left ( j \right )}$ are real. We can futher bound $\eta_{T-1}\lambda _j$ as
\begin{equation}\label{b131}
    \begin{aligned}
        \eta_{T-1}\lambda _j&=\frac{\eta_0\lambda _j}{T^2} \le\frac{A\left ( 1-\sqrt[]{\beta }  \right ) }{T^3}\le \frac{ \left ( 1-\sqrt[]{\beta }   \right ) ^2}{2}\le \frac{1}{2}.
    \end{aligned}
\end{equation}
The above inequality implies that 
\begin{equation}\label{b132}
    \begin{aligned}
        \Delta _{T-1,j}^2&=\left ( \left ( 1-\sqrt{\beta}  \right ) ^2-\eta_{T-1}\lambda _j \right ) \left ( \left ( 1+\sqrt{\beta}  \right ) ^2-\eta_{T-1}\lambda _j \right )
\ge \frac{1}{4} \left ( 1-\sqrt{\beta}  \right ) ^2.
    \end{aligned}
\end{equation}
By Lemma~\ref{bias diedai}, we can bound $\left | \left ( \mathbf{A}_{T-1}^{\left ( j \right )}\cdots\mathbf{A}_{0}^{\left ( j \right )}  \begin{bmatrix}
1 \\1

\end{bmatrix} \right ) _1 \right |$ from above by
\begin{equation}\nonumber
    \begin{aligned}
        \left | \left ( \mathbf{A}_{T-1}^{\left ( j \right )}\cdots\mathbf{A}_{0}^{\left ( j \right )}  \begin{bmatrix}
1 \\1

\end{bmatrix} \right ) _1 \right | 
\le \left ( \prod_{i=0}^{n-1} a_{i} \right ) +\sum_{s=1}^{n-1}b_{s-1}\left ( \prod_{i=-1}^{s-2} a_{i} \right ) \left \| \left ( \mathbf{A}_{\left ( n-1 \right )K }^{\left ( j \right )} \right )^{K} \cdots \left ( \mathbf{A}_{sK }^{\left ( j \right )} \right )^{K} \right \|.
    \end{aligned}
\end{equation}
Lemma~\ref{a b} indicates that $0\le a_i<1$ and $b_s\le 8\eta_0\lambda_j\frac{1}{1-\sqrt[]{\beta }}$. For $0\le s\le T-1$, the eigenvalues of $\mathbf{A}_s^{\left ( j \right )}$ are real, according to Lemma~\ref{Fvs2} and Lemma~\ref{dandiao}, we have 
\begin{equation}\nonumber
    \left \| \left ( \mathbf{A}_{\left ( n-1 \right )K }^{\left ( j \right )} \right )^{K} \cdots \left ( \mathbf{A}_{sK }^{\left ( j \right )} \right )^{K} \right \|\le \frac{4}{\sqrt{\Delta _{\left ( n-1 \right )K,j}^2} } \left ( \rho \left ( \mathbf{A}_{\left ( n-1 \right )K }^{\left ( j \right )} \right )  \right )^{\left ( n-s \right )K-1 } .
\end{equation}
 Lemma~\ref{rho js} and \eqref{b132} indicate that $\rho \left ( \mathbf{A}_{\left ( n-1 \right )K }^{\left ( j \right )} \right ) <1$ and $\frac{1}{\sqrt{\Delta _{\left ( n-1 \right )K,j}^2} }\le\frac{2}{1-\sqrt[]{\beta } } $. Therefore, we have
 \begin{equation}\label{AT:0}
     \begin{aligned}
& \left | \left ( \mathbf{A}_{T-1}^{\left ( j \right )}\cdots\mathbf{A}_{0}^{\left ( j \right )}  \begin{bmatrix}
1 \\1

\end{bmatrix} \right ) _1 \right | 
\le 2+\sum_{s=1}^{n-1}32\frac{A\left ( 1-\sqrt[]{\beta }  \right ) }{T}\frac{1}{1-\sqrt[]{\beta } }\frac{1}{1-\sqrt[]{\beta } }
\le 32\mathrm{log}_2 T  .
 \end{aligned}
\end{equation}
Combining the above inequalities, we can bound $\lambda _j\left ( \mathbf{A}_{T-1}^{\left ( j \right )}\cdots \mathbf{A}_{0}^{\left ( j \right )} \begin{bmatrix}
1 \\1

\end{bmatrix}  \right ) ^2_1$ from above as follows
\begin{equation}\label{bi3}
    \begin{aligned}
        & \lambda _j\left ( \mathbf{A}_{T-1}^{\left ( j \right )} \cdots\mathbf{A}_{0}^{\left ( j \right )} \begin{bmatrix}
1 \\1

\end{bmatrix}  \right ) ^2_1
\le2^{10}\lambda _j\left ( \mathrm{log}_2 T \right ) ^2
\le\frac{2^{10}\left ( \mathrm{log}_2 T \right ) ^2A\left ( 1-\sqrt[]{\beta }  \right ) }{\eta_0T} .
    \end{aligned}
\end{equation}

 \noindent Consequently, combining \eqref{bi1}, \eqref{bi2}, and \eqref{bi3}, for $1\le j \le k^*$, we have
\begin{equation}\nonumber
    \lambda _j\left ( \mathbf{A}_{T-1}^{\left ( j \right )} \cdots\mathbf{A}_{0}^{\left ( j \right )} \begin{bmatrix}
1 \\1

\end{bmatrix}  \right ) ^2_1
\le\frac{2^{10}\left ( \mathrm{log}_2 T \right ) ^2A\left ( 1-\sqrt[]{\beta }  \right ) }{\eta_0T} .
\end{equation}

For $j>k^*$, by \eqref{AT:0} we have
\begin{equation}\nonumber
    \begin{aligned}
        & \lambda _j\left ( \mathbf{A}_{T-1}^{\left ( j \right )} \cdots\mathbf{A}_{0}^{\left ( j \right )} \begin{bmatrix}
1 \\1

\end{bmatrix}  \right ) ^2_1
\le2^{10}\lambda _j\left ( \mathrm{log}_2 T \right ) ^2.
    \end{aligned}
\end{equation}
\end{proof}
\renewcommand{\proofname}{Proof}

\renewcommand{\proofname}{Proof of Theorem~\ref{thm: opt rigem}}
\section{Proof of Theorem~\ref{thm: opt rigem} }
\begin{proof}
For the region of $1<b<a\le 2b$,  setting $\beta=\left ( 1- A\cdot T^{-\frac{a-b}{b}}  \right ) ^2$ and $\eta_0=\left ( 2\lambda_1 \right ) ^{-1}$, the value of  $k^*$ in Theorem~\ref{sec: result-upb} is $A^{-\frac{1}{a } }T^{\frac{1}{b}}$. According to Theorem~\ref{sec: result-upb}, the error of the output of Algorithm~\ref{alg} with initial point $\mathbf{w}_0=\mathbf{0}$, momentum $\beta=\left ( 1- A\cdot T^{-\frac{a-b}{b}}  \right ) ^2$, and initial step size  $\eta_0=\left ( 2\lambda_1 \right ) ^{-1}$ can be bounded from above by the sum of the $\mathrm{Bias}$ term and the $\mathrm{Variance}$ term. The $\mathrm{Bias}$ term is bounded from above by
    \begin{equation}\nonumber
        \begin{aligned}
             \mathrm{Bias} \lesssim &\sum_{j=1}^{A^{-\frac{1}{a } }T^{\frac{1}{b}}}\frac{\left ( \mathrm{log}_2T \right ) ^6}{T^{\frac{a}{b}}} j^{a-b}+\left ( \mathrm{log}_2T \right ) ^2\sum_{A^{-\frac{1}{a } }T^{\frac{1}{b}}}^dj^{-b}\\
\lesssim &\left ( \mathrm{log}_2T \right ) ^{6  
}\cdot T^{-1+\frac{1}{b}   }\sigma^{2-\frac{2}{b} }.
        \end{aligned}
    \end{equation}
    The $\mathrm{Variance}$ term is bounded from above by
    \begin{equation}\nonumber
        \begin{aligned}
            \mathrm{Variance} \lesssim&\left ( \mathrm{log}_2T \right ) ^4\sigma^2\left ( T^{-1+\frac{1}{b} } +\sum_{j=A^{-\frac{1}{a} }T^{\frac{1}{b}}+1}^dj^{-2a}T^{\frac{2a-b}{b}} \right ) \\ \lesssim&\left ( \mathrm{log}_2T \right ) ^4\cdot  T^{-1+\frac{1}{b}   }.
        \end{aligned}
    \end{equation}
    Consequently, the error of the output is bounded from above by 
\begin{equation}\nonumber
    \begin{aligned}
\mathbb{E} \left [ f\left ( \hat{\mathbf{w} }_T^{SHB} \right )  \right ]-f\left (\mathbf{w}_*   \right )  \le \mathrm{Bias}+\mathrm{Variance}\lesssim \left ( \mathrm{log}_2T \right ) ^{6
} \cdot T^{-1+\frac{1}{b}   }.
    \end{aligned}
\end{equation}

For the region of $1<a\le b$,
set momentum $\beta=0$, initial point $\mathbf{w}_0=\mathbf{0}$ and initial step size $\eta_0=T^{-1+\frac{a}{b} }$. Let $k^*_{SGD}=\max \left \{ k:\lambda _k\eta_0\ge \frac{\left ( 128\vee \left ( \frac{b}{a}  \right )  \right ) \mathrm{log}_2T\mathrm{ln}T}{T}  \right \}$, then $k^*_{SGD}=\mathcal{O} \left (  T^{\frac{1}{b} }\left ( \mathrm{log}_2T \right ) ^{-\frac{2}{a}}\right ) $.
For $j\le k^*_{SGD}$, by \eqref{var1} and \eqref{var2}, we have
\begin{equation}\nonumber
    \lambda _j^2\eta_t^2\left \| \mathbf{A}_{T-1}^{\left ( j \right )} \cdots \mathbf{A}_{t+1}^{\left ( j \right )}   \right \| ^2 \lesssim \frac{1}{T^2}.
\end{equation}
Additionally, by \eqref{bi1} and \eqref{bi2}, we have
\begin{equation}\nonumber
   \begin{aligned}
      & \lambda _j\left (  \mathbf{A}_{T-1}^{\left ( j \right )}\cdots \mathbf{A}_{0}^{\left ( j \right )}  \begin{bmatrix}
1 \\1

\end{bmatrix}  \right ) ^2_1\left (  \mathbf{w} _*^{\left ( j \right ) }\right ) ^2\\
=&\lambda _j\left ( 1-\eta_{T-1}\lambda _j \right )^2\cdots \left ( 1-\eta_{0}\lambda _j \right )^2\left ( \mathbf{w}_0^{\left ( j \right ) } - \mathbf{w}_*^{\left ( j \right ) }\right ) ^2 \\
\le& \mathrm{exp}\left ( -\frac{T\eta_0\lambda _j}{\mathrm{log}_2T }  \right )  \lambda _j\left ( \mathbf{w}_*^{\left ( j \right ) } \right ) ^2\\
\lesssim &T^{-\frac{b}{a}  }.
   \end{aligned}
\end{equation}
For $j> k^*_{SGD}$, we have
\begin{equation}\nonumber
    \lambda _j^2\eta_t^2\left \| \mathbf{A}_{T-1}^{\left ( j \right )} \cdots \mathbf{A}_{t}^{\left ( j \right )}  \right \| ^2
\overset{Lemma~\ref{dandiao} }{\le} \eta_0^2 \lambda _j^2\frac{16}{\Delta _{T-1,j}^2} \left ( \rho \left ( \mathbf{A}_{T-1}^{\left ( j \right )} \right )  \right )^{2\left ( T-t-1 \right )}
\overset{\ Lemma~\ref{rho js} }{\le}   64\eta_0^2\lambda _j^2T ,
\end{equation}
and
\begin{equation}\nonumber
   \begin{aligned}
      & \lambda _j\left (  \mathbf{A}_{T-1}^{\left ( j \right )}\cdots \mathbf{A}_{0}^{\left ( j \right )}  \begin{bmatrix}
1 \\1

\end{bmatrix}  \right ) ^2_1\left (  \mathbf{w} _*^{\left ( j \right ) }\right ) ^2\\
=&\lambda _j\left ( 1-\eta_{T-1}\lambda _j \right )^2\cdots \left ( 1-\eta_{0}\lambda _j \right )^2\left ( \mathbf{w}_0^{\left ( j \right ) } - \mathbf{w}_*^{\left ( j \right ) }\right ) ^2 \\
\le&\lambda _j\left ( \mathbf{w}_*^{\left ( j \right ) } \right ) ^2.
   \end{aligned}
\end{equation}
By Lemma~\ref{SHB VZK}, we can bound the $\mathrm{Variance}$ term from above by
\begin{equation}\label{small SV}
    \begin{aligned}
        \mathrm{Variance}\le &\sigma ^2\sum_{j=1}^{d} \sum_{t=0}^{T-1} \lambda _j^2\eta_t^2\left \| \mathbf{A}_{T-1}^{\left ( j \right )} \cdots \mathbf{A}_{t+1}^{\left ( j \right )}  \right \| ^2\\
        \lesssim &k^*_{SGD}\frac{\sigma^2}{T}+\sum_{j=k^*_{SGD}+1}^{d} \sigma ^2 \lambda _j^2\eta_0^2T\\
\lesssim &\left ( \mathrm{log}_2T\right )^4 \cdot T^{-1+\frac{1}{b} }.
    \end{aligned}
\end{equation}
By Lemma~\ref{SHB BZK}, we can bound the $\mathrm{Bias}$ term from above by

\begin{equation}\label{small SB}
    \begin{aligned}
       \mathrm{Bias} \le&  \sum_{j=1}^{d}\lambda _j\left (  \mathbf{A}_{T-1}^{\left ( j \right )}\cdots \mathbf{A}_{0}^{\left ( j \right )} \begin{bmatrix}
1 \\1

\end{bmatrix}  \right ) ^2_1\left (  \mathbf{w} _*^{\left ( j \right ) }\right ) ^2\\
\le &\sum_{j=1}^{k^*_{SGD}}T^{-\frac{b}{a}}+ \sum_{j=k^*_{SGD}+1}^{d}j^{-b}\\
\lesssim & \left ( \mathrm{log}_2T\right )^{\frac{2\left ( b-1 \right ) }{a} }\cdot T^{-1+\frac{1}{b} }.
    \end{aligned}
\end{equation}
Combining \eqref{small SV} and \eqref{small SB}, and using Lemma~{\ref{SHB decomposition}} we have 
\begin{equation}\nonumber
    \mathbb{E} \left [ f\left ( \hat{\mathbf{w} }_T^{SGD} \right )  \right ]-f\left (\mathbf{w}_*   \right )  \le \mathrm{Bias}+\mathrm{Variance} \lesssim \left ( \mathrm{log}_2T\right ) ^{4\vee \frac{2\left ( b-1 \right ) }{a} } \cdot T^{-1+\frac{1}{b}   }\sigma^{2-\frac{2}{b} }.
\end{equation}

\end{proof}
\renewcommand{\proofname}{Proof}

\end{document}

%% file: mathdefs.tex
\theoremstyle{plain}
\newtheorem{theorem}{Theorem}[section]
\newtheorem{proposition}[theorem]{Proposition}
\newtheorem{lemma}[theorem]{Lemma}
\newtheorem{corollary}[theorem]{Corollary}
\newtheorem{definition}[theorem]{Definition}
\newtheorem{assumption}[theorem]{Assumption}
\newtheorem{remark}[theorem]{Remark}

\newcommand{\ab}{\mathbf{a}}
\newcommand{\bbb}{\mathbf{b}}
\newcommand{\cbb}{\mathbf{c}}
\newcommand{\db}{\mathbf{d}}
\newcommand{\eb}{\mathbf{e}}
\newcommand{\fb}{\mathbf{f}}
\newcommand{\gb}{\mathbf{g}}
\newcommand{\hb}{\mathbf{h}}
\newcommand{\ib}{\mathbf{i}}
\newcommand{\jb}{\mathbf{j}}
\newcommand{\kb}{\mathbf{k}}
\newcommand{\lb}{\mathbf{l}}
\newcommand{\mb}{\mathbf{m}}
\newcommand{\nbb}{\mathbf{n}}
\newcommand{\ob}{\mathbf{o}}
\newcommand{\pb}{\mathbf{p}}
\newcommand{\qb}{\mathbf{q}}
\newcommand{\rb}{\mathbf{r}}
\newcommand{\sbb}{\mathbf{s}}
\newcommand{\tb}{\mathbf{t}}
\newcommand{\ub}{\mathbf{u}}
\newcommand{\vb}{\mathbf{v}}
\newcommand{\wb}{\mathbf{w}}
\newcommand{\xb}{\mathbf{x}}
\newcommand{\yb}{\mathbf{y}}
\newcommand{\zb}{\mathbf{z}}

\newcommand{\lambdab}{\mathbf{\lambda}}
\newcommand{\Lambdab}{\mathbf{\Lambda}}
\newcommand{\thetab}{\bm{\theta}}
\newcommand{\gammab}{\bm{\gamma}}

\newcommand{\bb}{\bm{b}}
\newcommand{\bc}{\bm{c}}
\newcommand{\bd}{\bm{d}}
\newcommand{\be}{\bm{e}}
\newcommand{\bbf}{\bm{f}}
\newcommand{\bg}{\bm{g}}
\newcommand{\bh}{\bm{h}}
\newcommand{\bi}{\bmf{i}}
\newcommand{\bj}{\bm{j}}
\newcommand{\bk}{\bm{k}}
\newcommand{\bbm}{\bm{m}}
\newcommand{\bn}{\bm{n}}
\newcommand{\bo}{\bm{o}}
\newcommand{\bp}{\bm{p}}
\newcommand{\bq}{\bm{q}}
\newcommand{\bs}{\bm{s}}
\newcommand{\bt}{\bm{t}}
\newcommand{\bu}{\bm{u}}
\newcommand{\bv}{\bm{v}}
\newcommand{\bw}{\bm{w}}
\newcommand{\bx}{\bm{x}}
\newcommand{\by}{\bm{y}}
\newcommand{\bz}{\bm{z}}

\newcommand{\Ab}{\mathbf{A}}
\newcommand{\Bb}{\mathbf{B}}
\newcommand{\Cb}{\mathbf{C}}
\newcommand{\Db}{\mathbf{D}}
\newcommand{\Eb}{\mathbf{E}}
\newcommand{\Fb}{\mathbf{F}}
\newcommand{\Gb}{\mathbf{G}}
\newcommand{\Hb}{\mathbf{H}}
\newcommand{\Ib}{\mathbf{I}}
\newcommand{\Jb}{\mathbf{J}}
\newcommand{\Kb}{\mathbf{K}}
\newcommand{\Lb}{\mathbf{L}}
\newcommand{\Mb}{\mathbf{M}}
\newcommand{\Nb}{\mathbf{N}}
\newcommand{\Ob}{\mathbf{O}}
\newcommand{\Pb}{\mathbf{P}}
\newcommand{\Qb}{\mathbf{Q}}
\newcommand{\Rb}{\mathbf{R}}
\newcommand{\Sbb}{\mathbf{S}}
\newcommand{\Tb}{\mathbf{T}}
\newcommand{\Ub}{\mathbf{U}}
\newcommand{\Vb}{\mathbf{V}}
\newcommand{\Wb}{\mathbf{W}}
\newcommand{\Xb}{\mathbf{X}}
\newcommand{\Yb}{\mathbf{Y}}
\newcommand{\Zb}{\mathbf{Z}}

\newcommand{\bA}{\bm{A}}
\newcommand{\bB}{\bm{B}}
\newcommand{\bC}{\bm{C}}
\newcommand{\bD}{\bm{D}}
\newcommand{\bE}{\bm{E}}
\newcommand{\bF}{\bm{F}}
\newcommand{\bG}{\bm{G}}
\newcommand{\bH}{\bm{H}}
\newcommand{\bI}{\bm{I}}
\newcommand{\bJ}{\bm{J}}
\newcommand{\bK}{\bm{K}}
\newcommand{\bL}{\bm{L}}
\newcommand{\bM}{\bm{M}}
\newcommand{\bN}{\bm{N}}
\newcommand{\bO}{\bm{O}}
\newcommand{\bP}{\bm{P}}
\newcommand{\bQ}{\bm{Q}}
\newcommand{\bR}{\bm{R}}
\newcommand{\bS}{\bm{S}}
\newcommand{\bT}{\bm{T}}
\newcommand{\bU}{\bm{U}}
\newcommand{\bV}{\bm{V}}
\newcommand{\bW}{\bm{W}}
\newcommand{\bX}{\bm{X}}
\newcommand{\bY}{\bm{Y}}
\newcommand{\bZ}{\bm{Z}}

\newcommand{\x}{{\mathbf{x}}}
\newcommand{\y}{{\mathbf{y}}}
\newcommand{\z}{{\mathbf{z}}}
\newcommand{\w}{{\mathbf{w}}}
\newcommand{\ba}{{\mathbf{a}}}
\newcommand{\beps}{{\boldsymbol{\epsilon}}}
\newcommand{\bepss}{{\boldsymbol{\epsilon}}^2}
\newcommand{\bphi}{{\boldsymbol{\phi}}}
\newcommand{\tmu}{{\tilde\mu}}
\newcommand{\mulambda}[1]{\mu_{\lambda, #1}}
\newcommand{\alambda}{a_\lambda}
\newcommand{\tepsilon}{\tilde\epsilon}

\newcommand{\gauss}{{\boldsymbol{\xi}}}
\newcommand{\gz}{{\boldsymbol{\zeta}}}
\newcommand{\argmin}{\mathop{\mathrm{argmin}}}

\newcommand{\I}{{\mathbf{I}}}
\newcommand{\A}{{\mathbf{A}}}
\newcommand{\B}{{\mathbf{B}}}
\newcommand{\U}{{\mathbf{U}}}
\newcommand{\D}{{\mathbf{D}}}
\newcommand{\bXi}{\boldsymbol\Xi}

\newcommand{\bbeta}{\bm{\beta}}

\newcommand{\N}{{\mathbb{N}}}
\newcommand{\tlam}{\tilde{\lambda}}

\newcommand{\tB}{{\tilde{\mathbf{B}}}}

\newcommand{\tr}{\mathrm{tr}}

\newcommand{\AGD}{\mathrm{AGD}}
\newcommand{\cO}{\mathcal{O}}
\newcommand{\hntf}{\hat\nabla_\rho\tilde f_\delta}

\newcommand{\effdim}{r}
\newcommand{\efftrace}{\mathrm{ET}}

\newcommand{\lys}[1]{{\color{blue} #1}}
\newcommand{\zhz}{\textcolor{red}}
\newcommand{\xyy}{\textcolor{cyan}}
\newcommand{\ypy}{\textcolor{purple}}
\newcommand{\congfang}{\textcolor{red}}

%% file: notation.tex
% Domains

\global\long\def\R{\mathbb{R}}%

\global\long\def\Rn{\mathbb{R}^{n}}%

\global\long\def\Rm{\mathbb{R}^{m}}%

\global\long\def\Rd{\mathbb{R}^{d}}% 

\global\long\def\Rr{\mathbb{R}^{r}}% 

\global\long\def\Rmn{\mathbb{R}^{m \times n}}%

\global\long\def\Rnm{\mathbb{R}^{n \times m}}%

\global\long\def\Rmm{\mathbb{R}^{m \times m}}%

\global\long\def\Rnn{\mathbb{R}^{n \times n}}%

\global\long\def\Z{\mathbb{Z}}%

\global\long\def\Rp{\R_{> 0}}%

\global\long\def\dom{\mathrm{dom}}%

\global\long\def\dInterior{K}%

\global\long\def\Rpm{\R_{> 0}^{m}}%

% Norms

\global\long\def\ellOne{\ell_{1}}%
 
\global\long\def\ellTwo{\ell_{2}}%
 
\global\long\def\ellInf{\ell_{\infty}}%
 
\global\long\def\ellP{\ell_{p}}%

\global\long\def\otilde{\widetilde{O}}%

\global\long\def\argmax{\argmaxTex}%

\global\long\def\argmin{\argminTex}%

\global\long\def\sign{\signTex}%

\global\long\def\rank{\rankTex}%

\global\long\def\diag{\diagTex}%

\global\long\def\im{\imTex}%

\global\long\def\enspace{\quad}%

% bold variable
\global\long\def\mvar#1{\mathbf{#1}}%

% matrix variable
\global\long\def\vvar#1{#1}%

% vector variable

% Symbol for definitions

\global\long\def\defeq{\stackrel{\mathrm{{\scriptscriptstyle def}}}{=}}%

\global\long\def\diag{\mathrm{diag}}%

\global\long\def\mDiag{\mvar{Diag}}%
 
\global\long\def\ceil#1{\left\lceil #1 \right\rceil }%

\global\long\def\E{\mathbb{E}}%

\global\long\def\jacobian{\mvar J}%

\global\long\def\onesVec{1}%
 
\global\long\def\indicVec#1{1_{#1}}%
\global\long\def\cordVec#1{e_{#1}}%
\global\long\def\op{\mathrm{spe}}%

\global\long\def\va{\vvar a}%
 
\global\long\def\vb{\vvar b}%
 
\global\long\def\vc{\vvar c}%
 
\global\long\def\vd{\vvar d}%
 
\global\long\def\ve{\vvar e}%
 
\global\long\def\vf{\vvar f}%
 
\global\long\def\vg{\vvar g}%
 
\global\long\def\vh{\vvar h}%
 
\global\long\def\vl{\vvar l}%
 
\global\long\def\vm{\vvar m}%
 
\global\long\def\vn{\vvar n}%
 
\global\long\def\vo{\vvar o}%
 
\global\long\def\vp{\vvar p}%
 
\global\long\def\vs{\vvar s}%
 
\global\long\def\vu{\vvar u}%
 
\global\long\def\vv{\vvar v}%
 
\global\long\def\vx{\vvar x}%
 
\global\long\def\vy{\vvar y}%
 
\global\long\def\vz{\vvar z}%
 
\global\long\def\vxi{\vvar{\xi}}%
 
\global\long\def\valpha{\vvar{\alpha}}%
 
\global\long\def\veta{\vvar{\eta}}%
 
\global\long\def\vphi{\vvar{\phi}}%
\global\long\def\vpsi{\vvar{\psi}}%
 
\global\long\def\vsigma{\vvar{\sigma}}%
 
\global\long\def\vgamma{\vvar{\gamma}}%
 
\global\long\def\vphi{\vvar{\phi}}%
\global\long\def\vDelta{\vvar{\Delta}}%
\global\long\def\vzero{\vvar 0}%

\global\long\def\ma{\mvar A}%
 
\global\long\def\mb{\mvar B}%
 
\global\long\def\mc{\mvar C}%
 
\global\long\def\md{\mvar D}%
 
\global\long\def\mf{\mvar F}%
 
\global\long\def\mg{\mvar G}%
 
\global\long\def\mh{\mvar H}%
 
\global\long\def\mj{\mvar J}%
 
\global\long\def\mk{\mvar K}%
 
\global\long\def\mm{\mvar M}%
 
\global\long\def\mn{\mvar N}%

\global\long\def\mO{\mvar O}%

\global\long\def\mq{\mvar Q}%
 
\global\long\def\mr{\mvar R}%
 
\global\long\def\ms{\mvar S}%
 
\global\long\def\mt{\mvar T}%
 
\global\long\def\mU{\mvar U}%
 
\global\long\def\mv{\mvar V}%
 
\global\long\def\mx{\mvar X}%
 
\global\long\def\my{\mvar Y}%
 
\global\long\def\mz{\mvar Z}%
 
\global\long\def\mSigma{\mvar{\Sigma}}%
 
\global\long\def\mLambda{\mvar{\Lambda}}%
\global\long\def\mPhi{\mvar{\Phi}}%
 
\global\long\def\mZero{\mvar 0}%
 
\global\long\def\iMatrix{\mvar I}%
\global\long\def\mi{\mvar I}%
\global\long\def\mDelta{\mvar{\Delta}}%

% Common Variables
\global\long\def\oracle{\mathcal{O}}%
 
\global\long\def\mw{\mvar W}%

\global\long\def\runtime{\mathcal{T}}%

% weights

\global\long\def\mProj{\mvar P}%

\global\long\def\vLever{\sigma}%
 
\global\long\def\mLever{\mSigma}%
 
\global\long\def\mLapProj{\mvar{\Lambda}}%

\global\long\def\penalizedObjective{f_{t}}%
 
\global\long\def\penalizedObjectiveWeight{{\color{red}f}}%

\global\long\def\fvWeight{\vg}%
 
\global\long\def\fmWeight{\mg}%

\global\long\def\vNewtonStep{\vh}%

\global\long\def\norm#1{\|#1\|}%
 
\global\long\def\normFull#1{\left\Vert #1\right\Vert }%
 
\global\long\def\normA#1{\norm{#1}_{\ma}}%
 
\global\long\def\normFullInf#1{\normFull{#1}_{\infty}}%

\global\long\def\normFullSquare#1{\normFull{#1}_{\square}}%
 
\global\long\def\normInf#1{\norm{#1}_{\infty}}%
 
\global\long\def\normOne#1{\norm{#1}_{1}}%
 
\global\long\def\normTwo#1{\norm{#1}_{2}}%
 
\global\long\def\normLeverage#1{\norm{#1}_{\mSigma}}%
 
\global\long\def\normWeight#1{\norm{#1}_{\fmWeight}}%

\global\long\def\cWeightSize{c_{1}}%
 
\global\long\def\cWeightStab{c_{\gamma}}%
 
\global\long\def\cWeightCons{{\color{red}c_{\delta}}}%

\global\long\def\TODO#1{{\color{red}TODO:\text{#1}}}%
\global\long\def\mixedNorm#1#2{\norm{#1}_{#2+\square}}%

\global\long\def\mixedNormFull#1#2{\normFull{#1}_{#2+\square}}%
\global\long\def\CNorm{C_{\mathrm{norm}}}%
\global\long\def\Pxw{\mProj_{\vx,\vWeight}}%
\global\long\def\vq{q}%
\global\long\def\cnorm{\CNorm}%

\global\long\def\next#1{#1^{\mathrm{(new)}}}%

\global\long\def\trInit{\vx^{(0)}}%
 
\global\long\def\trCurr{\vx^{(k)}}%
 
\global\long\def\trNext{\vx^{(k + 1)}}%
 
\global\long\def\trAdve{\vy^{(k)}}%
 
\global\long\def\trAfterAdve{\vy}%
 
\global\long\def\trMeas{\vz^{(k)}}%
 
\global\long\def\trAfterMeas{\vz}%
 
\global\long\def\trGradCurr{\grad\Phi_{\alpha}(\trCurr)}%
 
\global\long\def\trGradAdve{\grad\Phi_{\alpha}(\trAdve)}%
 
\global\long\def\trGradMeas{\grad\Phi_{\alpha}(\trMeas)}%
 
\global\long\def\trGradAfterAdve{\grad\Phi_{\alpha}(\trAfterAdve)}%
 
\global\long\def\trGradAfterMeas{\grad\Phi_{\alpha}(\trAfterMeas)}%
 
\global\long\def\trSetCurr{U^{(k)}}%
\global\long\def\vWeightError{\vvar{\Psi}}%
\global\long\def\code#1{\texttt{#1}}%

\global\long\def\nnz{\mathrm{nnz}}%
\global\long\def\tr{\mathrm{tr}}%
\global\long\def\vones{\vec{1}}%

\global\long\def\volPot{\mathcal{V}}%
\global\long\def\grad{\mathcal{\nabla}}%
\global\long\def\hess{\nabla^{2}}%
\global\long\def\hessian{\nabla^{2}}%

\global\long\def\shurProd{\circ}%
 
\global\long\def\shurSquared#1{{#1}^{(2)}}%
\global\long\def\solver{\mathrm{\mathtt{S}}}
\global\long\def\time{\mathrm{\mathcal{T}}}
\global\long\def\trans{\top}%

\global\long\def\lpweight{w_{p}}%
\global\long\def\mlpweight{\mw_{p}}%
\global\long\def\lqweight{w_{q}}%
\global\long\def\mlqweight{\mw_{q}}%

\newcommand{\bracket}[1]{[#1]}
\global\long\def\interiorPrimal{\Omega^{\circ}}%
\global\long\def\interiorDual{\Omega^{\circ}}%

\newcommand{\tx}{\tilde{\mathbf{x}}}
\newcommand{\errorA}{\epsilon_A}
\newcommand{\errorB}{\epsilon_A}
\newcommand{\errorC}{\epsilon_B}
\newcommand{\errorD}{\epsilon_C}
\newcommand{\BSa}{\mathop{\mathtt{CBinarySearch}}}
\newcommand{\bbv}{\mathbf{v}}
\newcommand{\BSb}{\mathop{\mathtt{CCubicBinarySearch}}}
\newcommand{\rtemp}{r_{\mathrm{temp}}}
\newcommand{\ltemp}{\gamma_{\mathrm{temp}}}
\newcommand{\tilder}{\tilde{r}}
\newcommand{\elltemp}{l_{\mathrm{temp}}}